\documentclass[12pt, letterpaper]{article}
\pdfoutput=1 
\usepackage[utf8]{inputenc}
\usepackage{mathptmx}

\usepackage[square,sort,comma,numbers]{natbib}
\usepackage[a4paper, total={6in, 10in}]{geometry}
\usepackage{lineno,hyperref}
\usepackage{mathtools}
\usepackage{amsmath}
\usepackage[table]{xcolor}
\usepackage{xcolor,colortbl} 
\usepackage{booktabs}
\usepackage{graphicx}
\usepackage{subcaption}
\usepackage{multirow}
\usepackage{xr}
\usepackage{authblk}
\usepackage{wrapfig}
\usepackage{lscape}
\usepackage{rotating}
\usepackage{epstopdf}
\usepackage{verbatim,amssymb,amsmath}
\usepackage{graphicx,epsfig,bm}
\usepackage{tabularx}
\usepackage{subcaption} 
\usepackage{cleveref}
\usepackage{multicol, blindtext}
\usepackage{appendix}

\usepackage{mwe}

\providecommand{\keywords}[1]{\textbf{\textit{Index terms---}} #1}
\newcommand{\ttvar}{\begingroup\@makeother\#\@ttvar}
\newcommand*{\myfont}{\fontfamily{ptm}\selectfont}

\title{\myfont{\Large{REFINED (\textbf{RE}presentation of \textbf{F}eatures as \textbf{I}mages with \textbf{NE}ighborhood \textbf{D}ependencies): A novel feature representation for Convolutional Neural Networks}}}

\author[1]{Omid Bazgir}
\author[1]{Ruibo Zhang}
\author[1]{Saugato Rahman Dhruba}
\author[1]{Raziur Rahman}
\author[2,3]{Souparno Ghosh}
\author[1]{Ranadip Pal *}
\affil[1]{\footnotesize Department of Electrical and Computer Engineering, Texas Tech University}
\affil[2]{Department of Mathematics and Statistics, Texas Tech University}
\affil[3]{Department of Statistics, University of Nebraska-Lincoln}
\affil[ ]{\text{ranadip.pal@ttu.edu}}

\date{\vspace{-5ex}}

\begin{document}
	\maketitle
\noindent\hrulefill 
\begin{abstract}
	Deep learning with Convolutional Neural Networks has shown great promise in various areas of image-based
	classification and enhancement but is often unsuitable for predictive modeling involving non-image based
	features or features without spatial correlations. We present a novel approach for representation of high dimensional feature vector in a compact image form, termed REFINED (\textbf{RE}presentation of \textbf{F}eatures as \textbf{I}mages with \textbf{NE}ighborhood \textbf{D}ependencies), that is conducible for convolutional neural network based deep learning. We consider the correlations between features to generate a compact representation of the features in the form of a two-dimensional image using minimization of pairwise distances similar to multi-dimensional scaling. We hypothesize that this approach enables embedded feature extraction and integrated with Convolutional Neural Network based Deep Learning can produce more accurate predictions as compared to other methodologies such as Artificial Neural Networks, Random Forests and Support Vector Regression. We illustrate the superior predictive performance of the proposed representation, as compared to existing approaches, using synthetic datasets, cell line efficacy prediction based on drug chemical descriptors for NCI60 dataset and drug sensitivity prediction based on transcriptomic data and chemical descriptors using GDSC dataset.  Results illustrated on both synthetic and biological datasets shows the higher prediction accuracy of the proposed framework as compared to existing methodologies while maintaining desirable 	properties in terms of bias and feature extraction.
\end{abstract}

\noindent\hrulefill 

\section{Introduction}

In recent years, machine learning has been able to produce numerous insights from the surge of data generated in diverse areas. For instance, the area of computational biology has benefited from the availability of high throughput information for genome, transcriptome, proteome and metabolome. These large datasets often have the issue of numerous features with limited samples which necessitates the use of feature selection or feature extraction prior to modeling. A predictive modeling framework that has high accuracy and incorporates inbuilt feature extraction or selection can be highly useful in such circumstances. In pharmacogenomics studies, in order to predict drug efficacy, based on genomic characterizations, various types of machine learning approaches such as Random Forests, Elastic Net, Kernelized Bayesian Multi Task learning, Support vector Regression have been proposed \cite{Costello:2014, rahman2017heterogeneity, rahman2017integratedmrf, Wan:PLOS} where an initial step of feature selection or extraction has been included before model building.
 Although, sparse linear regression approaches such as Lasso and Elastic Net do offer embedded feature selection but the accuracy of the models are significantly lower than ensemble, kernel and non-linear regression approaches under model misspecification \cite{Costello:2014}. On the other hand, deep convolutional neural networks (CNN) has the potential to provide high accuracy prediction while automatically discovering multiple levels of joint representation of the data and thus eliminating the need for feature engineering or selection \cite{lecun2015deep}. 
CNN bypasses the {\it a priori} manual extraction of features by learning them from data \cite{angermueller2016deep}. Furthermore, their representational richness often allows capturing of nonlinear dependencies at multiple scales \cite{wainberg2018deep}, and minimizing generalization error rather than the training error \cite{bengio2012practical}.

CNN based Deep Learning methods have shown improved performance in speech recognition, object recognition \cite{iandola2016squeezenet}, natural language processing \cite{xu2018prediction}, genomics \cite{alipanahi2015predicting} and cancer therapy \cite{coudray2017classification}. Deep (multi-layered) neural networks are especially well-suited for learning representations of data that are hierarchical in nature, such as images or videos \cite{ruff2018deep}. CNN-based methods have achieved close to human-level performance in object classification, where a CNN learns to classify the object contained in an image \cite{esteva2019guide}.
In the computational biology area, Alipanahi et al \cite{alipanahi2015predicting} used 1-D CNN architecture to predict specificities of DNA and RNA binding proteins by directly training on the raw DNA sequence. 
Note that 1-D CNN can be directly applied to scenarios where the features have relationships with neighbors such as DNA or RNA sequences. However, a 1-D CNN will not be highly effective in scenarios where ordering of features does not describe the dependencies among features. For instance, gene expressions or chemical descriptors, in their raw form, do not exhibit any form of ordering and hence not amenable to 1-D CNN.   

If, on the other hand, the predictors are in form of images, a CNN model is often effective because the spatial correlation among the neighbors can be exploited to reduce the number of model parameters compared to fully connected network by applying convolutional operations and sharing parameters. In classification setup, \cite{coudray2017classification} demonstrated the efficiency of this approach  to distinguish  the most prevalent subtype of lung tumor from normal lung tissue using whole slide images of The Cancer Genome Atlas (TCGA) dataset and validating on independent histopathology images.
Thus, the ability to represent a collection of potentially high dimensional scalar features as images, with correlated neighborhoods, has the potential of benefiting from the automated feature extraction and high accuracy predictions of CNN based Deep Learning. To our knowledge, the only other approach for representing data as images is OmicsMapNet \cite{ma2018omicsmapnet} that has been proposed at the same time while we were developing our REFINED idea. OmicsMapNet uses treemap \cite{shneiderman1998tree} to rearrange omics data into 2D images which requires preliminarily knowledge extracted from KEGG. OmicsMapNet cannot be used when there is no ontology knowledge on the omics data, or when the covariates are non-omics data such as drug descriptors.

In this paper, we present a novel methodology, termed REFINED (\textbf{RE}presentation of \textbf{F}eatures as \textbf{I}mages with \textbf{NE}ighborhood \textbf{D}ependencies), for representing high dimensional feature vectors as mathematically justifiable 2D Images that can be processed by standard convolutional neural network based deep learning methodologies. We illustrate the advantages of our proposed framework in terms of accuracy and bias characteristics on both synthetic and biological datasets.

\section{Materials and Methods}

In this section we introduce our proposed REFINED algorithm that maps high-dimensional feature vectors to images, describe the datasets used for performance evaluation, followed by the CNN architecture used as the predictive model.

\subsection{REpresentation of Features as Images using NEighborhood Dependencies (REFINED)}

As mentioned earlier, the main idea of the REFINED CNN approach is to map high-dimensional vectors to mathematically justifiable images for training by traditional CNN architecture. Evidently, a mapping of features from the high dimensional vector to a 2-D image matrix serially in a row by row or column by column fashion will not guarantee any spatial correlations in the image. Instead, we first 
obtain the euclidean distance matrix of the features and use it as a distance measure to generate a compact 2D feature representation where neighborhood features are closely related. A potential solution to achieve this 2-D projection is to apply dimensionality reduction approach such as Multidimensional Scaling (MDS) \cite{davison1983multidimensional} on a distance measure such as euclidean distance of features. However, that will not guarantee that each mapped point will have a unique pixel representation in the image and might result in sparse images due to the overlap \cite{urpa2019focused}. For instance, if we have 900 features, the features can potentially be represented by a $30 \times 30$ matrix and a direct MDS like approach on a $30 \times 30$ dimension space might not spread out each feature in such a manner that each pixel contains at most one feature. To ensure that the features are spaced out in a discrete grid and to incorporate the discrete nature of the image pixels, we apply a Bayesian version of metric MDS. We start with the MDS algorithm to create an \emph{initial feature map} (a 2-D space with feature coordinates) that preserves the feature distances in the 2-D space with minor computational cost. Next, we apply the \emph{Bayesian MDS} (BMDS) to estimate the feature location on a bounded domain with the constraint that each pixel can at most contain one feature. However, the location of the features are estimated up to an automorphism.  Therefore, we apply a hill climbing algorithm, with a cost function that measures the absolute difference in the Euclidean distances among the new feature locations (as represented by the 2D image map) to the  estimated true distances ($\hat{\bm{\delta}}$, anticipating the following section) among the features, to arrive at an {\it optimal} configuration. 

More specifically, starting from the BMDS location estimates, we considered all the configurations in the map sequentially in row-order. For each feature, we tried different permutations of the features by interchanging the position of the central feature position with its neighboring features and selected the permutation that  minimizes the above mentioned cost function. Once the cost function is minimized, a set of unique coordinates in a 2D space was produced for each feature. Using those coordinates, we mapped the features into a 2D space and create an image per sample. The created images were then used to train the REFINED CNN.

\begin{figure}[!htbp]
	\centering
	\includegraphics[width=5in]{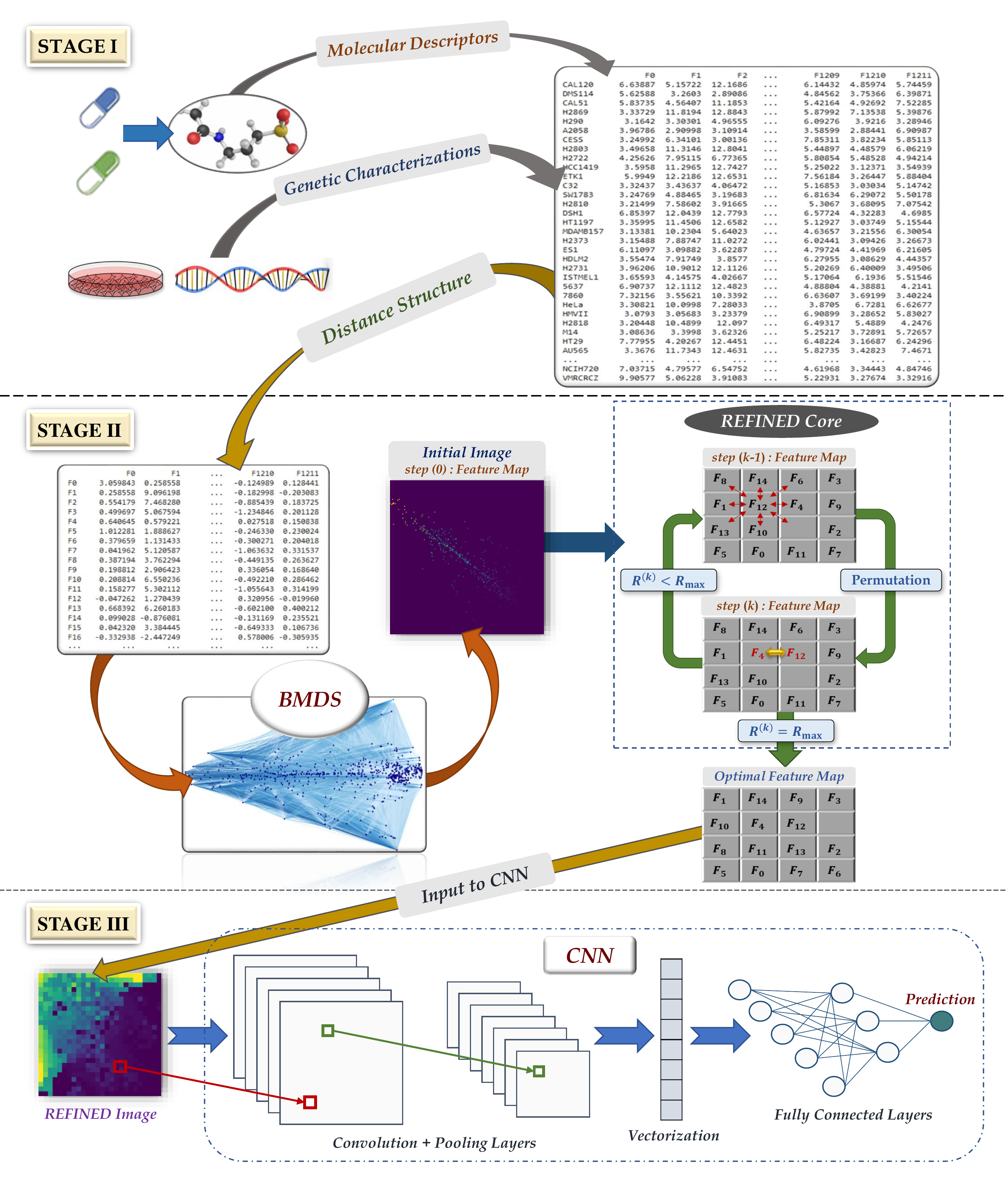}
	\caption{Overview of training a CNN using REFINED images which mainly transforms cell line-drug pairs data structures into images, \textbf{(STAGE I)} distance matrix of the 672 PaDEL descriptors of the cell line-drug pairs are used to calculate the distance between features, \textbf{(STAGE II)} Initial BMDS images generated using the initial distance matrix, then hill climbing algorithm utilized as an iterative optimizer to maximize the correlation between features, by relocating the features (descriptors) in the distance matrix. \textbf{(STAGE III)} A CNN is trained using REFINED images and drug responses of the 60 drugs of NCI60 dataset to predict drug sensitivity of paired cell lines.}
	\label{fig:MDS_IMAGE_gen}
\end{figure}

The general idea of the REFINED CNN approach is shown pictorially in Figure \ref{fig:MDS_IMAGE_gen} for the application case of predicting drug efficacy over a cell line using genetic characteristics of cell lines and chemical descriptors of the drug as predictors.\footnote{Note that we use PaDEL features as chemical descriptors of each drug.} In Figure \ref{fig:MDS_IMAGE_gen}, an example case is shown where \emph{F-12} has been interchanged with its neighboring features, and after each exchange, we checked the similarities/correlation among distances of features from the map and estimated distance matrix of descriptors. If we can find a better exchange case in the feature map, we exchange that feature pair and arrive at a new feature map. The entire process was repeated iteratively until we reached the optimized feature map that is close to the \emph{benchmark} distance matrix ($\hat{\bm{\delta}}$) of the initial features (in this case PaDEL descriptors).

At the conclusion of this iterative algorithm, we arrive at a REFINED feature map with all features having a unique position in a bounded 2-D space and similar features are placed close by and dissimilar features are far apart. Without loss of generality, we have considered feature maps on unit square and the BMDS specification induced sparsity in the image.

\begin{figure*}
	\centering
	\begin{subfigure}[b]{0.495\textwidth}
		\centering
		\includegraphics[width=\textwidth]{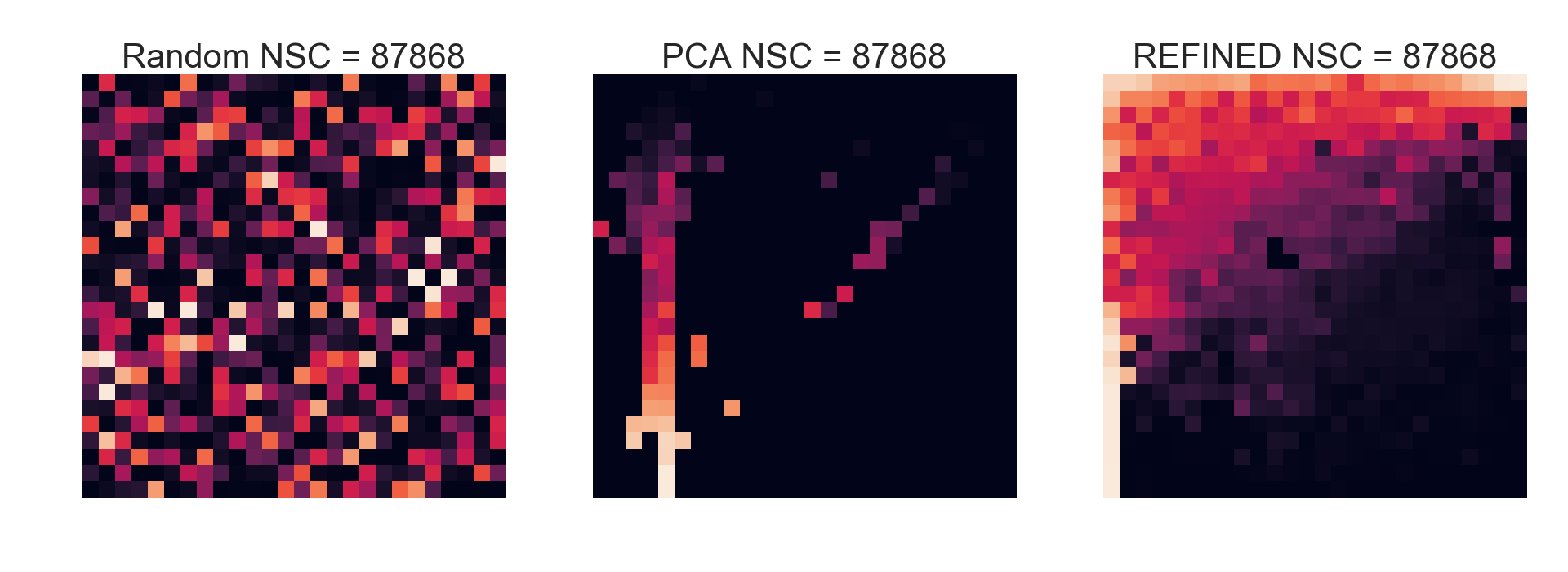}
		
		\label{IM_Gen:a}
	\end{subfigure}
	\hfill
	\begin{subfigure}[b]{0.495\textwidth}  
		\centering 
		\includegraphics[width=\textwidth]{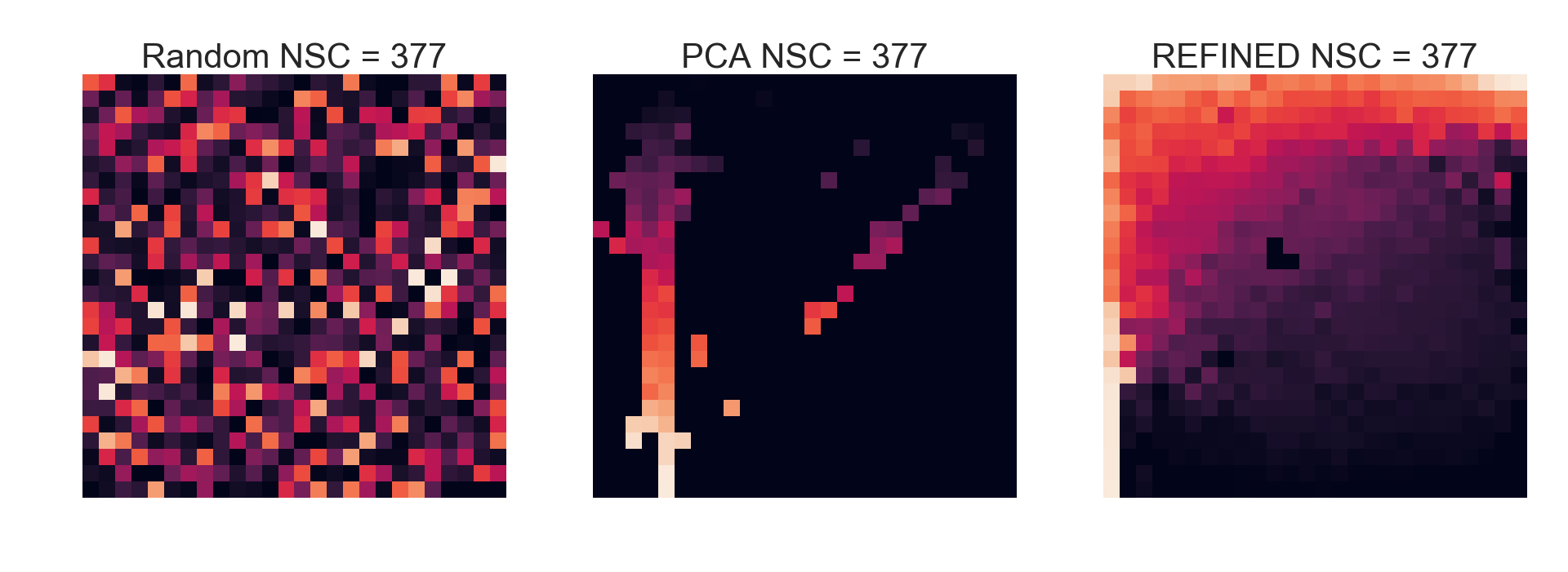}
		
		\label{IM_Gen:b}
	\end{subfigure}

	\caption{Illustration of generated images randomly selected from the NCI60 dataset, where the caption denotes the method that is utilized to generate the image followed by the NSC ID associated with the image.}
	\label{IM_Gen}
\end{figure*}

Figure \ref{IM_Gen} shows some generated REFINED images for different drugs. Each image varies from another depending on the value of the PaDEL descriptors of the drug, but the descriptor coordinates are same for all the cases.

\subsubsection{Theoretical Basis for REFINED} 
\label{thre}

Consider the predictor matrix $\bm{X}=\{x_{ij}\}, i=1,2,..., n; \; j=1,2,...,p$ with $x_{ij}$ being the value of the $j$th predictor for the $i$th subject. Suppose, the predictors are generated from a latent zero mean, square integrable stochastic process $\{Z(s)\}$ where the index $s$ belongs to a compact subset of $\mathbb{R}^m$. Let $s_{j}$ denote the original position of the $j$th predictor produced by $Z(s)$ and the observed data is randomly permuted version of the original data, i.e. $x_{ij}= Z_i(s_j)$. 

{\it Case 1:} There is a underlying true ordering of the predictors , i.e.,  there exists a permutation $\{\pi(1),...,\pi(p)\}$ of $\{1,2,...,p\}$ such that $s_{\pi(1)}<s_{\pi(2)}<...< s_{\pi(p)}$ is the true, but unknown, ordering of the predictors. If such ordering exists, we can take $m=1$ and the predictors can be projected on $[0,1]$ via unidimensional scaling (UDS). Let $\{\hat{s}_1,...,\hat{s}_p\}$ be the estimated locations of the $p$ predictors on $[0,1]$ obtained via UDS. Let $\{\psi(1),...,\psi(p)\}$ be the permutation of $\{1,2,...,p\}$ that orders $\{\hat{s}_1,...,\hat{s}_p\}$. Then under some regularity conditions $\psi(j)=\pi(j),\; 1\leq j\leq p,\; \forall p$. Thus UDS can correctly identify the true relative pairwise distances among the predictors.

For proof, see \cite{chen2011stringing}.

{\it Case 2:} Suppose the ordering does not exist. For example, suppose three predictors 
are equidistant from one another. Clearly, $m=1$ may not be a valid assumption and results corresponding to Case 1 become untenable in this situation. For the second order approximation, we start with $m=2$, i.e., we would like to obtain the location of the predictors in a compact subset of $\mathbb{R}^2$. Without loss of generality, we project the locations on unit square ($[0,1]^2$).

Let $d_{jk}$ be the observed distance between the $j$th and the $k$th predictor and $\delta_{jk}$ be their true, but unobserved, distances. Under the assumption of Euclidean metric,$\delta_{jk}=\sqrt{\sum_l (s_{j,l}-s_{k,l})^2}$, where $s$ is now 2D coordinate system denoting the true location of the predictors $j$ and $k$ in unit square. As in Case 1, we can assume that $\pi(.)$ is the underlying true permutation of  2D configurations of the $p$ predictors. Our goal is to draw inference on the locations of each predictor, i.e. estimate $s_j \in [0,1]^2$.

\cite{oh2001bayesian} developed a Bayesian estimation procedure to estimate $s$ based on observed distance by assuming  $d_{jk}\sim N(\delta_{jk},\sigma^2)I(d_{jk}>0)$ at the data level. For the location process, we specify a spatial Homogeneous Poisson Process (HPP) with constant intensity $\lambda=p/[0,1]^2$ which essentially distributes locations of $p$ predictors randomly in an unit square. Since this corresponds to complete spatial randomness, an alternative specification of location process is given by $\bm{s}=\{s_1,s_2,...,s_p\}\sim Uniform([0,1]^2)$ \cite{chandler2013spatially}. Note that, the properties of HPP guarantee that the HPP operating on the unit square can be further partitioned into disjoint cells and the entire location process can be expressed as the superposition of the HPPs operating on these disjoint cells. Futhermore, as the volume of each cell (within the unit square) goes to zero, so does the probability of observing more than one event in that cell \cite{illian2008statistical}. We also note that, we do not assume that the location process shows any clustering tendency {\it a-priori}, however the uniform specification is flexible enough to capture clustering {\it a-posteriori}  \cite{chandler2013spatially}.   
Let us denote the set of observed and true distances by $\bm{d}$ and $\bm{\delta}$, respectively. Our data model is then given by
\begin{equation}
f(\bm{d}|\bm{s},\sigma^2)\propto (\sigma^2)^{-q/2}\exp\left[-\frac{1}{2\sigma^2}\sum_{j>k} (d_{jk}-\delta_{jk})^2 -\sum_{j>k}\log \Phi(\delta_{jk}/\sigma)\right]
\label{likelihood}
\end{equation}
where $q$=$p\choose 2$ is the total number of distances in the dataset and $\Phi(.)$ is the usual standard normal cdf.
At the process level, we have 
\begin{equation}
\bm{s}|p  \sim Uniform([0,1]^2)
\end{equation}
Finally, the prior is given by  $\sigma^2 \sim IG(a,b)$ with  $a>2,\;\; b>0$ and $IG$ denoting the Inverse Gamma distribution. 
Consequently, the full posterior distribution is given by 

\begin{equation}
[\bm{s},\sigma^2|\bm{d}] \propto (\sigma^2)^{-(q/2+a+1)}\exp\left[-\frac{1}{2\sigma^2}\sum_{j>k} (d_{jk}-\delta_{jk})^2 -\sum_{j>k}\log \Phi(\delta_{jk}/\sigma)-b/\sigma^2\right]
\label{posterior 1}
\end{equation}

\noindent When $q$ is large, $\sum \log \Phi(.) \approx 0$, the full conditional posterior of $\sigma^2|.$ is approximated by 
$IG(q/2+a,\frac{1}{2}\sum_{j>k} (d_{jk}-\delta_{jk})^2+b)$. However, the conditional posterior of $\bm{s}$ is not available in closed form, a Metropolis-in-Gibbs sampler is used to obtain posterior realizations of the locations. Since $\bm{s}$ are identifiable only up to an automorphism, convergence of the Markov Chain Monte Carlo (MCMC) is assessed on $\bm{\delta}$ and $\sigma^2$. Furthermore, following the recommendation of \cite{oh2001bayesian}, we used the posterior mode of $\bm{s}$ as the point estimate of the predictor {\it location}.
    
Once these locations are estimated, we have a set of point-referenced predictor location in the domain of interest. The domain is then subjected to regular square tessellation such that each pixel contains at most one location -- essentially rewriting the point process {\it a-posteriori} as superposition of constant intensity local point processes operating on disjoint cells that the unit square in partitioned into. Additionally, the constraint of allowing at most one predictor location per-pixel implies that the posterior location process is approximated by a Non-homogeneous Poisson process \cite{illian2008statistical}. The foregoing hill climbing algorithm is then applied to arrive at {\it locally optimal} configuration \footnote{The location process is not uniquely identifiable. Local adjustment via hill climbing is done to mimic the swapping step of the partition-around-medoid algorithm.}.

Once the tiled surface associated with the feature space is obtained, we have the observed value, $x(s_j)$, for each row of $\bm{X}$. The intensity at each pixel is, therefore, determined by $x(s_j)$. Pixels that do not contain any predictor location are assigned  null values. This pixelated image on unit square is our second order approximation of the random functions developed in \cite{chen2011stringing}. We can then deploy any suitable smoothing operation (for example, autoregressive spatial smoothing \cite {macnab2001autoregressive}) to generate the corresponding predictor {\it images} (such as shown in figure \ref{IM_Gen}). Furthermore, each posterior realization of $\bm{s}$ can be used for data augmentation purpose in CNN architecture. Also, since Euclidean metric is invariant under translation, rotation and reflection about the origin, any such perturbation will not affect the relationship between the response and predictors. 

Note that, even if there exists ordering among the covariates, we can still generate these images in the following way. Since \cite{chen2011stringing} guarantees that the  relative pairwise distances  among the predictors, estimated from UDS, are consistent estimators of the true relative distances, we can posit a calibration model for these estimates $\hat{d}_{jk}$ to connect with the true distance, i.e. $\hat{d}_{jk}\sim N(\alpha_0+\alpha_1\delta_{jk},\sigma^2)I(d_{jk}>0)$.

\subsection{Datasets and Preprocessing}

To evaluate our framework, we considered three datasets: (a) A synthetically generated dataset (b) \emph{NCI 60 dataset} consisting of drug responses following application of more than 52,000 unique compounds on 60 human cancer cell lines \cite{shoemaker2006nci60} (c) \emph{ Genomics of Drug Sensitivity in Cancer (GDSC) }\cite{yang2012genomics} dataset that contains responses to 222 anticancer drugs across approximately 972 cancer cell lines with known genomic information.
In scenario (b), we use the chemical descriptors of drugs to predict drug responses in a specific cell line. In scenario (c), we consider two heterogeneous predictor set- (i) gene expressions for cancer cell lines and (ii) chemical descriptors for applied drugs and use both these type of predictors to predict drug responses . 

\subsubsection*{(a) Synthetic dataset}

We simulated a synthetic dataset with $P$ correlated features for $N$ samples, where for each simulation 20\%, 50\% and 80\% of the features were spurious. The features were simulated from a zero mean Gaussian process with stationary isotropic covariance matrix whose $(i,j)$th element is given by $\gamma^{|i-j|}$. $P$ ranged from $20, 50,100,400,800,1000,2000,4000$ and for each P, the number of simulated sampled ranged from ${50,200,600,1000,2000,5000,10000}$. We simulated the target values by simply multiplying random weights to the features. For example, $N$ target values with 100 features ($\bm{X}^{N\times 100})$, with 20\% spurious features were generated using the relation $\bm{X} [\bm{\beta}_r,\bm{\beta}_0]^T$ where $\bm{\beta}_r^{80\times 1}$ are non-zero random weights and $\bm{\beta}_0^{20\times 1}$ are zeros.

\subsubsection*{(b) NCI dataset}

\begin{figure}[!htbp]
	\centering
	\includegraphics[width=\textwidth,height=0.3\textheight]{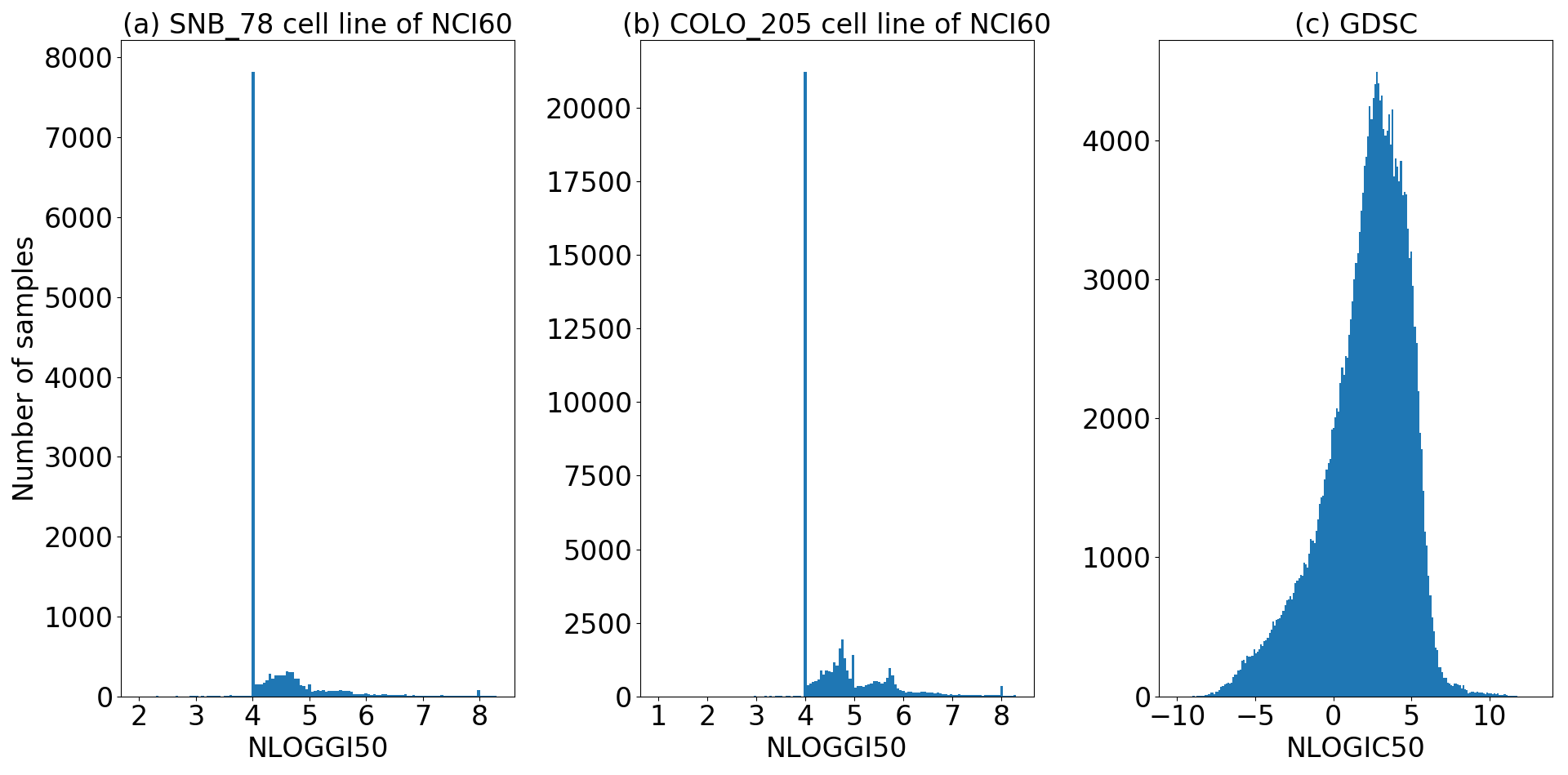}
	\caption{(a), (b) Drug response distribution of $SNB\_78$ and $COLO\_205$ cell lines of NCI60 dataset in natural logarithmic scale, where all the cell lines response distribution with more than 10,000 samples have the same large peak, (c) represents the drug response distribution in natural logarithmic scale of all the drugs that had been applied on all the cell lines available in the GDSC dataset.}
	\label{Histogram}
\end{figure}

The US National Cancer Institute (NCI) screened more than $52,000$ unique chemicals on around 60 human cancer cell lines. The chemical (drug) response is reported as GI50 which is the concentration required to achieve 50\% of maximal inhibition of cell proliferation \cite{shoemaker2006nci60}. All the chemicals have an associated unique NSC identifier number which is assigned to identify agents when they are submitted for clinical trials to the Cancer Therapy Evaluation Program (CTEP). We used the  NSC identifiers to obtain the chemical descriptor features and then used PaDEL software \cite{yap2011padel} to extract these features for each one of the chemicals.  The chemicals with more than 10 \% of their descriptor values being zero or missing were discarded. The final dataset consists of $52,126$ chemicals, each with 672 descriptor features and 59 cancer cell lines. To incorporate the logarithmic nature of dose administration protocol, we calculated the negative-log concentration of GI50s (NLOGGI50). The drug response distribution for one illustrative cell lines is shown in figures \ref{Histogram} (a). We selected 17 cell lines with more than 10k drugs, to ensure availability of enough data points for training deep learning models. The number of drugs applied on each selected cell line are provided in table \ref{NCI_sample} of the appendix. All these preprocessing steps were conducted on the training phase.

\subsubsection*{(c) GDSC dataset}

For validation of our framework, we also considered the Genomics of Drug Sensitivity in Cancer (GDSC) \cite{yang2012genomics} dataset which describes the responses to 222 anticancer drugs across approximately 972 cancer cell lines. The resulting drug-cell line pairs have the responses reported in the form of IC50 which denotes the concentration of the drug to reduce the response by 50\%. The normalized drug response distribution is shown in figure \ref{Histogram}(c). 
We used the gene expression of the cell lines data (17,737 genes), and PaDel descriptors of drug data as predictors. Descriptors with more than 10 \% zero or missing values were removed from the dataset, and the rest of missing values were imputed using KNN. Finally, 171K drug-cell line pairs (953 Cell lines, and 206 Drugs) are available for training.

Since in prior drug sensitivity studies \cite{Costello:2014} \cite{Wan:PLOS} \cite{pal2016predictive}, an initial feature selection was used to reduce the number of genes before training the predictive model, we also used an {\it a-priori} RELIEFf based feature selection \cite{Kira:1992:FSP:1867135.1867155}  for scenario (c) for easier comparison with earlier studies. Since multiple drugs were tested on each cell line, a common subset of 1211 genes were selected that were found to be common among top 8048 $ \sim $ 8000 genes selected for each drug. Similar to NCI60 scenario, we used PaDEL software to obtain the chemical descriptor features. The chemical descriptors with more than 10 \% of their descriptor values being zero or missing were discarded. Finally each drug has 992 descriptors as features. The gene selection process was conducted on the training set, then the same set of genes were used for validation and test. All the gene expressions and drug descriptors were normalized between 0 and 1.

\subsection{Predictive Models}

We used the REFINED images to train a CNN regressor to predict drug sensitivity of the NCI60, and GDSC datasets, as well as the simulated target values of the synthetic datasets. We compared performance of the REFINED CNN with Random Forest (RF), Support Vector Regressor (SVR), a deep Artificial Neural Network (ANN) for synthetic dataset. For other datasets, along with RF, SVR, and ANN, we compared the REFINED CNN with Elastic Net (EN), CNN with images created by random 2D projection matrix (Random CNN), and CNN with images created using principal component analysis (PCA CNN) coordinates. The details on Random CNN and PCA CNN are included in section \ref{section:comparison}. We randomly picked 17 cell lines where each cell line contained more than 10k drug responses so that we have sufficient number of samples to train a deep learning model. Subsequently, 17 set of models were trained to predict NLOGGI50 of drugs tested on the selected cell line. In each set, all the above-mentioned models were trained on the same set of samples and tested on a separate set of same samples.
 
Note that for the synthetic dataset and the NCI60 dataset, the predictors are a vector of real values (chemical descriptor values for NCI60) that are converted to images. For the GDSC dataset, we have two types of input features: chemical descriptors describing the drugs and gene expression describing the cell lines. We generated individual images for each feature type and used both of them as inputs to the CNNs. For the RF, SVR, ANN, and EN these two types of features were appended and used as the predictors. We trained REFINED CNN and 6 other competing models on same set of samples, where each sample is a combination of one drug tested on one cell line. All the models were tested on a separate set of same samples.

The distribution of NLOGGI50 shows a massive point mass at 4 (Figure \ref{Histogram} (a) and (b)) indicating that an overwhelming majority of drugs are not sensitive for majority of the NCI60 cell lines. Thus, we also considered a classification problem of whether a drug is sensitive or resistive among the NCI60 cell lines. Based on the NLOGGI50 distribution for different cell lines, the sensitivity threshold was empirically fixed at 4.5. 
All the drugs with NLOGGI50 less than the threshold (4.5) were considered resistive and the rest as sensitive. Similar to the regression scenario, we compared the CNN performance with RF, SVM, ANN, logistic regression (LR), Random CNN, and PCA CNN for the classification scenario. 

\subsubsection{Convolutional Neural Network}
\label{section:CNN_arch}
Convolutional Neural Networks (CNN) are designed to model multidimensional arrays, where convolutional layers along with pooling layer, are adaptive feature extractors connected to sequential fully connected layers \cite{angermueller2016deep}. A convolutional layer consists of multiple kernels connected to a local path of neurons in the previous layer, where all neurons share same parameters to generate a feature map. Thus, all neurons within the feature map scan same features in different locations of the previous layer. The pooling layer summarizes the feature map by finding the maximum/average of each adjacent kernel, which reduces the number of model parameter \cite{lecun2015deep}. We used two different CNN architectures, a sequential CNN for modeling the NCI60 and synthetic dataset, and a hybrid CNN, that can accommodate drug and genetic image for the GDSC datasets.

The sequential CNN regressor contains six learned layers: one input layer, two convolutional layer, two fully connected layer and one output layer. The CNN input dimension is same as the input image dimension, $ 26 \times 26$. The convolutional layer contains 64, $7 \times 7$ kernels convolving with valid border mode and stride of 2, followed by batch normalization, and ReLu activation. Each fully connected layer is followed by a batch normalization layer and ReLu activation layer. Number of neurons of the fully connected layers are respectively, 256, and 64. A dropout layer with retaining probability of 0.7 was added before the output layer. The above architecture remains same for Random CNN, PCA CNN and REFINED CNN.

The sequential CNN classifier contains input layer with the same size as the CNN regressor, three convolutional layer with 16 kernels of  size $7 \times 7$, 32 kernels of size  $7 \times 7$, and 64 kernels of size $3 \times 3$. Each convolutional layer is followed by a batch normalization \cite{ioffe2015batch} and a rectified linear unit (ReLu) \cite{glorot2011deep} activation function layer. The third ReLu layer is followed by two fully connected layers with 256 and 64 neurons respectively. Same as the CNN regressor, each fully connected layer is followed by a batch normalization, ReLu and a drop out layer. The CNN classifier architecture remains same for Random, PCA and REFINED CNN. We used adam optimizer to train both the CNN regressor and classifier. The CNN regressor and classifier architecture is shown in figure \ref{NCI_Arch} of the appendix.

We used hybrid CNNs with two inputs to model GDSC dataset, where two separate images are used as inputs to the two arms of the CNN. Each arm contains three convolutional layers, where the last convolutional layers are concatenated, and followed by two
sequential fully connected layers. The two input layers represent the cell lines and drug images, which defines each convolutional layer dimension. The three convolutional layer of each arm have 60 kernels with the size of $5 \times 5$ and stride of 1, 72 kernels with the size of $6 \times 6$ and stride of 2, and 72 kernels with the size of $5 \times 5$ respectively. The last convolutional layer of the arm that takes drug images as the input has stride of 1 and the other arm that takes cell images as the input has a stride of 2. Each convolutional layer is followed by batch normalization and ReLu activation function layers. The last two convolutional arms of the CNN are concatenated and connected to two sequential fully connected layers with 305 and 175 neurons. A batch normalization and a ReLu activation function is included after each fully connected layer. A dropout layer with retaining probability of 0.7 is placed before the output layer. The CNN model architecture is shown in figure  \ref{GDSC_Arch} of the appendix. The hybrid CNN was trained by an adam optimizer. The same architecture was utilized for Random, PCA and REFINED images. There are some variations in the architecture depending on the training set size that are explained in section \ref{section:gdsc_results}. All tuning parameters were chosen via a comprehensive search that is detailed in the next section.

\subsubsection{Hyper parameter search}

To tune the competitive models, we did a grid search on the following hyper parameters for
each model using hold-out for partitioning the data. We randomly partition both NCI60 and GDSC data to 80 \% training, 10 \% validation and 10 \% test. The same training, validation and test datasets were used for all models. The hyper parameters were tuned using the training and validation sets. The test sets were hold out for evaluating the final performance of each tuned model.

	\begin{itemize}
		\item RF: The number of decision trees in the forest ranged from 100 to 700 trees. For the maximum number of features for the best split evaluated at each node, we tried all the options provided by scikit-learn \cite{scikit-learn} including: number of features, square root of number of features, and logarithm base 2 of number of features.
		\item SVM: Gamma parameter of the radial basis function (RBF) kernel using both \textit{'scale'} and \textit{'auto'} options of scikit-learn, and the regularization parameter, \textit{'C'}, ranging from 0.01 to 100.
		\item EN and LR: Alpha penalty term and \textbf{L1} ratio values between (0.3,0.7) and (1e-6,1e-4), respectively.
		\item ANN: Number of hidden layers (3-6), number of neurons per each hidden layer and learning rate of the Adam optimizer.  
		\begin{itemize}
			\item 1 hidden layer: 800-1200 neurons
			\item 2 hidden layer: 600-1000 neurons
			\item 3 hidden layer: 400-700 neurons
			\item 4 hidden layer: 200-500 neurons
			\item 5 hidden layer: 50-250 neurons
			\item 6 hidden layer: 20-80 neurons
			\item learning rate: 1e-6 - 1e-3
		\end{itemize}
		\item CNN: As we did not have access to GPUs, we did not do comprehensive hyper parameter grid search for the CNNs. The current parameters were chosen over about hundreds of run. The hyper parameter space that we searched was number of convolutional layers, number of dense (fully connected) layers, and the learning rate of the Adam optimizer. In each convolutional layer, we seek optimum number of kernels, kernel size, and stride. Per each dense layer, we checked multiple number of neurons. The insights that we gained from thousands of runs are as follows:
		\begin{itemize}
			
			\item We observed that $7 \times 7$ kernels were better than smaller kernels of 3x3 for the convolutional operator. For larger number of features, larger kernels might be desirable.
			
			\item Using strides larger than 1 is more effective in embedded dimensionality reduction than the pooling layer.

			\item Using two or more sequential convolution layers with kernel size of $7 \times 7 $ and stride of 2 reduces the feature map dimension considerably compared to the input image. Therefore, large number of kernels is recommended at least for the last convolution layer, to provide sufficient number of extracted features for the dense layer.
			\item The width of network is as important as the depth of the network.
			\item The Adam optimizer is recommended.
		\end{itemize}
	\end{itemize}

For unbiased evaluation of machine learning models, nested cross-validation \cite{cawley2010over} (CV) is often considered where an inner CV is used for model selection (hyper parameter selection) and an outer CV is used for evaluating the model tuned by the inner CV. However, nested CV is often extremely computationally intensive and thus we considered a training-validation-test (hold-out) approach where the hyper-parameters are tested on the validation set and the selected model is evaluated based on the separate test set. We used training-validation-test approach as the sample size is relatively large and thus, both training-validation-test and nested CV approaches are expected to provide similar results for comparing different modeling approaches. To illustrate the similar behavior, we compared the results of nested CV and training-validation-test (hold-out) approach using three cell lines randomly selected from the NCI60 dataset. As the results provided in figure \ref{HyperPar} indicates, the difference in the results are minimal. On the other hand, to compare the time complexity of the two approaches, we selected one cell line with a fixed CNN architecture. Within the CNN architecture, a grid of hyper parameters was defined. In a single run (architecture) that takes 48 hours on Texas Tech University high performance computer center, 50 different models can be tried using training-validation-test, whereas only 4 models can be tried using nested cross-validation. Given the time complexity and similar performance of the two approaches, we used training-validation-test approach for hyper parameter selection and model evaluation as we could search considerably larger space of hyper parameters.

\section{Results}

In this section, we report the performance of our REFINED-CNN methodology on the previously described synthetic, NCI60, and GDSC datasets. In each case, the performance of the REFINED CNN was compared to ANN, RF and SVR models. We also compared the REFINED CNN with EN, Random CNN, and PCA CNN for biological datasets. 

\noindent \emph{Evaluation Metric}:
We evaluated the performance of each regression model using (a) normalized root mean square error (NRMSE), (b) Pearson correlation coefficient (PCC) between the predicted and target values and (c) bias reduction. 
The NRMSE (\ref{NRMSE}) is the ratio of the root mean squared error (RMSE) of a given model to the RMSE with mean as the predictor. It represents the overall  potential of the model to minimize prediction error.
\begin{equation} \label{NRMSE}
NRMSE = \sqrt{\frac{\sum_{i=1}^{N}(y_{i}-\hat{y_{i}})^2}{\sum_{i=1}^{N}(y_{i}-\bar{y})^2}}
\end{equation}
where the $y$, $\bar{y}$ , and $\hat{y}$ are respectively the target data, mean of the target values, and predicted target values.
We offer NRMSE in order to implicitly compare all the models with respect to the baseline intercept-only model.

PCC indicates collinearity between the predicted and observed responses. Lack of collinearity often implies model misspecification and lack of predictive capability.

To represent the bias, we first generated the scatter plot of the residual (ordinate) and the observed response (abscissa).  We captured the bias via  the angle ($\theta$) between the best fitted line through the residuals and the abscissa. An unbiased model is expected to produce an angle of $0^{\circ}$. Therefore, a smaller value of $\theta$ indicates that the model is less biased.

We used Gap statistics \cite{tibshirani2001estimating} to report the significance of the difference in performance across methods of the regression tasks. We paired each model with a null model \cite{Costello:2014} with bootsrap sampling, where the bootstrap sampling was done on the dose-response values of the test set along with their corresponding predicted values for each model and the null model randomly predicted dose-response values for the sampled test set using the distribution of the training set dose-responses. The process is repeated for 10,000 times and a distribution of NRMSE, PCC, and Bias, is made for each model along with the null model. The distribution per each metric for each model is paired with null models' metric distribution. Then cluster centroids are calculated using K-means (k = 2) clustering, after the Gap statistics shows appropriateness of using 2 clusters. The difference between each model and the null model cluster centroids per metric represents the difference between them. In addition to the Gap statistics, all models were subjected to a robustness analysis \cite{Costello:2014}, where we calculate how many times the REFINED CNN outperforms other competing models in 10,000 repetition of bootstrap sampling process.

We calculated 95 \% confidence interval for each metric that is used to measure the performance of the methodologies in modeling NCI60 and GDSC dataset using a psuedo Jackknife-After-Bootstrap confidence interval generation approach \cite{rahman2017integratedmrf,efron1992bootstrap}. Multiple Bootstrap sets were selected from the test samples and error metrics generated resulting in a distribution for each error metric which was used to calculate the confidence interval for a given cell line for NCI60 dataset or a given pair of cell line and drug for the GDSC dataset.

The classification models that predict the sensitive and resistive drugs applied on the NCI60 data cell lines, were evaluated using accuracy, precision, recall, F1 score, and area under the receiver operating characteristic curve (AUROC) metrics. Accuracy is the ratio between correct predictions; true positive (TP) and true negative (TN), over all the predictions; summation of TP, TN, false positive (FP), and false negative (FN).  

\begin{equation} \label{accuracy}
Accuracy = \frac{TP + TN}{TP + TN + FP + FN}
\end{equation}

Precision is the ability of predicting positive instances of the classifiers, or in other words, the ratio of correctly predicted positive instances.
\begin{equation} \label{precision}
Precision = \frac{TP}{TP + FP}
\end{equation}

Recall or true positive rate (TPR) is the ability of the classifiers in predicting positive instances, which corresponds to the proportion of of positive instances that are correctly predicted as positive.

\begin{equation} \label{recall}
Recall = \frac{TP}{TP + FN}
\end{equation}

F1 score is simply harmonics mean of the precision and recall, 
\begin{equation} \label{F1-score}
F1 score = \frac{2TP}{2TP + FP + FN}
\end{equation}

False positive rate (FPR) is a ratio between negative instances that are mistakenly predicted as positive instances. AUROC combines TPR and FPR in many different threshold for each classifier on a single curve, where the area under the curve is considered as AUROC.

\begin{equation} \label{FPR}
FPR = \frac{FP}{FP + TN}
\end{equation}

The 95 \% confidence interval of each of the above-mentioned classification metrics was calculated using the Binomial proportion confidence interval \cite{witten2005practical,vollset1993confidence}.

\subsection{Comparison with other image generation methods}
\label{section:comparison}

In this section, we consider two alternative image generation methods for comparison purposes: random projection-based and PCA-based. In the random projection method, we assume each image is a matrix and the location of each entry in the vector is randomly mapped to a location in the matrix. Thus, we placed each element of the drug descriptors or the gene expression on the image (matrix) coordinates one after another. 

Principal component analysis (PCA) \cite{wold1987principal} is mainly used for dimensionality reduction and visualization purposes, where each sample could be represented on a 2D plane aligned with their first two principal eigen vectors. The first two principal components of the transposed covariance matrix with rows being features and columns representing samples, were selected as the feature coordinates. Some of the generated images using the random and PCA methods are shown in figure \ref{IM_Gen}.

\subsection{Synthetic Data}

\begin{figure}[!htbp]
	\centering
	\includegraphics[width=\textwidth]{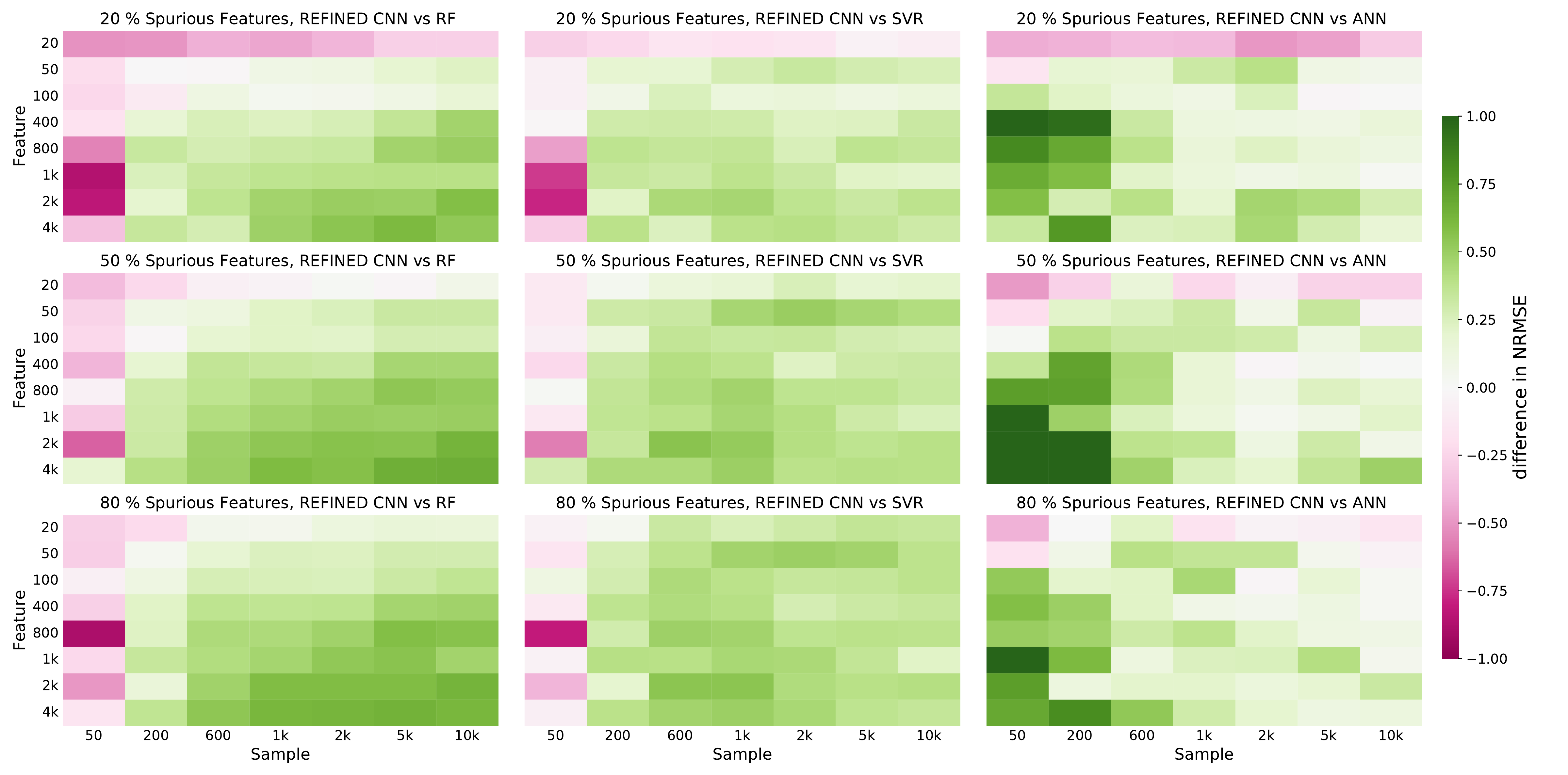} 
	\caption{NRMSE difference of REFINED CNN with RF, SVR and ANN for different sample sizes and with different number of features where the 20 \%, 50 \% and 80 \% of the features are respectively spurious. The green regions of each heatmap represents where REFINED CNN outperform other models.} 
	\label{Sim100} 
\end{figure}

In this section, we offer the comparative performance of the candidate models on the simulated dataset. First we generate a lexicographic ordering of $P$ features. The features are then generated from a $0$ mean Gaussian process with covariance depending only on the lexicographic distance between the features. In each case, a subset of features were randomly selected as spurious. Subsequently, random weights generated for non-spurious features were used to generate the target values. The generated target values were normalized between 0-1.
We have used REFINED to generate images for different $N/P$ scenarios and then trained a CNN for each scenario.
In each case, the same dataset was used to train RF, SVR and ANN for comparison. We used five-fold cross validation for all the models on fixed training, validation and test sets. The results are summarized in figure \ref{Sim100}, as heat maps. As shown in figure \ref{Sim100}, the green regions represents the cases where the REFINED-CNN NRMSE is less than the competing models. The separate heatmap for all the model results are provided in the figures \ref{SVRHeat},\ref{RFHeat},\ref{ANNHeat},\ref{CNNHeat} of the appendix. The heat maps clearly shows, that the REFINED-CNN methodology outperforms others when the number of features and samples are relatively high ($P > 100, N > 600$) regardless of the percentage of spurious features present in the dataset. We also observe that the performance of the posited methodology improves as the $N$ increases.  \cite{chen2011stringing} also reported that the performance of their UDS based projection improved with increase in both $N$ and $P$. Therefore, our findings suggest that our second order REFINED approximation are in agreement with the first order {\it stringing} approximation of \cite{chen2011stringing}. Furthermore, we also observe that as the ratio of spurious features increases, the predictive performance of our REFINED-CNN also improves as compared to the competing models. Recall, we are not performing any feature pre-selection for REFINED-CNN for the synthetic data. This exercise demonstrates the ability of our approach to automatically remove spurious features without performing an explicit feature selection {\it a-priori}. Additionally, by comparing the REFINED CNN versus ANN heatmap we observe that increasing the number of samples reduces the gap between their performance as more samples are available to train the large number of parameters of ANN.

We next investigated the effect of the REFINED-CNN approach on the bias characteristics of the prediction. Figure \ref{Simulation} shows the scatter plot of prediction versus actual responses of the four models when 80 \% and 20 \% of the features are spurious. Clearly, the scatter plot for REFINED-CNN closely follows a straight line with unit slope indicating predictive accuracy of our approach. RF and SVR reveal their  well-known tendency  to under-predict higher valued observation and over-predict lower valued observations \cite{matlock2018investigation}. REFINED-CNN bias is also better than the bias observed for the ANN scenario. Thus, it appears that the REFINED-CNN approach can automatically improve the prediction bias which some of the other existing models are known to suffer from.

\subsection{NCI60 dataset}

\subsubsection{Classification}

\begin{figure}[!htbp]
	\centering
	\includegraphics[width=\textwidth]{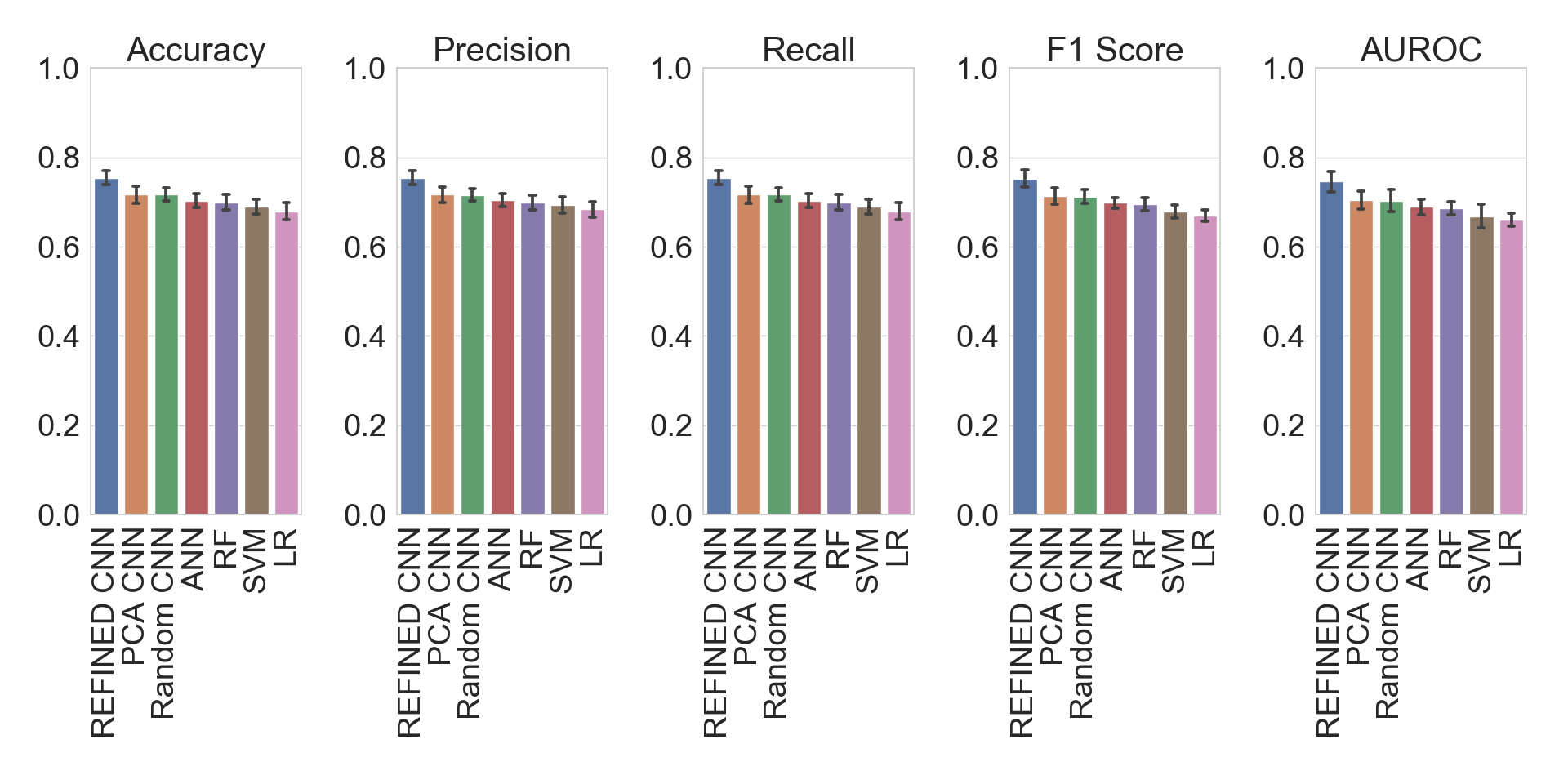} 
	\caption{Summary of REFINED-CNN and 6 other competing classifier models performance on randomly selected cell lines of NCI60 database, using the accuracy, precision, recall, f1-score, and AUROC metrics.}
\label{NCI Classification results} 
\end{figure}

We investigated the discriminative power of REFINED-CNN as compared to other models in predicting resistant and non-resistant drugs on different cell lines of the NCI60 dataset.
The threshold for defining resistant and non-resistant classes, was selected based on the drug response distribution shown in figure \ref{Histogram}. The drugs with NLOGGI50 smaller than $4.25$ was considered resistant and the rest as non-resistant.
Since, we have sufficiently large number of drugs for each cell line \footnote{Each unique drug for a particular cell line is a sample in this scenario}, we randomly considered 80\% of the drugs for training, 10\% for validation and 10\% for testing. As shown in the figure \ref{NCI Classification results}, the REFINED CNN outperforms other classifiers for all 17 cell lines. The average classification accuracy of REFINED-CNN was 75.4\% - considerably higher than the average classification accruacy obtained for Random CNN (71.6\%), PCA CNN (71.7\%), ANN (70.3\%), RF (70\%), SVM (69\%) and LR(67.9\%). For each model, we also report precision, recall, f1-score, and AUROC. REFINED-CNN outperforms other models considering all the metrics. Detailed classification results are provided in appendix tables \ref{Classification}, and \ref{Classification2}. The 95 \% intervals for each metric for different cell lines are provided in table \ref{Classification_inervals} of appendix.

To compare the statistical significance of the difference in performance between competing classifiers and REFINED CNN, we used the McNemar’s test which is a paired nonparametric statistical hypothesis test. McNemar’s test evaluate whether two models disagree in the same way or not. To compare classifiers with REFINED CNN classifier pairwise, a contingency table is formed and McNemar’s test is applied \cite{dietterich1998approximate}. The null hypothesis is whether two classifiers disagree by the same amount. Therefore, if the p-value is smaller than a threshold (0.05), null hypothesis is rejected, and the conclusion is: there is a significant disagreement between the two classifiers. The results of comparing REFINED CNN with other models using McNemar’s test is provided in table \ref{Classification_statistical}  of the appendix.  We observe that the REFINED CNN classifier performance is significantly different than the other classifiers (LR, RF, SVM, ANN, Random CNN, PCA CNN).

\subsubsection{Regression}

The NCI60 dataset was randomly partitioned into 80\% , 10\% , and 10\% segments for training, validation and test purpose, respectively. The same training, validation and test set were used for model comparisons. The performance of each model was evaluated using normalized root mean square error (NRMSE), Pearson correlation coefficient (PCC), and bias. 

\begin{figure}[!htbp]
	\centering
	\includegraphics[width=\textwidth]{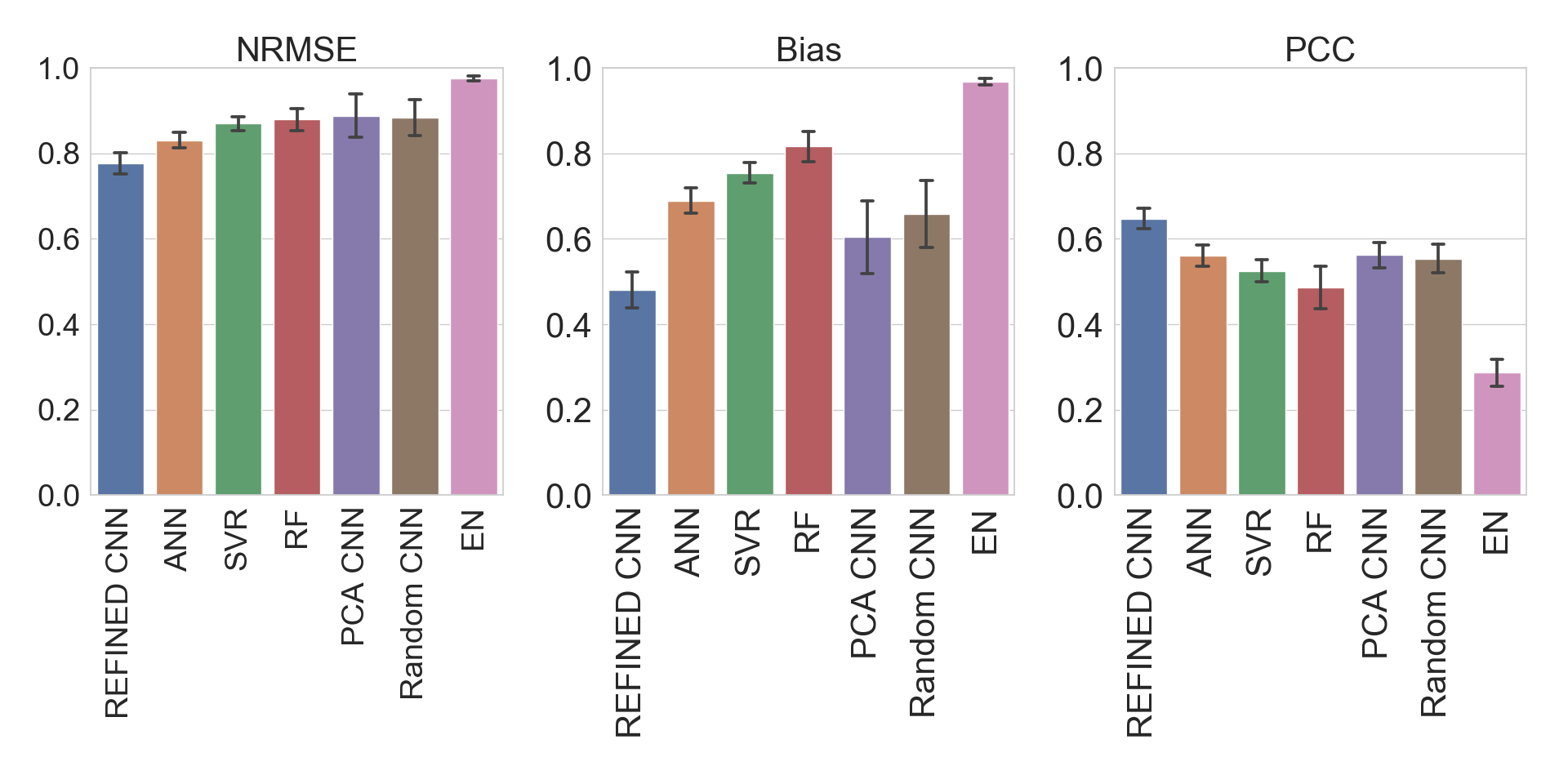} 
	\caption{Summary of REFINED-CNN and 6 other competing regressors models performance on randomly selected cell lines of NCI60 database, using the NRMSE and Pearson correlation coefficients metric.}
	\label{NCI Regression results}  
\end{figure}

Table \ref{Reression} in the appendix details the performance of each model with respect to the foregoing metrics for different cell lines. Table \ref{Reression} is summarized in figure \ref{NCI Regression results}, as bar plots. The 95 \% confidence interval for all the models per each cell line is provided in the figure \ref{NCI_NRMSE_Conf} \ref{NCI_PCC_Conf} \ref{NCI_Bias_Conf} of the appendix.
We note that CNN outperforms all the competing models in all 17 cell lines. The average improvement in NRMSE, PCC and bias for REFINED-CNN as compared to other competing models are 6-20\%, 8-36\%, and 12-38\%, respectively . 

We used Gap statistics and robustness analysis as described in the \textit{evaluation metric} section to compare REFINED CNN with other competing models for the regression task. The robustness results are provided in table 11 of the appendix which indicates that REFINED CNN has better performance in terms of: NRMSE between 89.51-100\% of the times, PCC between 94.42-100\% of the times, and Bias 91.14-100\% of the times on average as compared to other models. The Gap statistics results are provided in table \ref{NCI_Gap} of the appendix, which indicates superior performance of REFINED CNN as compared to the other models in all the metrics both in average and per each cell line. The NRMSE and PCC distribution of all the seven models along with the null model are plotted for three cell lines of the NCI60 dataset in figure \ref{GapDist} of the appendix.

\subsubsection{Data Augmentation}

This section analyzes the effect of augmenting the dataset using samples from the less represented regions.  As shown in figure \ref{Histogram}(a), the massive point mass associated with the non-sensitive (resistant) drugs severely impacts a global regression model for NLOGGI50. This problem is analogous to zero-inflation problem in classical statistical literature. In such situation, the discrete point mass is modeled separately from the continuous part (see \cite{ghosh2012k, ghosh2006bayesian} and references therein). In our situation, it boils down to classification into sensitive/resistive category followed by a regression in the sensitive category. 

We have already demonstrated superiority of REFINED-CNN in both classification and regression in Tables \ref{Classification} and \ref{Reression}, respectively.  In this section, we explore if REFINED-CNN's performance could be improved by synthetically oversampling the sensitive category to arrive at a more balanced dataset \cite{chawla2002smote}. To that end, we used a version of SMOTE technique and generated bootstrap replicates from the sensitive category. The NRMSE and NMAE (Normalized Mean Absolute Error) improvement of REFINED-CNN regression model on different cell lines is illustrated in table \ref{Regression_BootStrap}. The bootstrap data augmentation systematically decreases the NRMSE and NMAE for the cell lines indicating the negative impact of the point mass in the response distribution.

\begin{table}[!htbp]\caption{REFINED CNN NRMSE and NMAE improvement by data augmentation of the sensitive drug region using bootstrap sampling. }\label{Regression_BootStrap}
	\centering
	\begin{tabular}{@{}c||ccc||ccc@{}}
		
		& \multicolumn{3}{c||}{No data augmentatoin} & \multicolumn{3}{c}{Bootstrap} \\ 
		\hline
		\hline
		Cell Lines     & NRMSE  & NMAE  & \# Samples   		& NRMSE		& NMAE       & \#Samples     \\
		\hline
		\hline
		SNB\_78       & 0.784    & 0.713      & 13940        & 0.744  & 0.688    & 19613       \\
		MDA\_MB\_435  & 0.787    & 0.739      & 36868        & 0.762  & 0.729  	 & 59570        \\ 
		NCI\_ADR\_RES & 0.798    & 0.745      & 37156        & 0.755  & 0.717 	 & 59250        \\
		786\_0		  &	0.752	 & 0.688	  &	49344		 & 0.713  & 0.663 	 & 76908		 \\
		COLO\_205	  &	0.741	 & 0.664	  & 48946		 & 0.722  & 0.660 	 & 75158		 \\

	\end{tabular}
\end{table}

\subsubsection{Sample Size Analysis}
 
\begin{figure}[!htbp]
	\centering
	\includegraphics[width=\textwidth]{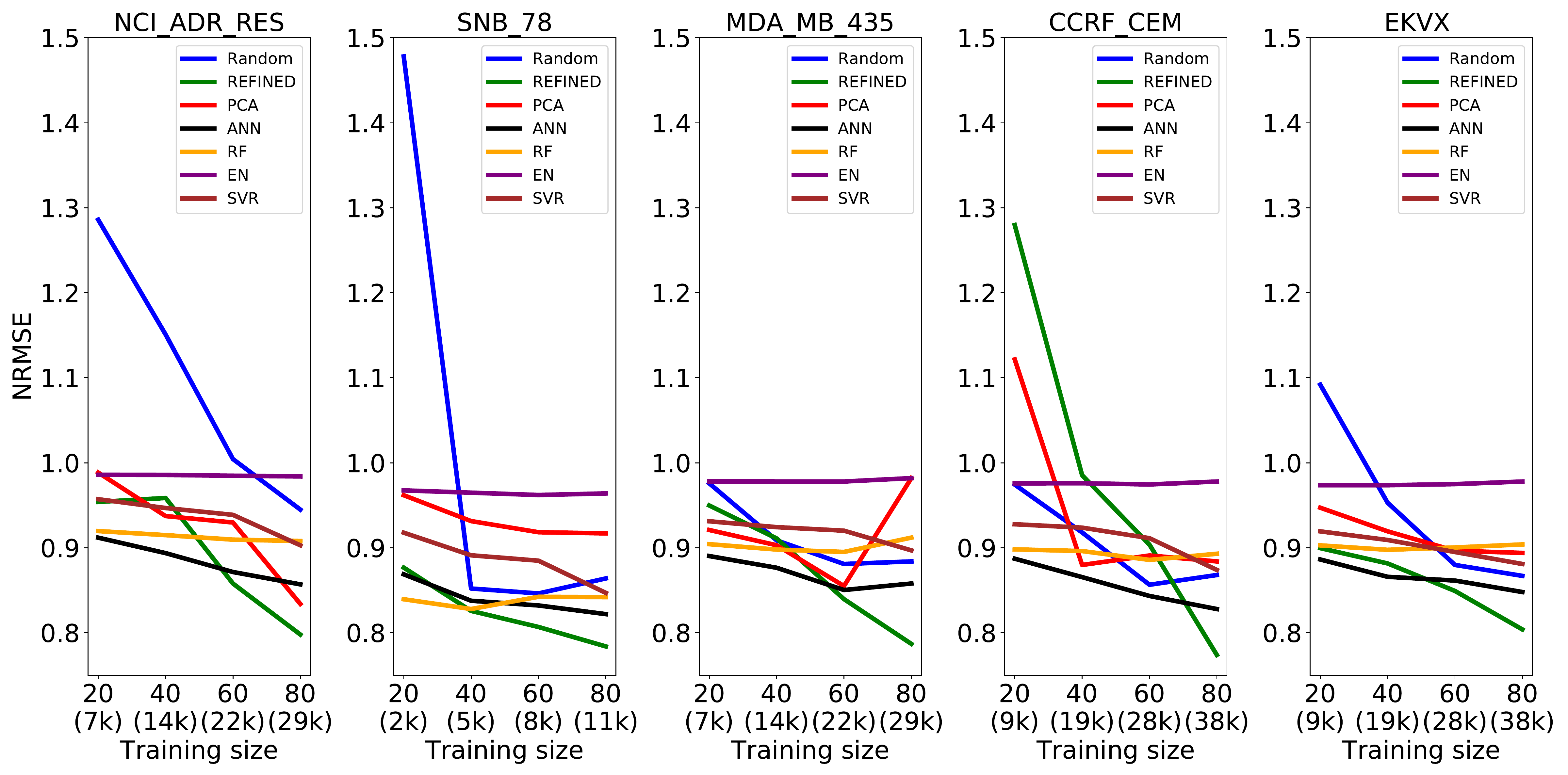} 
	\caption{Comparison of six competing models with REFINED CNN for different training sizes applied to five randomly selected cell lines of NCI60 dataset. The x-axis represents the percentage of data used for training of each cell line along with the actual number of samples used in brackets. To fit the figures, we shortened REFINED CNN as REFINED, Random CNN as Random, and PCA CNN as CNN in the legends.}
	\label{NCI_Samplesize}  
\end{figure}

Deep CNN models are expected to perform better with larger number of samples as compared to smaller number of samples. Therefore, we trained our model on different portion of training sets for randomly selected cell lines to test this hypothesis. We trained our model on 20 \%, 40 \%, 60 \% and 80 \% of the available drugs applied on the selected cell lines and
kept rest of the data for testing, considering NRMSE as a comparison metric. The results of five cell lines are summarized in figure \ref{NCI_Samplesize} which illustrates that REFINED-CNN outperforms the other models as sample size increases. This trend was also observed on the synthetic data (figure figure \ref{Sim100}).

\subsubsection{Model Stacking}

\begin{figure}[!htbp]
	\centering
	\includegraphics[width=\textwidth]{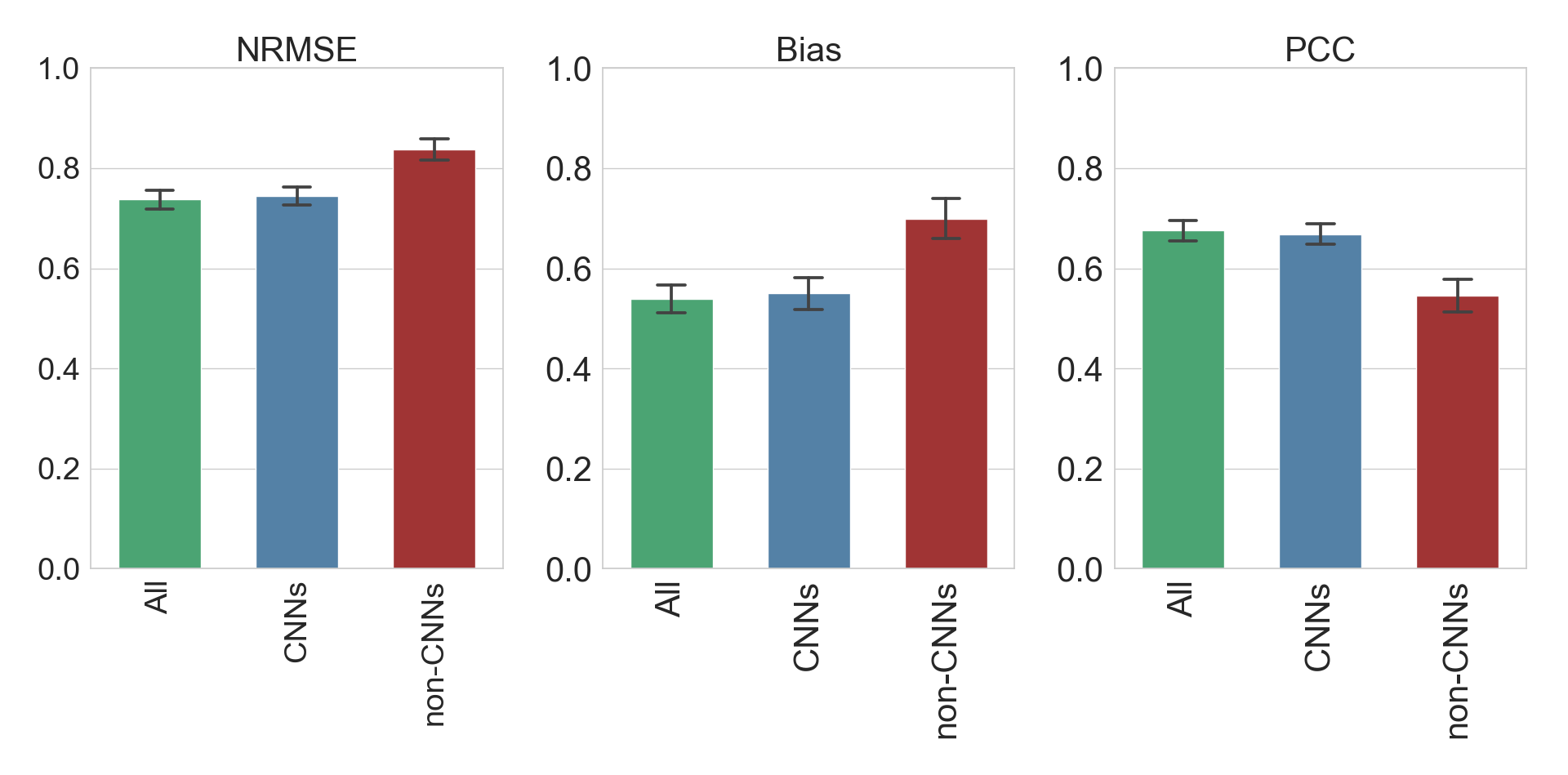} 
	\caption{Comparison of stacking all models, all CNN models and non-CNN models using NRMSE, bias and Pearson correlation coefficients (PCC).}
	\label{Stacking}  
\end{figure}

To explore whether stacking of multiple models can improve prediction performance, we stacked predicted drug sensitivity of the validation set of all models--in three different combinations--as the covariates in a linear regression model to find the weight of each model which are then employed to predict drug sensitivity of the test set. 
These three combinations are stacking non-CNN models (RF, SVR, ANN, and EN); stacking CNN models (PCA CNN, Random CNN, and REFINED CNN); and stacking all models (REFINED CNN and 6 other competitors). Figure \ref{Stacking} represents the stacking results where the average NRMSE of stacking all models is 0.738, stacking CNNs is 0.744 and stacking non-CNNs 0.837. By comparing the stacking results, with average NRMSE of each model in table \ref{NCI Regression results} , the stacking produced a significant improvement as compared to non-CNN models individually. It is notable to mention that REFINED CNN average NRMSE in table \ref{NCI Regression results} is significantly lower than stacked of non-CNN models.

\subsubsection{Bias Analysis}

\begin{figure}[!htbp]
	\centering
	\includegraphics[width=\textwidth]{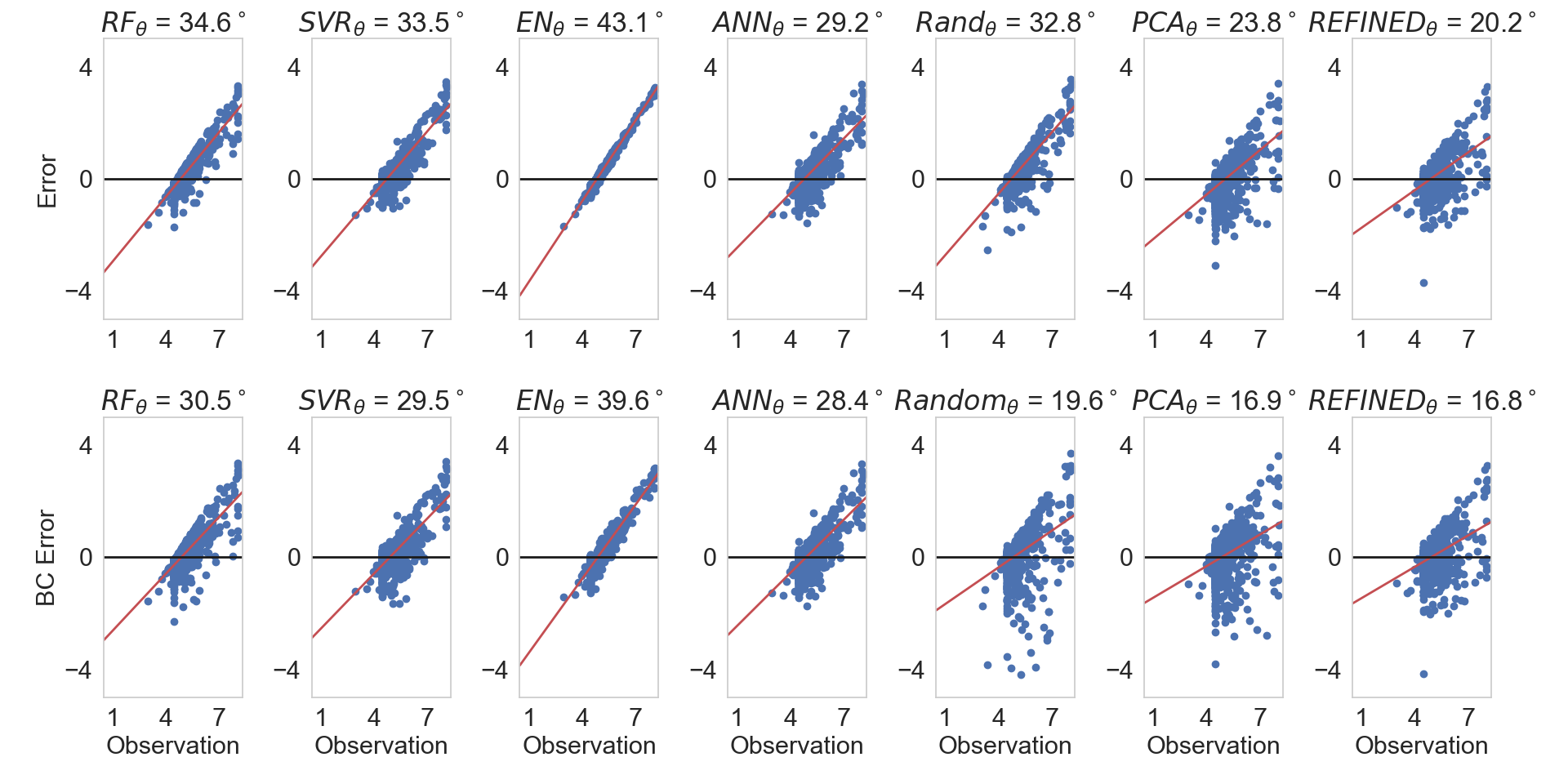} 
	\caption{Residual plots for RF, SVR, EN, ANN, Random CNN, PCA CNN, and REFINED CNN for $SNB\_78$ cell line test set data prior bias correction (BC) (a) and post bias correction (b). The corresponding NRMSE for scenario (a) are RF = 0.842, SVR = 0.847, EN = 0.964, ANN = 0.822 , Random CNN = 0.864 , PCA CNN = 0.917, REFINED CNN = 0.784 and for scenario (b) are RF = 0.830, SVR = 0.836, EN = 0.937, ANN = 0.817 , Random CNN = 0.832 , PCA CNN = 0.840, REFINED CNN = 0.763.}
	\label{Error Analysis}  
\end{figure}

In this section, we re-investigate the effect of the REFINED-CNN approach on the bias characteristics of the prediction using actual biological data. Figure \ref{Error Analysis} shows the plot of the residuals against the observed values.  Similar to the earlier presented synthetic data scenario, Figure \ref{Error Analysis} (a) shows that REFINED-CNN has the lowest bias ($20.2^{\circ}$) as compared to Random-CNN ($32.8^{\circ}$), PCA-CNN ($23.8^{\circ}$), ANN ($29.2^{\circ}$), RF ($34.6^{\circ}$), SVR ($33.5^{\circ}$) and EN ($43.1^{\circ}$). To investigate whether bias correction erodes the advantage of REFINED-CNN in terms of bias, we considered BC1 bias correction algorithm proposed in \cite{zhang2012bias} where we fit a second model (linear regression was used in this case) on the residuals. The results shown in Figure \ref{Error Analysis} (b) illustrate the superiority of  REFINED-CNN in terms of lowering bias even after bias correction is applied to the competing models.

\subsection{GDSC dataset}
\label{section:gdsc_results}

In this section, we consider the application of REFINED CNN in integrating two types of heterogeneous datasets. Our predictors now consist of (a) PaDel chemical descriptors representing the drugs and (b) gene expression profiles for each cell line. The response consists of the experimentally obtained IC50 for each drug-cell line pair. We used the REFINED approach to generate the images corresponding to the gene expressions for each cell line and drug descriptors for each drug compound in the GDSC dataset.

\begin{figure}[!htbp]
	\centering
	\includegraphics[width=\textwidth]{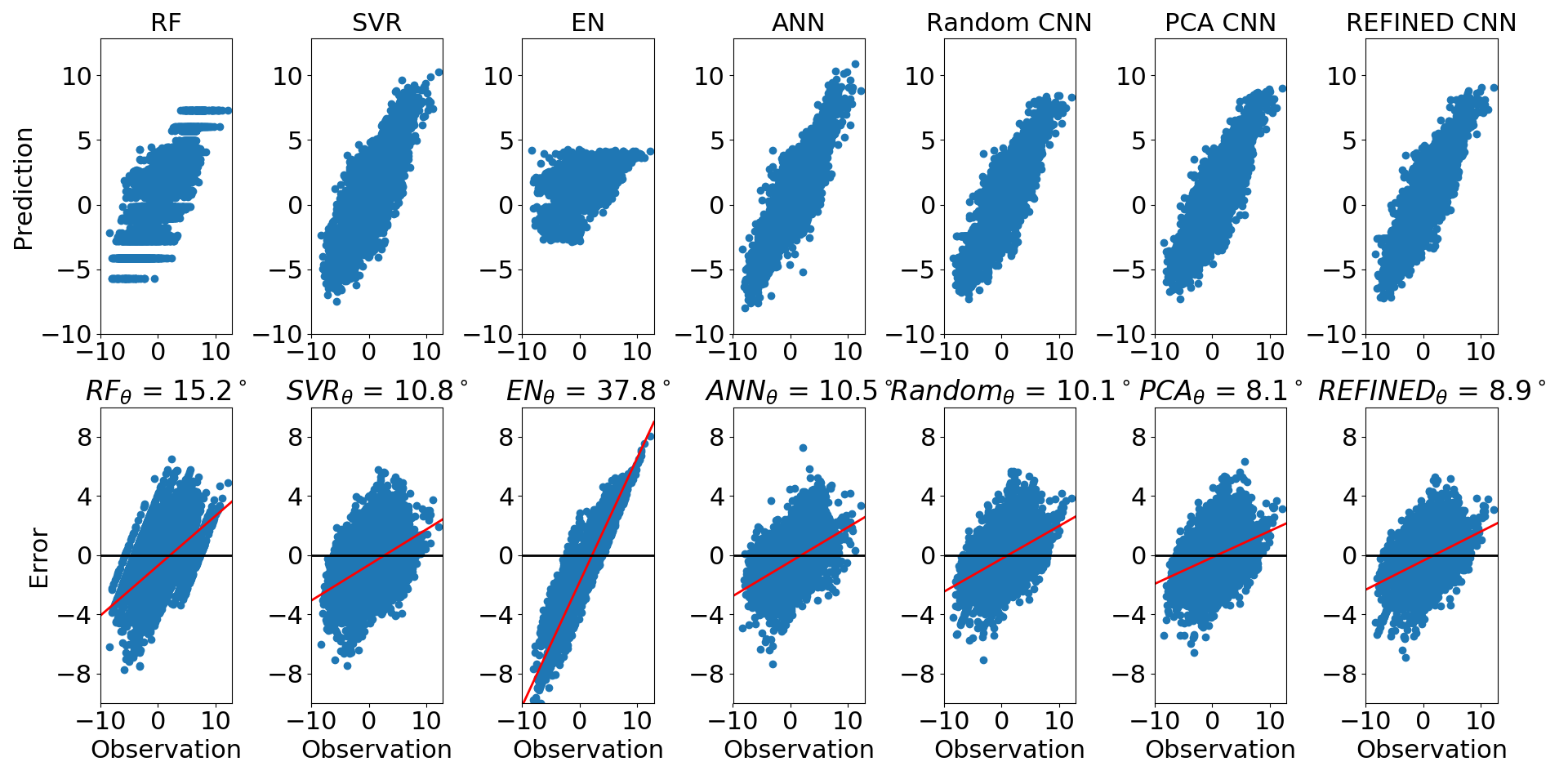} 
	\caption{Scatter plot of predicted NLOGIC50s and their residual (error) for each model, in the case of models were trained on 80 \% of the available data. In each scatter plot, observation is the NLOGIC50 values if the test set. On the bottom row, bias of each model is depicted by the method explained in the bias analysis section. }
	\label{GDSC_results}  
\end{figure}

Considering 222 drugs and around 972 cell lines for each drug, the total number of samples in the dataset is close to 177K. We randomly divided the dataset into 80\% training, 10\% validation and 10\% test sets, where each set covariates contains 1211 genes and 992 chemical drug descriptors. Figure \ref{GDSC_results} represents the scatter plot of the natural log IC50 prediction using our REFINED CNN approach, Random CNN, PCA CNN, ANN, RF, SVR, and EN along with their corresponding residual plots. Table \ref{GDSC_Table} summarizes the performance of each model using NRMSE, PCC, and bias.

Similar to NCI60 data analysis, we used Gap statistics and robustness analysis for further comparison of the models. The robustness analysis and Gap statistics results are provided in tables \ref{GDSC_Robust} and \ref{GapGDSC} of the appendix respectively. The Gap statistics distribution plots per each metric of each model paired with the null model along with their corresponding cluster centroids are provided in figures \ref{GapGDSCNRMSE}-\ref{GapGDSCBias} of the appendix.

As shown in table \ref{GDSC_Table}, REFINED-CNN model achieves improvement as compared to other models in the range of  1-47\% for NRMSE, 1-42\%, for PCC. We also train the REFINED CNN on 50\% and 20\% of the GDSC data and compared it with other competing models. As the CNN architecture (please see figure \ref{GDSC_Arch} of the appendix) trained on the 80\% of the REFINED images was too complex to be trained on the 20\% of the data, we reduced the network complexity of REFINED CNN and ANN models. We removed the last convolutional layer and its latter dense layer for the CNN model, and also we removed the first two dense layer of the ANN model. The detailed corresponding results are provided in table \ref{GDSC_results}. The predicted and residual scatter plots are provided in figures \ref{GSC 0.5}, \ref{GSC 0.2} of the appendix.

\begin{table}[!htbp]\caption{Comparison of REFINED CNN, Random CNN, PCA CNN, ANN, RF, SVR, and EN peformance on the GDSC dataset in terms of NRMSE, Pearson Correlation Coefficients (PCC), and bias}\label{GDSC_Table}
	\centering
	\resizebox{\textwidth}{!}{%
		\begin{tabular}{c||ccc||ccc||ccc}
			\multirow{2}{*}{Models} & \multicolumn{3}{c||}{Trained on 20 \%} & \multicolumn{3}{c||}{Trained on 50 \%} & \multicolumn{3}{c}{Trained on 80 \%} \\
			& NRMSE   & PCC   & Bias  & NRMSE  & PCC   & Bias  & NRMSE  & PCC  & Bias      \\
			\hline
			\hline
			EN        & 0.890   & 0.488  &  0.848 & 0.889  & 0.484  & 0.849  & 0.887  & 0.486 &  0.840  \\
			RF        & 0.609   & 0.797  &  0.433 & 0.620  & 0.785  & 0.417 & 0.569  & 0.821 &  0.337 \\
			SVR       & 0.750   & 0.847  &  0.257 & 0.742  & 0.845  & 0.273  & 0.525  & 0.853 &  0.241  \\
			ANN       & 1.407   & 0.519  &  0.784 & 0.475  & 0.883  & \textbf{0.153}  & 0.435  & 0.901 &  0.233 \\
			Random CNN  & 0.579 & 0.836 & 0.215   & 0.456  & 0.892  &  0.193  & 0.441  & 0.903  & 0.222  \\
			PCA CNN   & 0.612   & 0.820  & \textbf{0.201}  & 0.461  & 0.891 & 0.228  & 0.443   & 0.901 &  0.179 \\
			REFINED CNN & \textbf{0.541}  & \textbf{0.845} &  0.255  & \textbf{0.439}   & \textbf{0.899} & 0.173 & \textbf{0.414}  & \textbf{0.911}  & \textbf{0.197}
		\end{tabular}%
	}
\end{table}

We also incorporated larger number of genes to experiment the impact of creating larger images using REFINED. We selected larger common subset of genes including (2147 $\sim$ 2000, 2985 $\sim$ 3000, 4271 $\sim$ 4000, 8048 $\sim$ 8000) genes and created the REFINED images to train the same CNN architecture that we used to train the GDSC data with 1211 selected genes. We also trained other competing models with the same hyper parameters. In our experiment, the SVR model crashed after several hours due to the larger memory requirement while estimating the kernel by computing the distance function between each point in the dataset. Thus, we used a bag of SVRs, where we selected subset of 400 features (appended gene expression and drug descriptors) to train 100 SVRs in parallel. The complete results are provided in figure \ref{GDSC_gene_figure} and table \ref{GDSC_gene_table} along with the corresponding confidence intervals in figure \ref{GDSC_NRMSE_Conf} \ref{GDSC_PCC_Conf} \ref{GDSC_Bias_Conf} of the appendix. We observed that the trend of REFINED CNN outperforming other approaches is maintain irrespective of the number of genes used for training models.

\subsubsection{Comparison with state-of-the-art methods}

We have also compared REFINED-CNN with another state-of-the-art approach, Deep-Resp-Fores (DRF) proposed by Su et al. \cite{su2019deep}. The DRF is a deep cascaded forest model, designed to classify the anti-cancer drug response as “sensitive” or “resistive” based on integration of heterogeneous input data (gene expression and copy number alteration (CNA) of GDSC data. We switched the random forest classifier of the DRF to a regressor for prediction and used drug descriptors in the DRF's wing that takes CNAs as the input for training the DRF. 

We have also compared REFINED-CNN with heterogeneous graph neural networks (HGNN) proposed by Lim et al. \cite{lim2019predicting}. HGNNs automate feature engineering task by aggregating feature information of neighborhood nodes, where the input data of the network are from different sources \cite{zhang2019heterogeneous}. Lim's network embeds the 3D structural information of protein-ligand complexes in distance matrix, to predict drug-target interaction. We trained their network by encompassing gene-drug information in distance matrix, to predict drug sensitivity of the GDSC dataset. The results of applying DRF\cite{su2019deep} and HGNN \cite{lim2019predicting}  are provided in table \ref{state_art} for comparing with REFINED CNN. As the results indicate, REFINED CNN outperforms DRF and HGNN for drug response prediction of the GDSC dataset.
The robustness analysis and Gap statistics results for both DRF and HGNN methods are provided in the table \ref{GDSC_Robust} and \ref{GapGDSC} of the appendix, respectively.

\begin{table}[!htbp]
	\centering
	\caption{Comparison REFINED-CNN with DRF and HGNN for GDSC drug sensitivity prediction. }
	\label{state_art}
	\begin{tabular}{c||ccc}
		Models      & NRMSE          & PCC            & Bias           \\
		\hline
		\hline
		REFINED-CNN & \textbf{0.414} & \textbf{0.911} & \textbf{0.197} \\
		DRF \cite{su2019deep}             &  0.986  & 0.169   & 0.976     \\
		HGNN \cite{lim2019predicting}     &  0.637  & 0.805   & 0.446                  
	\end{tabular}
\end{table}

\subsection{Ablation study}

The main components of REFINED are dimensionality reduction followed by a search optimization algorithm, where we use Bayesian MDS as a global distortion minimizer and hill climbing as local adjustment to reach a locally optimal configuration among multiple automorphs. To investigate the contribution of each individual component to REFINED, evaluated the effect of different approaches in each step. For the dimensionality reduction step, we considered other dimensionality reduction techniques such as Isomap \cite{tenenbaum2000global}, Linear Local Embedding (LLE) \cite{roweis2000nonlinear}, and Laplacian Eigenmaps (LE) \cite{belkin2003laplacian} instead of Bayesian MDS. Isomap generalizes MDS with using Geodesic distance rather than Euclidean distance in nonlinear manifolds, where the Geodesic distance is approximated by summation of Euclidean distances. LLE is a local dimensionality reduction technique, which tries to reconstruct each sample based on k-nearest samples in a lower dimension locally. Laplacian eigenmap (LE) is similar to LLE except, it uses Laplacian graph to reconstruct k-nearest neighbor, where k eigen vectors corresponding to the k smallest eigen values are preserved for embedding \cite{bengio2004out}.
 
To investigate the contribution of the search optimization algorithm, we initialized each feature location randomly and applied the hill climbing methodology with the objective of minimizing the difference between features Euclidean distance matrix in the initial domain and the created REFINED image. We generated images using all the above-mentioned methods and then trained the same CNN structure that we had used for NCI regression task on the same cell lines. As an example, figure 12 shows created images using each described method for one of the randomly selected drugs of the NCI dataset. Further examples are provided through figures \cref{Image_Generation_2,Image_Generation_3,Image_Generation_4,Image_Generation_5,Image_Generation_6}
in the appendix.

\begin{figure}[t]
	\centering
	\includegraphics[width=\textwidth]{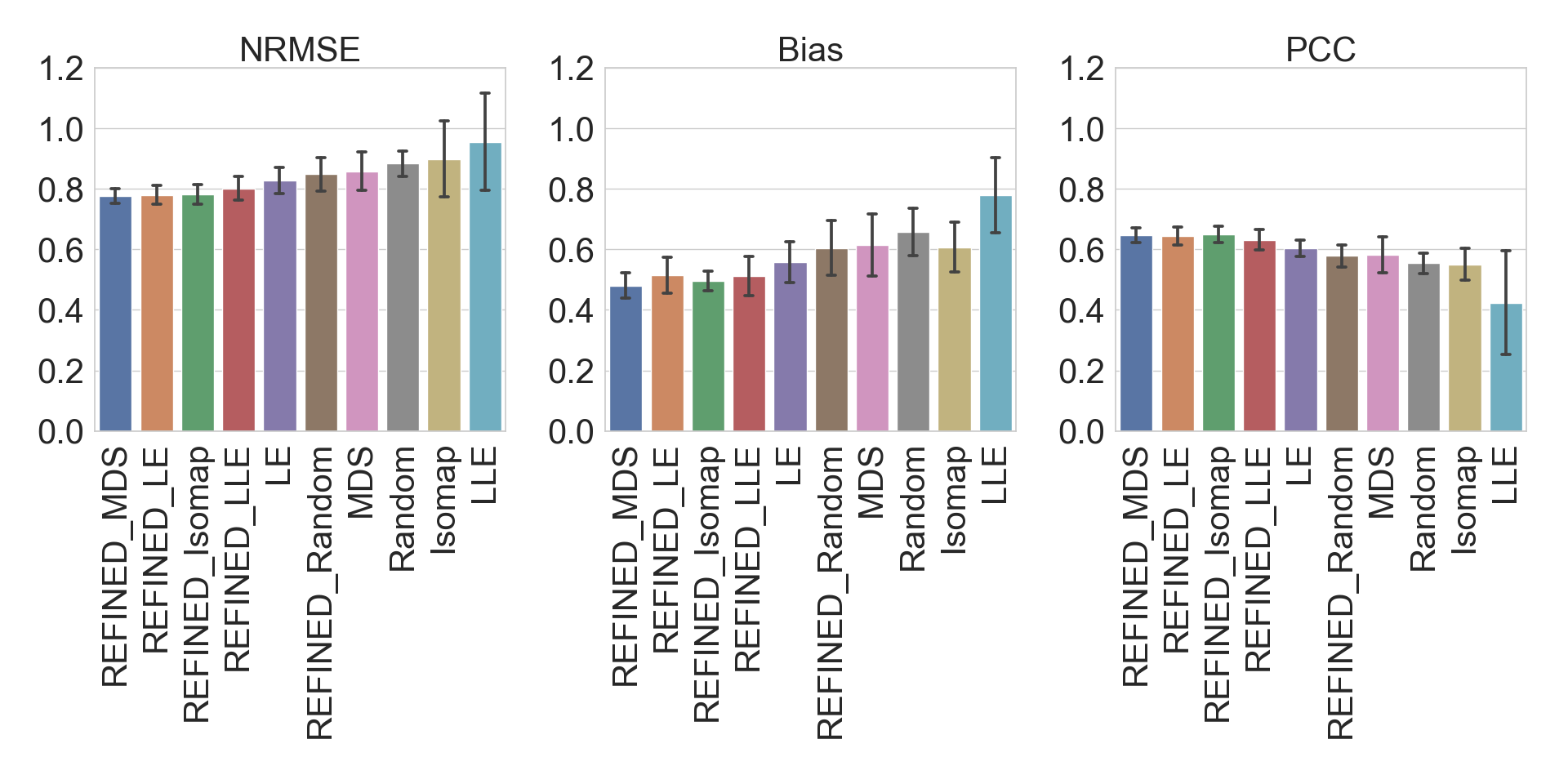} 
	\caption{Barplots of REFINED ablation study on NCI regression task using three metrics: NRMSE, PCC, and bias, where the bars show the CNN performance for different image generation methods.}
	\label{Ablation_barplot}  
\end{figure}

The average results are numerically  presented in table \ref{Ablation_Table} and the pictorial representation is included in figure \ref{Ablation_barplot} as barplots. As the results in the table \ref{Ablation_Table} indicate, NRMSE associated with REFINED initialized with Bayesian MDS is smaller than the NRMSE of all other approaches. We observed that both global and local embedding do better than random initialization followed by hill climbing. We also observed that the idea of non-overlapping pixel locations for the features borne out by the hill climbing step improves the performance if the dimensionality reduction approaches. Complete results of ablation study are provided in table \ref{Ablation_init} and \ref{Ablation_REFINED} of the appendix.

\begin{figure}[h]
	\centering
	\includegraphics[width=\textwidth]{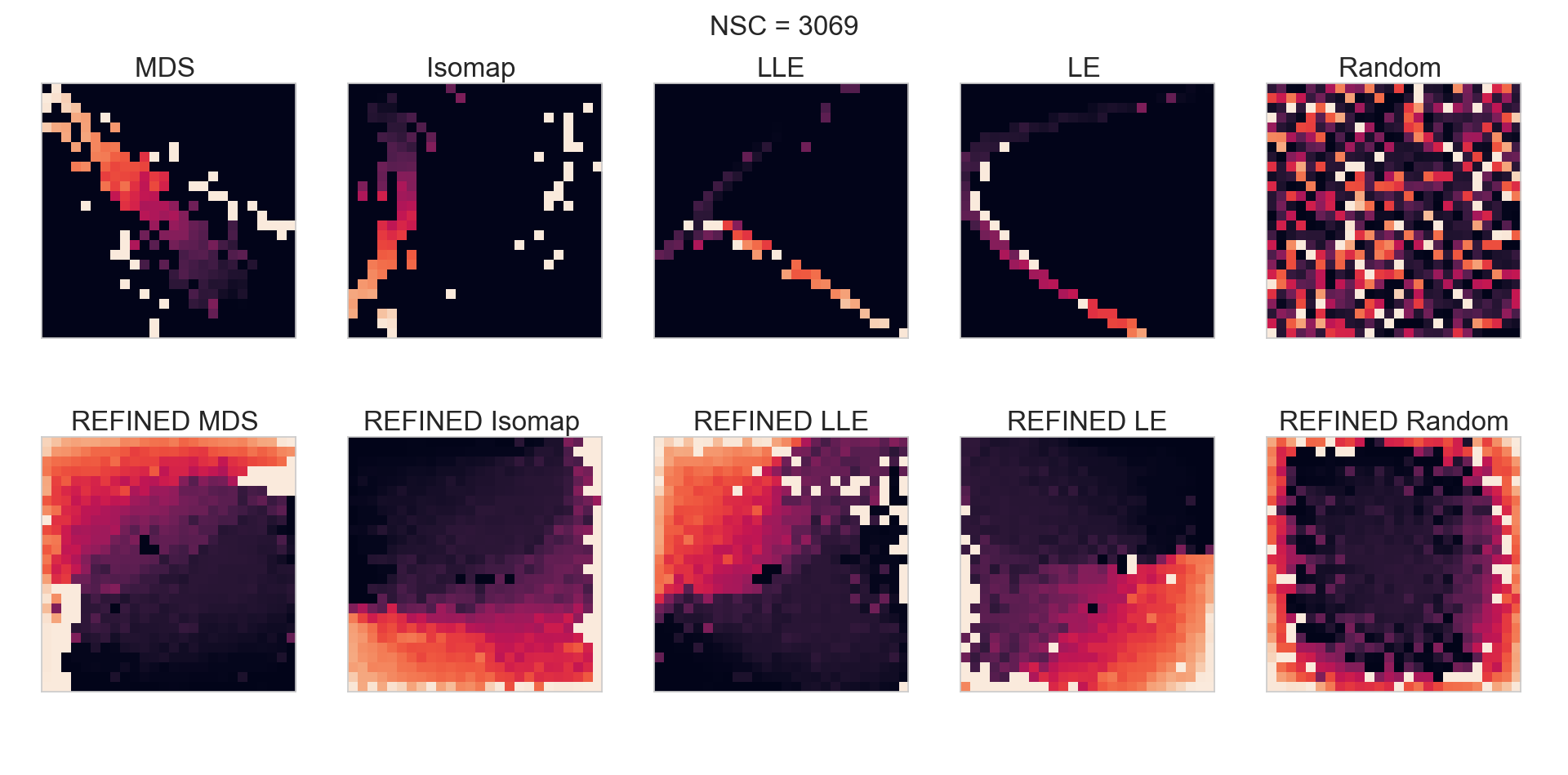} 
	\caption{Illustration of generated images of a randomly selected drug (NSC = 3069) of NCI dataset, where the first row represent pattern of images generated only with different initialization approach that included overlapping features and the second row shows the REFINED images initialized with different dimensionality techniques followed by a hill climbing step for avoiding overlap among features.}
	\label{Image_Generation}  
\end{figure}

\begin{table}[h]\caption{Average results of REFINED ablation study on NCI regression task, where the images are generated using different initialization methods with and without applying the hill climbing component.}\label{Ablation_Table}
	\centering
	\resizebox{0.7\textwidth}{!}{%
		\begin{tabular}{c||ccc||ccc}
			\multirow{2}{*}{Initialization} & \multicolumn{3}{c||}{No hill climbing} & \multicolumn{3}{c}{REFINED}                      \\
			& NRMSE      & PCC        & Bias      & NRMSE          & PCC            & Bias           \\
			\hline
			\hline
			Isomap                               & 0.900      & 0.551      & 0.608     & 0.783          & 0.649          & 0.496          \\
			Bayesian MDS                         & 0.858      & 0.582      & 0.616     & \textbf{0.776} & \textbf{0.647} & \textbf{0.481} \\
			LLE                                  & 0.956      & 0.424      & 0.780     & 0.802          & 0.632          & 0.512          \\
			LE                                   & \textbf{0.829}   & \textbf{0.603}  & \textbf{0.558}  & 0.781          & 0.645          & 0.514          \\
			Random                               & 0.884      & 0.554      & 0.659     & 0.849          & 0.578          & 0.605         
		\end{tabular}%
	}
\end{table}

\section{Discussion}

This paper presents a novel approach of converting high dimensional vectors into images with spatial neighborhood dependency that can be used as inputs to traditional Convolutional Neural Networks. The proposed methodology was conceived from the observation that deep learning CNN has increased the prediction accuracy in many scenarios especially when the inputs are images, but it is not usually appropriate when high dimensional vectors with limited neighborhood correlations are used as inputs. Our REFINED approach produces a mapping of the input features such that spatial neighbors are close and far away points in the map are distant in initial feature space. 
 
There are several advantages of the proposed REFINED methodology. First, the REFINED mapping to a compact image space appears to allow for \emph{automated feature extraction} using deep learning CNN architecture.  Using a synthetic dataset and altering the amount of spurious features, we observed that REFIEND-CNN is able to significantly outperform other approaches for scenarios with larger percentage of spurious features as shown in figure \ref{Sim100}. 

Second, the REFINED CNN approach provides a \emph{gain in predictive accuracy} as compared to other commonly used predictive models of Artificial Neural Networks, Random Forest, Support vector Machines, and Elastic Net. We have validated the performance of REFINED-CNN in  (a) synthetic dataset, (b) NCI60 drug response dataset and (c) GDSC dataset that combines chemical descriptors of drugs with genomic expressions of cell lines. REFINED CNN also outperforms other state of the art methods such as Deep-Resp-Forest and Heterogeneous graph models. REFINED CNN also outperforms other existing approaches in terms fo statistical significance and robustness.

Third,  REFINED-CNN methodology can also be used to seamlessly \emph{combine heterogeneous predictors} where each predictor can be mapped to an image as was done for the GDSC prediction scenario. Perhaps the biggest advantage of REFINED-CNN is that it has the  potential to combine multi-type predictors, where some predictors are images, some high dimensional vectors and some having functional forms. In principle, each type of predictor data can be individually mapped to images and the corresponding images can be used as inputs in a CNN architecture. Finally, we observe that REFINED-CNN has better ability to automatically perform  \emph{bias correction} as compared to ANN, RF and SVR as shown in all three application scenarios (Figures \ref{Simulation}, \ref{Error Analysis} and \ref{GDSC_results}).  The proposed REFINED approach can also be used for data augmentation by using different realizations of the mapping as was discussed in section \ref{thre}. We have provided a theoretical justification to motivate how the proposed approach can map to an ordering of features if such an ordering exists. 

In terms of applications, the REFINED approach can be applied to any predictive modeling scenario where the predictors are high dimensional vectors without an explicit neighborhood based dependence structure. We motivated the application of the scenario through the drug sensitivity prediction problem where both the gene expressions and chemical descriptors are not necessarily ordered based on correlations. 

Limitations of the approach will include scenarios where the covariance structure of the features is primarily diagonal with limited correlations between any features. Furthermore, the REFINED approach is expected to benefit from the traditional CNN architecture and thus the performance benefit will require large number of samples as is required for normal CNN scenarios. Also note that, REFINED is a second order approximation under Euclidean norm. If the predictor space is non-Euclidean, the current form of REFINED will not be suitable. 

To summarize, the paper presents a novel effective tool for feature representation and multi-object regression and classification. 

\section*{Data Availability}
\begin{itemize}
	\item \textbf{NCI60:} We downloaded the GI50 data from (\url{https://dtp.cancer.gov/databases_tools/bulk_data.htm}) and the associated drug's chemical information from (\url{https://pubchem.ncbi.nlm.nih.gov/}).
	\item \textbf{GDSC:} We downloaded the IC50 responses and cell line screened data from (\url{https://www.cancerrxgene.org/downloads/bulk_download}), and the drug's chemical information from (\url{https://pubchem.ncbi.nlm.nih.gov/}).
	\item \textbf{PaDel:} To convert the chemical information of each drug to their descriptors, we used (\url{http://www.yapcwsoft.com/dd/padeldescriptor/}) software. 
\end{itemize}

\section*{Code Availability}
 The code of this work will be available at \url{https://github.com/omidbazgirTTU/REFINED}.

\bibliographystyle{ieeetr}


\begin{thebibliography}{9}
		
	\bibitem{Costello:2014}
	J. C. Costello and \emph{et al.}
	\textit{A community effort to assess and improve drug sensitivity prediction algorithms}
	Nature biotechnology, 32, (12), 1202, 2014.

	\bibitem{Wan:PLOS}
	Q. Wan and  R. Pal.
	\textit{An ensemble based top performing approach for NCI-DREAM drug sensitivity prediction challenge}
	PLOS One, 9, (6), e101183, 2014
	
	
	\bibitem{rahman2017heterogeneity}
	Rahman, Raziur and Matlock, Kevin and Ghosh, Souparno and Pal, Ranadip.
	\textit{Heterogeneity aware random forest for drug sensitivity prediction}
	Scientific Reports, 7, (1), 11347, 2017.
	
	\bibitem{rahman2017integratedmrf}
	Rahman, Raziur and Otridge, John and Pal, Ranadip.
	\textit{IntegratedMRF: random forest-based framework for integrating prediction from different data types}
	Bioinformatics, 33, (9) , 1407:1410, 2017.
	
	\bibitem{lecun2015deep}
	LeCun, Yann and Bengio, Yoshua and Hinton, Geoffrey.
	\textit{Deep learning}
	nature, 521 (7553), 436, 2015.
	
	
	\bibitem{angermueller2016deep}
	Angermueller, Christof and P{\"a}rnamaa, Tanel and Parts, Leopold and Stegle, Oliver.
	\textit{Deep learning for computational biology}
	Molecular systems biology, 12, (7), 878, 2016.
	
	
	\bibitem{wainberg2018deep}
	Wainberg, Michael and Merico, Daniele and Delong, Andrew and Frey, Brendan J.
	\textit{Deep learning in biomedicine}
	Nature biotechnology, 36 (9) : 829, 2018.

	\bibitem{bengio2012practical}
	Bengio, Yoshua.
	\textit{Practical recommendations for gradient-based training of deep architectures}
	Neural networks: Tricks of the trade, Springer, 437:478, 2012.


	\bibitem{iandola2016squeezenet}
	Iandola, Forrest N and Han, Song and Moskewicz, Matthew W and Ashraf, Khalid and Dally, William J and Keutzer, Kurt.
	\textit{Squeezenet: Alexnet-level accuracy with 50x fewer parameters and< 0.5 mb model size}
	arXiv preprint arXiv:1602.07360, 2016.

	\bibitem{xu2018prediction}
	Xu, Bowen and Ye, Deheng and Xing, Zhenchang and Xia, Xin and Chen, Guibin and Li, Shanping.
	\textit{Predicting semantically linkable knowledge in developer online forums via convolutional neural network.}
	Proceedings of the 31st IEEE/ACM International Conference on Automated Software Engineering, 51--62, 2016.
	
	\bibitem{alipanahi2015predicting}
	Alipanahi, Babak and Delong, Andrew and Weirauch, Matthew T and Frey, Brendan J.
	\textit{Predicting the sequence specificities of DNA-and RNA-binding proteins by deep learning}
	Nature biotechnology, 33 (8), 831, 2015.
	
	\bibitem{coudray2017classification}
	Coudray, Nicolas and Moreira, Andre L and Sakellaropoulos, Theodore and Fenyo, David and Razavian, Narges and Tsirigos, Aristotelis.
	\textit{Classification and mutation prediction from non-small cell lung cancer histopathology images using deep learning}
	Nature medicine, 24, (10), 1559, 2018.
	
	\bibitem{ma2018omicsmapnet} 
	Ma, Shiyong and Zhang, Zhen.
	\textit{OmicsMapNet: Transforming omics data to take advantage of Deep Convolutional Neural Network for discovery}
	arXiv preprint arXiv:1804.05283, 2018.
	
	\bibitem{shneiderman1998tree}
	Shneiderman, Ben
	\textit{Tree visualization with tree-maps: A 2-d space-filling approach}
	Transactions on Graphics11 (1):92–99, 1998


	\bibitem{ruff2018deep}
	Ruff, Lukas and G{\"o}rnitz, Nico and Deecke, Lucas and Siddiqui, Shoaib Ahmed and Vandermeulen, Robert and Binder, Alexander and M{\"u}ller, Emmanuel and Kloft, Marius.
	\textit{Deep one-class classification.}
	International Conference on Machine Learning, 4390:4399, 2018.
	
	\bibitem{esteva2019guide}
	Esteva, Andre and Robicquet, Alexandre and Ramsundar, Bharath and Kuleshov, Volodymyr and DePristo, Mark and Chou, Katherine and Cui, Claire and Corrado, Greg and Thrun, Sebastian and Dean, Jeff.
	\textit{A guide to deep learning in healthcare}
	Nature medicine, 25, (1), 24, 2019.

	\bibitem{matlock2018investigation}
	Matlock, Kevin and De Niz, Carlos and Rahman, Raziur and Ghosh, Souparno and Pal, Ranadip.
	\textit{Investigation of model stacking for drug sensitivity prediction}
	BMC bioinformatics, 19, (3), 71, 2018.

	\bibitem{chang2018cancer}
	Chang, Yoosup and \emph{et al.}
	\textit{Cancer Drug Response Profile scan (CDRscan): A Deep Learning Model That Predicts Drug Effectiveness from Cancer Genomic Signature.}
	Scientific reports, 8,(1), 8857, 2018.
	
	\bibitem{chiu2019predicting}
	Chiu, Yu-Chiao and \emph{et al.}
	\textit{Predicting drug response of tumors from integrated genomic profiles by deep neural networks}
	BMC medical genomics, 12 , (1), 18, 2019.
	
	
	\bibitem{davison1983multidimensional}
	Davison, Mark Leonard.
	\textit{Multidimensional scaling}
	Wiley New York, (85), 1983.


	\bibitem{urpa2019focused}
	Urpa, Lea M and Anders, Simon
	\textit{Focused multidimensional scaling: interactive visualization for exploration of high-dimensional data}
	BMC bioinformatics, 20, (1), 221, 2019

	\bibitem{shoemaker2006nci60}
	Shoemaker, Robert H.
	\textit{The NCI60 human tumour cell line anticancer drug screen}
	Nature Reviews Cancer, 6 (10):813,2006.
	
	\bibitem{yap2011padel}
	Yap, Chun Wei.
	\textit{PaDEL-descriptor: An open source software to calculate molecular descriptors and fingerprints}
	Journal of computational chemistry, 32, (7) , 1466:1474, 2011.


	\bibitem{yang2012genomics}
	Yang and \emph{et al.}.
	\textit{Genomics of Drug Sensitivity in Cancer (GDSC): a resource for therapeutic biomarker discovery in cancer cells}
	Nucleic acids research, 41, (1), 955:961, 2012.
	
	\bibitem{barretina2012cancer}
	Barretina, Jordi and Caponigro, Giordano and Stransky, Nicolas and Venkatesan, Kavitha and Margolin, Adam A and Kim, Sungjoon and Wilson, Christopher J and Leh{\'a}r, Joseph and Kryukov, Gregory V and Sonkin, Dmitriy and others
	\textit{The Cancer Cell Line Encyclopedia enables predictive modelling of anticancer drug sensitivity}
	
	\bibitem{wold1987principal}
	Wold, Svante and Esbensen, Kim and Geladi, Paul
	\textit{Principal component analysis}
	Chemometrics and intelligent laboratory systems, 2, (1-3),37--52, 1987, Elsevier

	\bibitem{pal2016predictive}
	Pal, Ranadip.
	\textit{Predictive modeling of drug sensitivity.}
	Elsevier Inc., 2016. 342 p
	
	\bibitem{Kira:1992:FSP:1867135.1867155}
	Kira, Kenji and Rendell, Larry A.
	\textit{The Feature Selection Problem: Traditional Methods and a New Algorithm.}
	Proceedings of the Tenth National Conference on Artificial Intelligence, AAAI Press, 129--134, 6, 1992.
	

	
	\bibitem{zhang2012bias}
	Zhang, Guoyi and Lu, Yan.
	\textit{Bias-corrected random forests in regression}
	Journal of Applied Statistics, 39, (1), 151:160, 2012.
	
	
	\bibitem{smouse1986multiple}
	Smouse, Peter E and Long, Jeffrey C and Sokal, Robert R.
	\textit{Multiple regression and correlation extensions of the Mantel test of matrix correspondence.}
	Systematic zoology, 35, (4), 627:632, 1986.	
	
	\bibitem{efron1992bootstrap}
	Efron, Bradley.
	\textit{Bootstrap methods: another look at the jackknife}
	Breakthroughs in statistics, Springer, 569--593, 1992

	
	
	
	
	\bibitem{chen2011stringing}
	Chen, Kun and Chen, Kehui and M{\"u}ller, Hans-Georg and Wang, Jane-Ling.
	\textit{Stringing high-dimensional data for functional analysis}
	Journal of the American Statistical Association, Taylor \& Francis, 106, (493), 275--284, 2011.
	
	\bibitem{oh2001bayesian}
	Oh, Man-Suk and Raftery, Adrian E
	\textit{Bayesian multidimensional scaling and choice of dimension}
	Journal of the American Statistical Association, Taylor \& Francis, 96, (455), 1031--1044, 2001.
	
	\bibitem{macnab2001autoregressive}
	MacNab, Ying C and Dean, CB.
	\textit{Autoregressive spatial smoothing and temporal spline smoothing for mapping rates},
	  Biometrics, Wiley Online Library, 57, (3), 949--956, 2001.
	  
	\bibitem{illian2008statistical}
	Illian, Janine and Penttinen, Antti and Stoyan, Helga and Stoyan, Dietrich.
	\textit{Statistical analysis and modelling of spatial point patterns}
	John Wiley \& Sons, 70, 2008.
	  
	\bibitem{chawla2002smote}
    Chawla, Nitesh V and Bowyer, Kevin W and Hall, Lawrence O and Kegelmeyer, W Philip.
	\textit{SMOTE: synthetic minority over-sampling technique}
	Journal of Artificial Intelligence Research, 16, 321--357, 2002.
	
	\bibitem{ghosh2012k}
	Ghosh, Souparno and Gelfand, Alan E and Zhu, Kai and Clark, James S.
	\textit {The k-ZIG: flexible modeling for zero-inflated counts},
	Biometrics, Wiley Online Library, 68, (3), 878--885, 2012.
	
	\bibitem{chandler2013spatially}
	Chandler, Richard B and Royle, J Andrew and others.
	\textit{Spatially explicit models for inference about density in unmarked or partially marked populations}
	The Annals of Applied Statistics,,7, 2, 936--954, 2013.
	  
	\bibitem{ghosh2006bayesian}
	Ghosh, Sujit K and Mukhopadhyay, Pabak and Lu, Jye-Chyi JC.
	\textit{Bayesian analysis of zero-inflated regression models}  
	 Journal of Statistical planning and Inference, Elsevier, 136, (4), 1360--1375, 2006.
 
    \bibitem{dietterich1998approximate}
    Dietterich, Thomas G
    \textit{Approximate statistical tests for comparing supervised classification learning algorithms}
	Neural computation, MIT Press, 10, (7), 1895--1923, 1998.
	
	\bibitem{tibshirani2001estimating}
	Tibshirani, Robert and Walther, Guenther and Hastie, Trevor.
	\textit{Estimating the number of clusters in a data set via the gap statistic}
	Journal of the Royal Statistical Society: Series B (Statistical Methodology), 63, 2, 411--423, 2001
	
	\bibitem{witten2005practical}
	Witten, Ian H and Frank, Eibe and Hall, Mark A.
	\textit{Practical machine learning tools and techniques}
	Morgan Kaufmann, 578, 2005.
	
	\bibitem{vollset1993confidence}
	Vollset, Stein Emil
	\textit{Confidence intervals for a binomial proportion}
	Statistics in medicine, 12, 9, 809--824, 1993.
	
	\bibitem{tenenbaum2000global}
	Tenenbaum, Joshua B and De Silva, Vin and Langford, John C
	\textit{A global geometric framework for nonlinear dimensionality reduction}
	science, American Association for the Advancement of Science, 290, (5500), 2319--2323, 2000.

	\bibitem{roweis2000nonlinear}
	Roweis, Sam T and Saul, Lawrence K
	\textit{Nonlinear dimensionality reduction by locally linear embedding}
	science, American Association for the Advancement of Science, 290, (5500), 2323--2326, 2000.
	
	\bibitem{belkin2003laplacian}
	Belkin, Mikhail and Niyogi, Partha
	\textit{Laplacian eigenmaps for dimensionality reduction and data representation}
	Neural computation, MIT Press, 15, (6), 1373--1396, 2003.
	
	\bibitem{bengio2004out}
	Bengio, Yoshua and Paiement, Jean-fran{\c{c}}cois and Vincent, Pascal and Delalleau, Olivier and Roux, Nicolas L and Ouimet, Marie
	\textit{Out-of-sample extensions for lle, isomap, mds, eigenmaps, and spectral clustering}
	Advances in neural information processing systems, 177--184, 2004.
	
	\bibitem{su2019deep}
	Su, Ran and Liu, Xinyi and Wei, Leyi and Zou, Quan
	\textit{Deep-Resp-Forest: A deep forest model to predict anti-cancer drug response}
	Elsevier, Methods, 166, 91--102, 2019.
	
	\bibitem{zhang2019heterogeneous}
	Zhang, Chuxu and Song, Dongjin and Huang, Chao and Swami, Ananthram and Chawla, Nitesh V
	\textit{Heterogeneous graph neural network}
	Proceedings of the 25th ACM SIGKDD International Conference on Knowledge Discovery \& Data Mining, 793--803, 2019.
	
	\bibitem{lim2019predicting}
	Lim, Jaechang and Ryu, Seongok and Park, Kyubyong and Choe, Yo Joong and Ham, Jiyeon and Kim, Woo Youn.
	\textit{Predicting Drug--Target Interaction Using a Novel Graph Neural Network with 3D Structure-Embedded Graph Representation}
	ACS Publications, Journal of chemical information and modeling, 59, 9, 3981--3988, 2019.
	
	\bibitem{cawley2010over}
	Cawley, Gavin C and Talbot, Nicola LC.
	\textit{On over-fitting in model selection and subsequent selection bias in performance evaluation}
	Journal of Machine Learning Research 11, July, 2079--2107, 2010.
	
	\bibitem{scikit-learn}
	Pedregosa, F. and Varoquaux, G. and Gramfort, A. and Michel, V. and Thirion, B. and Grisel, O. and Blondel, M. and Prettenhofer, P. and Weiss, R. and Dubourg, V. and Vanderplas, J. and Passos, A. and Cournapeau, D. and Brucher, M. and Perrot, M. and Duchesnay, E.
	\textit{Scikit-learn: Machine Learning in {P}ython}
	Journal of Machine Learning Research, 12, 2825--2830, 2011.
	
	\bibitem{glorot2011deep}
	Glorot, Xavier and Bordes, Antoine and Bengio, Yoshua.
	\textit{Deep sparse rectifier neural networks}
	Proceedings of the fourteenth international conference on artificial intelligence and statistics, 315--323, 2011.
	
	\bibitem{ioffe2015batch}
	Ioffe, Sergey and Szegedy, Christian.
	\textit{Batch normalization: Accelerating deep network training by reducing internal covariate shift}
	arXiv preprint arXiv:1502.03167, 2015
	
	\bibitem{chollet2015keras}
	Chollet, Fran\c{c}ois and others.
	\textit{Keras}
	\url{https://keras.io}, 2015
	
\end{thebibliography}

\newpage
\appendix
\section*{Appendix}
\begin{subappendices}
\addcontentsline{toc}{section}{Appendices}
\renewcommand{\thesubsection}{\Alph{subsection}}

\subsection{Syntethic dataset}
	

\begin{figure}[!htbp]
	\centering
	\includegraphics[width=\textwidth]{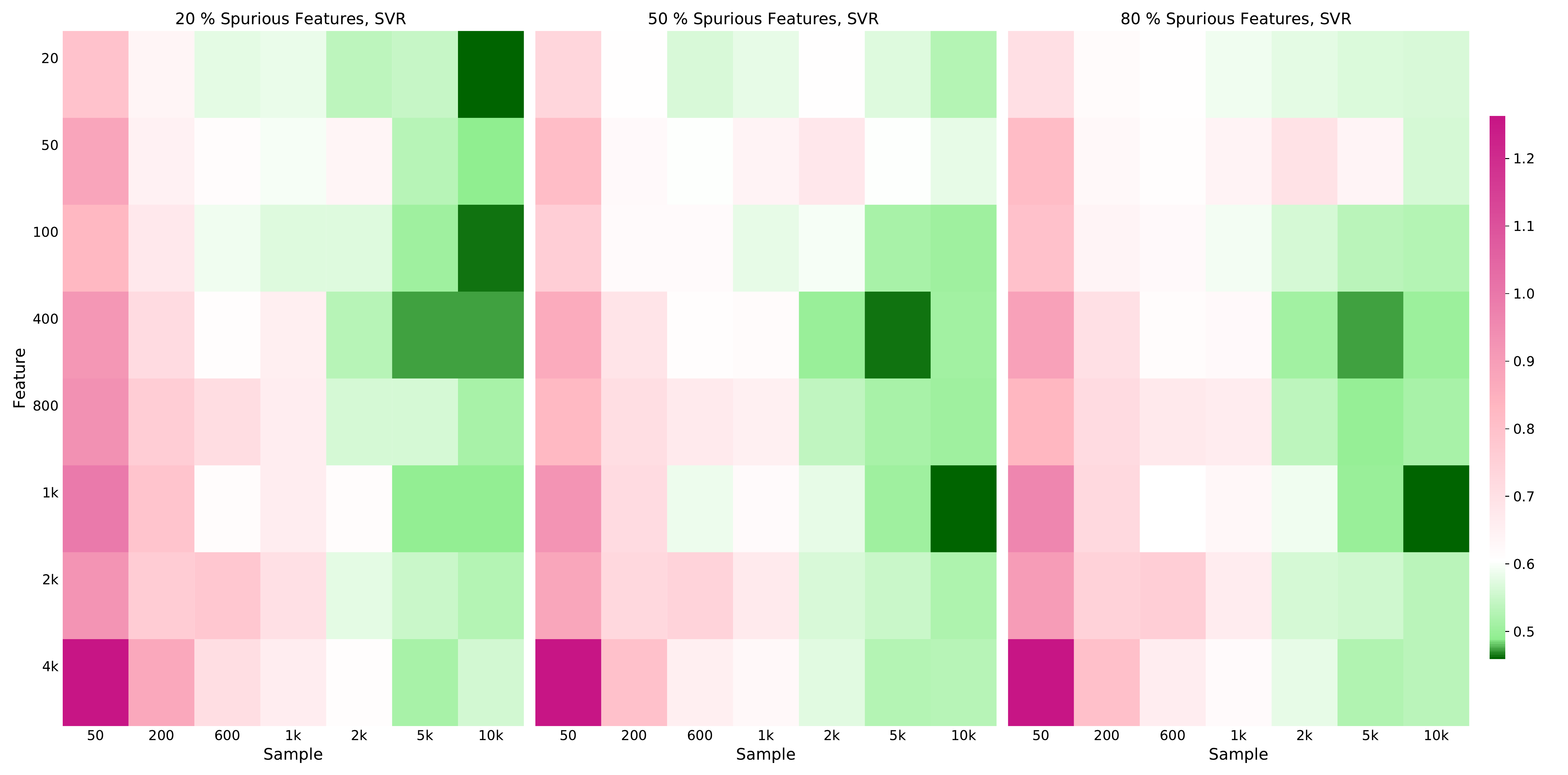} 
	\caption{SVR's NRMSE heatmap for different sample sizes and with different number of features where the 20 \%, 50 \% and 80 \% of the features are respectively spurious.} 
	\label{SVRHeat} 
\end{figure}

\begin{figure}[!htbp]
	\centering
	\includegraphics[width=\textwidth]{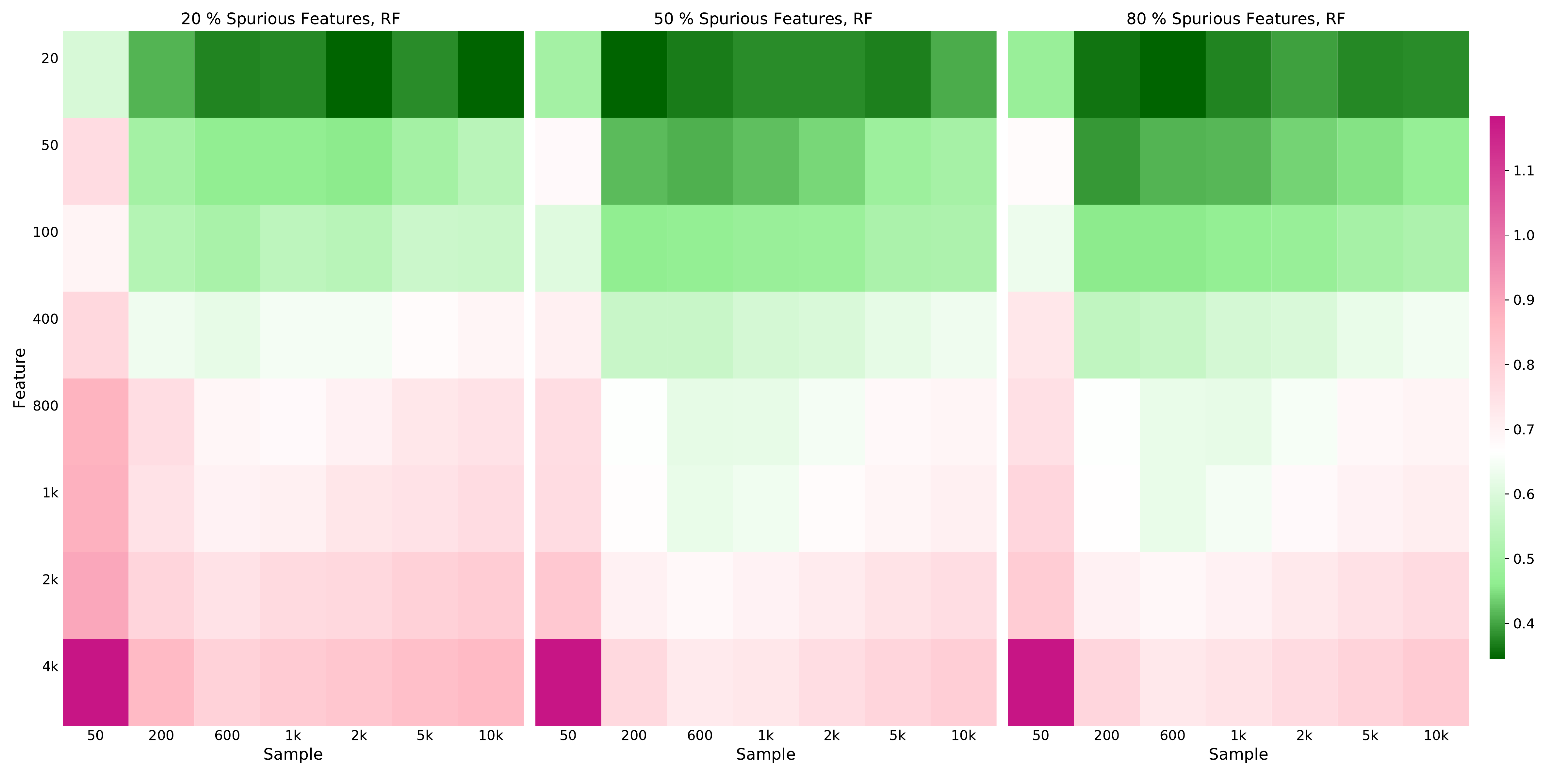} 
	\caption{RF's NRMSE heatmap for different sample sizes and with different number of features where the 20 \%, 50 \% and 80 \% of the features are respectively spurious.}
	\label{RFHeat} 
\end{figure}

\begin{figure}[!htbp]
	\centering
	\includegraphics[width=\textwidth]{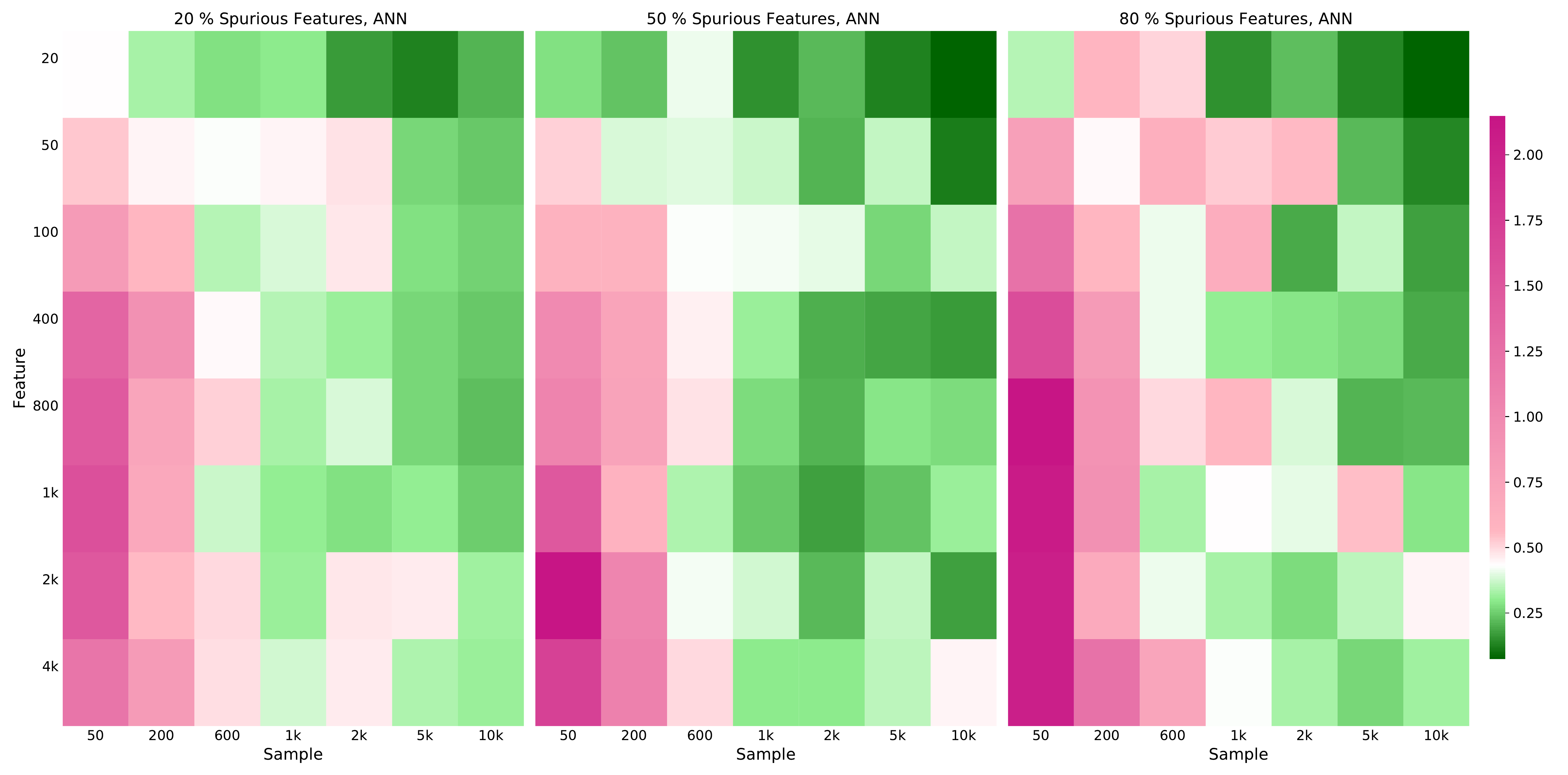} 
	\caption{ANN's NRMSE heatmap for different sample sizes and with different number of features where the 20 \%, 50 \% and 80 \% of the features are respectively spurious.}
	\label{ANNHeat} 
\end{figure}

\begin{figure}[!htbp]
	\centering
	\includegraphics[width=\textwidth]{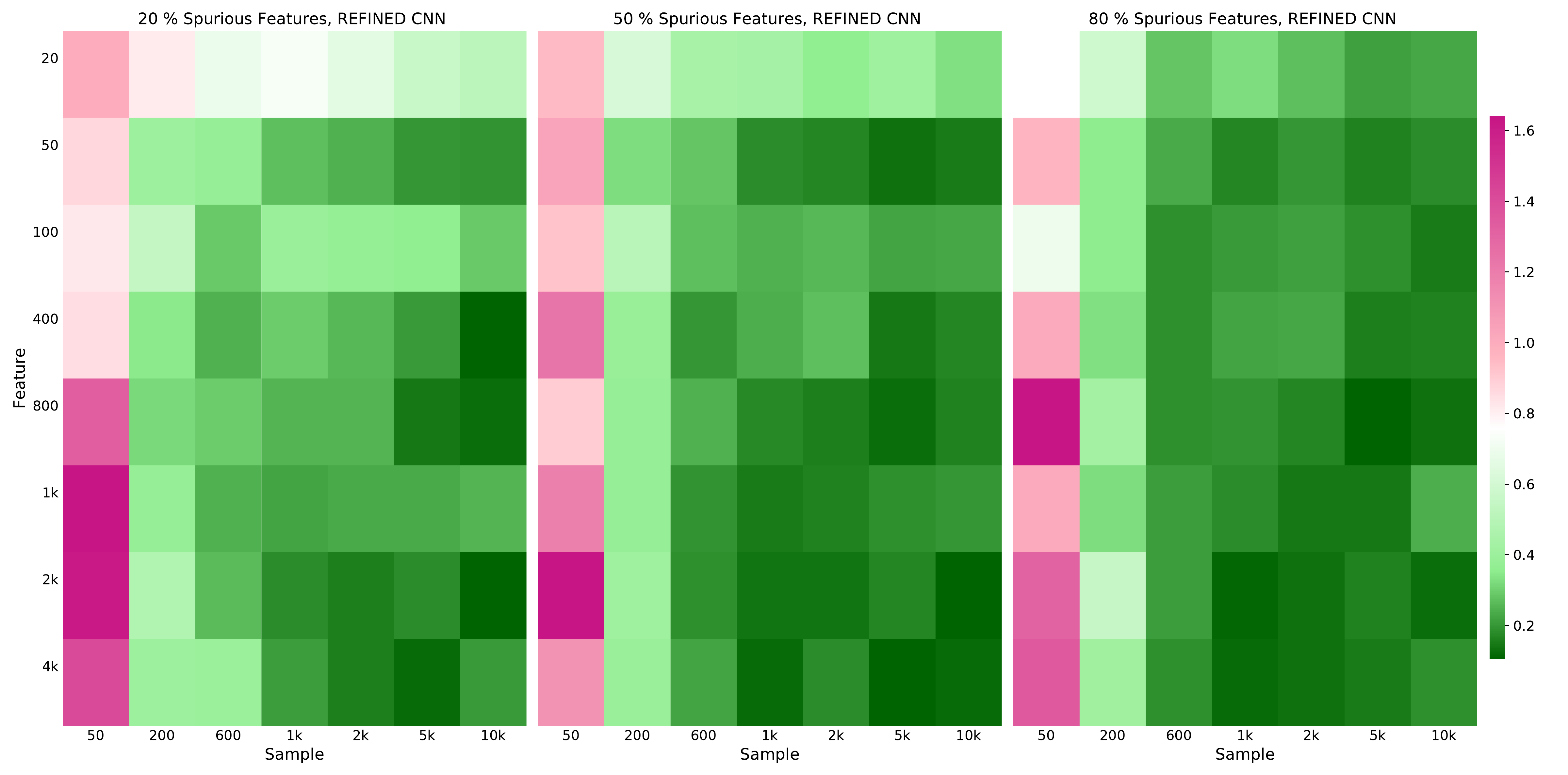} 
	\caption{REFINED CNN's NRMSE heatmap for different sample sizes and with different number of features where the 20 \%, 50 \% and 80 \% of the features are respectively spurious.}
	\label{CNNHeat} 
\end{figure}

\begin{figure}[!htbp]
	\begin{subfigure}[h]{\linewidth}
		\centering
		\includegraphics[width=\linewidth]{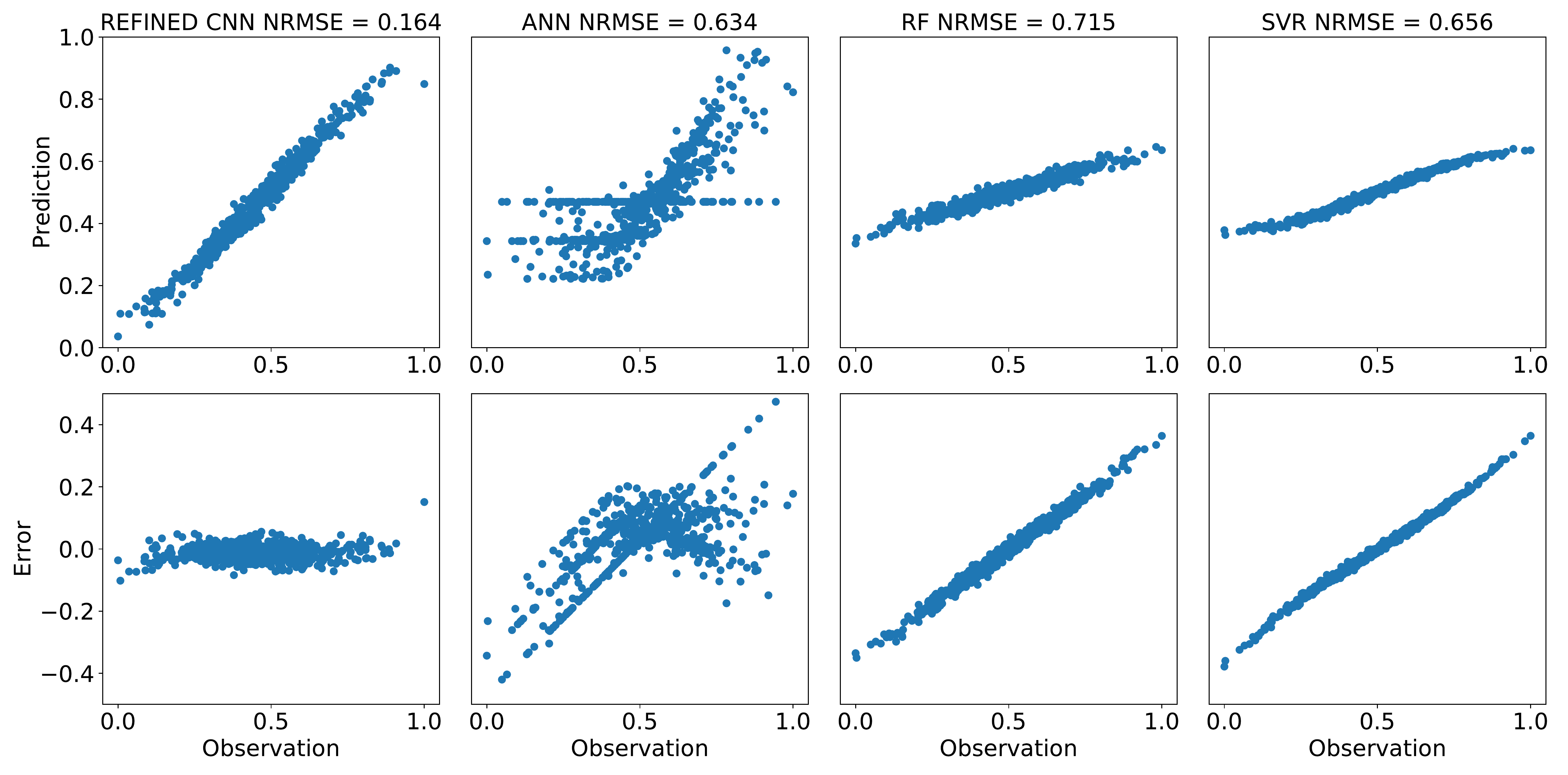} 
		\caption{4000 features with 80 \% spurious rate, 600 samples}
		\label{Synthetic:b} 
		\vspace{4ex}
	\end{subfigure} 
	\begin{subfigure}[h]{\linewidth}
		\centering
		\includegraphics[width=\linewidth]{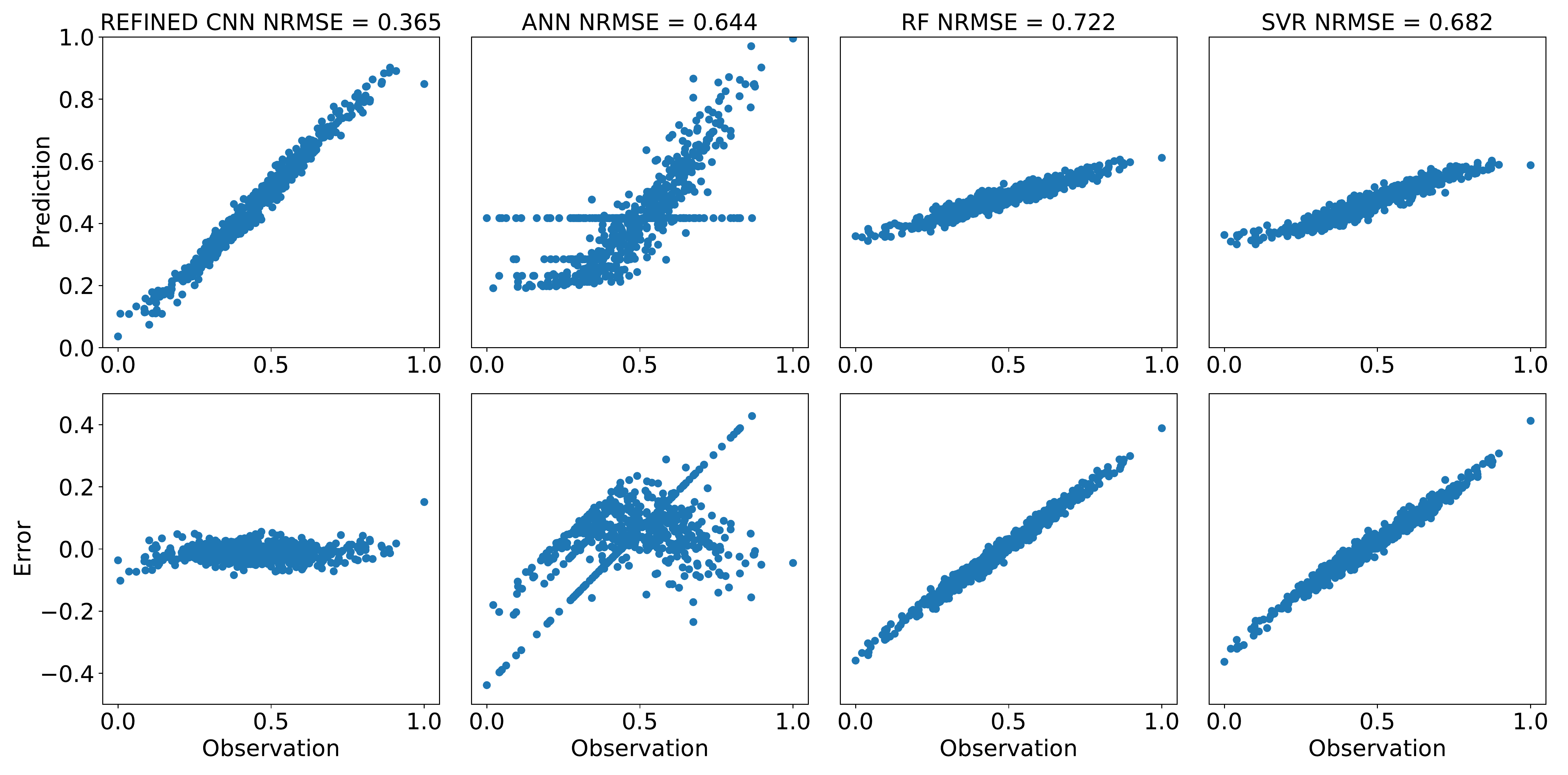} 
		\caption{4000 features with 20 \% spurious rate, 600 samples}
		\label{Synthetic:a} 
		\vspace{4ex}
	\end{subfigure}
	\caption{Prediction vs observation and Error vs observation of the simulated data for REFINED CNN, ANN, RF, and SVR, where number of features, number of samples are 4000 and 600  respectively. spurious feature percentage for figure (a) is 80 \% and for figure (b) is 20 \%. Note that the scatter plot for REFINED-CNN closely follows a straight line with unit slope indicating predictive accuracy of our approach. RF and SVR reveal their well-known tendency  to under-predict higher valued observation and over-predict lower valued observations. REFINED-CNN bias is also better than the bias observed for the ANN scenario.}
	\label{Simulation} 
\end{figure}

\newpage

\subsection{NCI60 dataset}

\begin{table}[b]
	\centering
	\caption{Number of drugs applied on each selected cell line of the NCI60 dataset}
	\label{NCI_sample}
	\resizebox{\textwidth}{!}{%
		\begin{tabular}{c||ccccccccccccccccc}
			Cell line         & CCRF\_CEM & COLO\_205 & DU\_145 & EKVX  & HCC\_2998 & MDA\_MB\_435 & SNB\_78 & NCI\_ADR\_RES & 786\_0 & A498  & A549\_ATCC & ACHN  & BT\_549 & CAKI\_1 & DLD\_1 & DMS\_114 & DMS\_273 \\
			\hline
			\hline
			\#drugs (samples) & 47571     & 49844     & 36622   & 47732 & 45192     & 36868        & 14006   & 37156         & 49453  & 43853 & 50703      & 49760 & 33466   & 47410   & 14650  & 14937    & 13728   
		\end{tabular}%
	}
\end{table}

\begin{figure}[!htbp]
	\begin{subfigure}[h]{\linewidth}
		\centering
		\includegraphics[width=\linewidth]{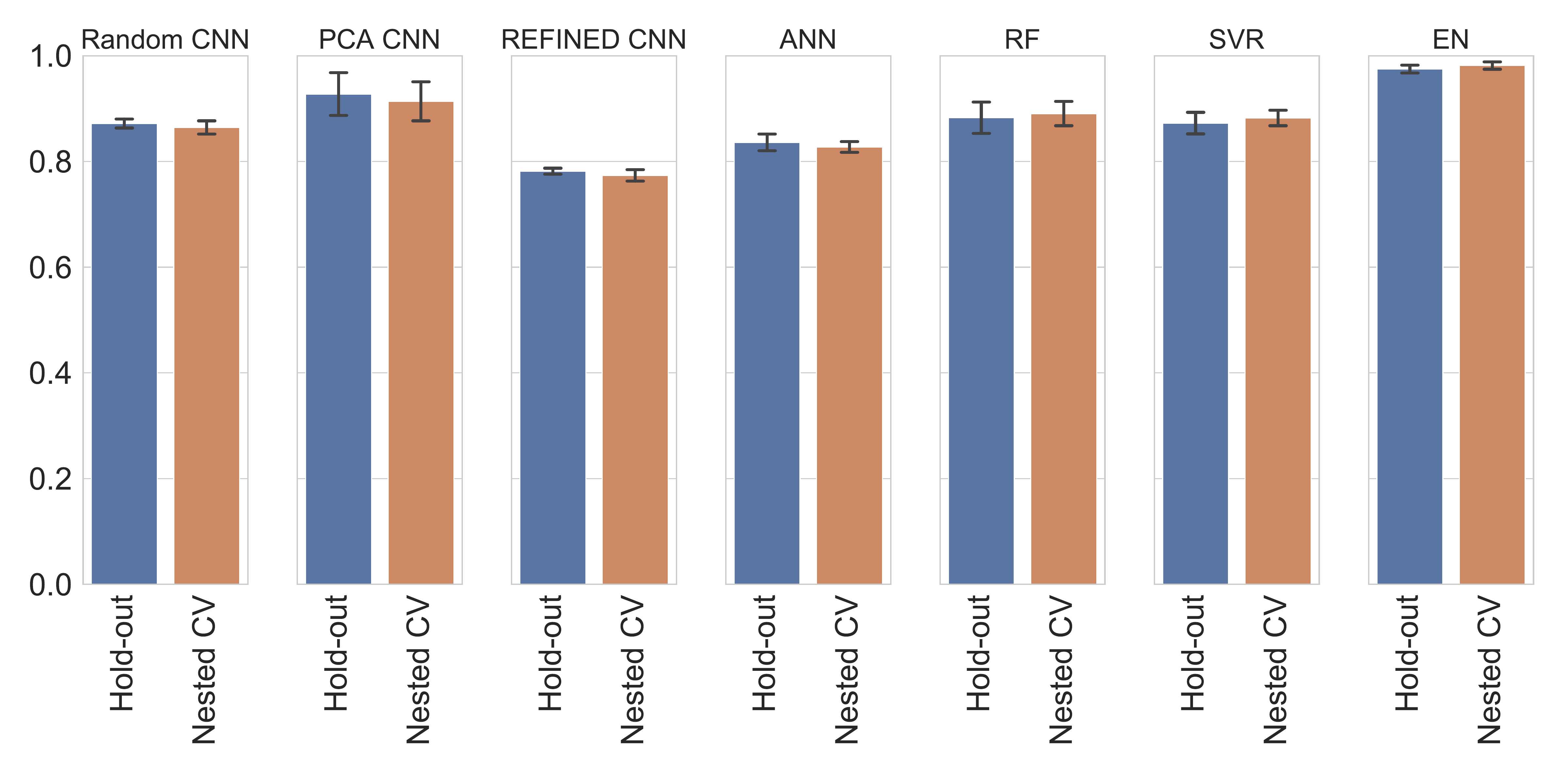} 
		\caption{Hold-out versus nested cross-validation pairwise comparison}
		\label{HyperPar:b} 
		\vspace{4ex}
	\end{subfigure} 
	\begin{subfigure}[h]{\linewidth}
		\centering
		\includegraphics[width=\linewidth]{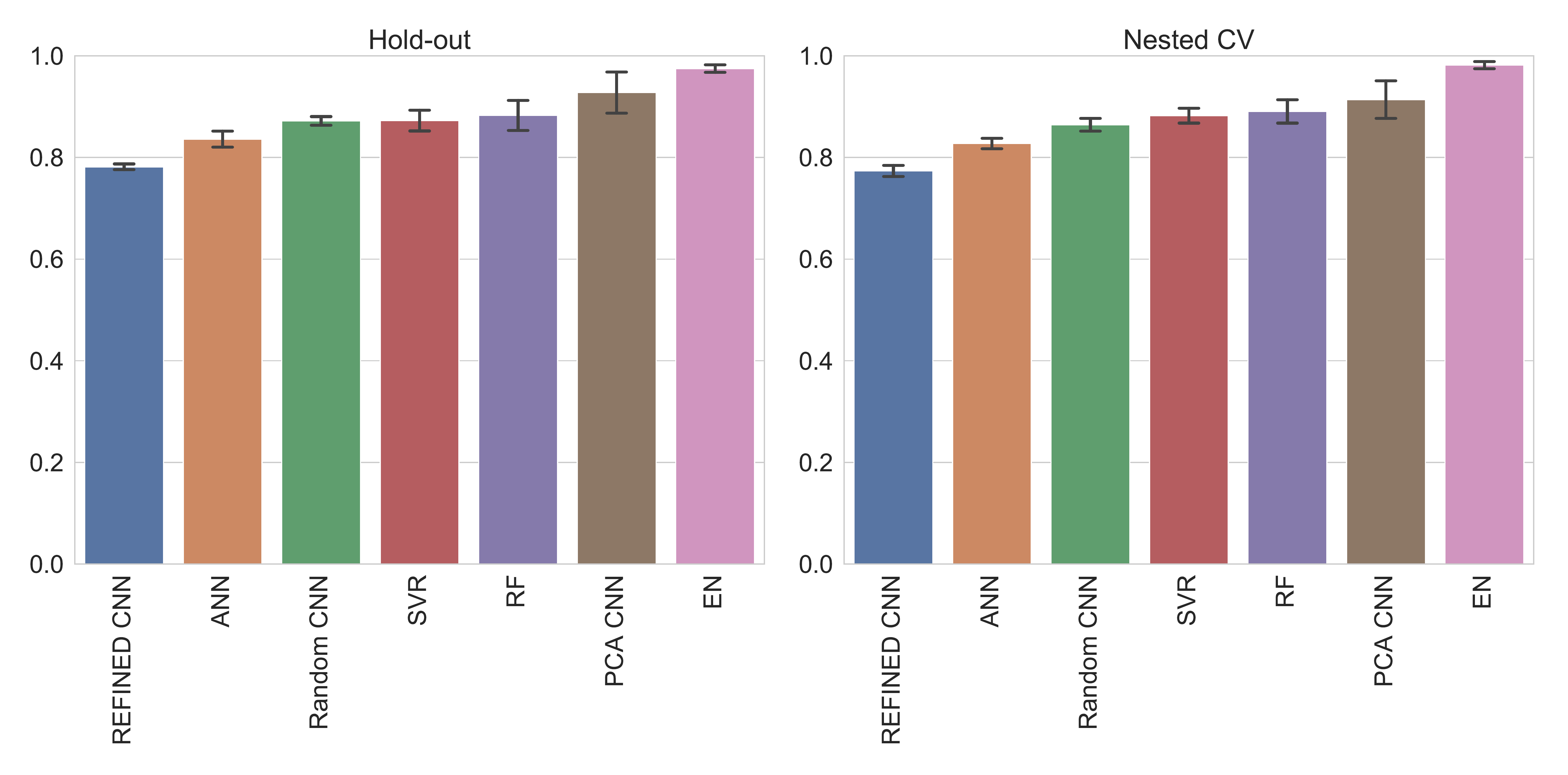} 
		\caption{Hold-out versus nested cross-validation trend comparison }
		\label{HyperPar:a} 
		\vspace{4ex}
	\end{subfigure}
	\caption{Hold-out versus nested cross-validation: a) The pairwise comparison of the models shows minimal difference between the two error estimation approaches. b) The trend comparison indicates that the comparative performance of the models does not change while using Hold-out and Nested-CV approaches. The cell lines that were selected randomly from NCI60 dataset for this comparison are, CCRF\_CEM, MDA\_MB\_435, and SNB\_78 with 47k, 37k and 14k number of drugs, respectively.}
	\label{HyperPar} 
\end{figure}

\begin{table}[!htbp]\caption{ Predicting sensitivity or resistivity of the drugs using seven classifiers part I, including Random CNN, REFINED CNN, and PCA CNN}\label{Classification}
	\centering
	\resizebox{\textwidth}{!}{%
		\begin{tabular}{c||ccccc||ccccc||ccccc}
			\multirow{2}{*}{Cell lines} & \multicolumn{5}{c||}{Random CNN} & \multicolumn{5}{c||}{REFIND CNN} & \multicolumn{5}{c}{PCA CNN}  \\
			& Accuracy & Precision & Recall & F1 Score & AUC   & Accuracy       & Precision      & Recall         & F1 Score       & AUC            & Accuracy & Precision & Recall & F1 Score & AUC   \\
			\hline
			\hline
			CCRF\_CEM                   & 0.725    & 0.723     & 0.725  & 0.723    & 0.716 & \textbf{0.752} & \textbf{0.751} & \textbf{0.752} & \textbf{0.75}  & \textbf{0.743} & 0.721    & 0.72      & 0.721  & 0.717    & 0.708 \\
			COLO\_205                   & 0.748    & 0.748     & 0.748  & 0.748    & 0.748 & \textbf{0.762} & \textbf{0.762} & \textbf{0.762} & \textbf{0.762} & \textbf{0.762} & 0.734    & 0.734     & 0.734  & 0.734    & 0.734 \\
			DU\_145                     & 0.734    & 0.734     & 0.734  & 0.734    & 0.734 & \textbf{0.753} & \textbf{0.753} & \textbf{0.753} & \textbf{0.752} & \textbf{0.752} & 0.713    & 0.713     & 0.713  & 0.713    & 0.713 \\
			EKVX                        & 0.713    & 0.713     & 0.713  & 0.713    & 0.712 & \textbf{0.747} & \textbf{0.748} & \textbf{0.747} & \textbf{0.747} & \textbf{0.747} & 0.709    & 0.709     & 0.709  & 0.708    & 0.707 \\
			HCC\_2998                   & 0.71     & 0.711     & 0.71   & 0.709    & 0.709 & \textbf{0.758} & \textbf{0.758} & \textbf{0.758} & \textbf{0.758} & \textbf{0.758} & 0.718    & 0.718     & 0.718  & 0.717    & 0.717 \\
			MDA\_MB\_435                & 0.713    & 0.712     & 0.713  & 0.706    & 0.692 & \textbf{0.757} & \textbf{0.757} & \textbf{0.757} & \textbf{0.754} & \textbf{0.742} & 0.692    & 0.696     & 0.692  & 0.693    & 0.69  \\
			SNB\_78                     & 0.733    & 0.723     & 0.733  & 0.721    & 0.672 & \textbf{0.768} & \textbf{0.764} & \textbf{0.768} & \textbf{0.765} & \textbf{0.734} & 0.755    & 0.75      & 0.755  & 0.737    & 0.683 \\
			NCI\_ADR\_RES               & 0.707    & 0.712     & 0.707  & 0.708    & 0.708 & \textbf{0.771} & \textbf{0.771} & \textbf{0.771} & \textbf{0.771} & \textbf{0.768} & 0.714    & 0.714     & 0.714  & 0.711    & 0.702 \\
			786\_0                      & 0.72     & 0.72      & 0.72   & 0.72     & 0.72  & \textbf{0.764} & \textbf{0.764} & \textbf{0.764} & \textbf{0.764} & \textbf{0.764} & 0.713    & 0.714     & 0.713  & 0.713    & 0.713 \\
			A498                        & 0.722    & 0.723     & 0.722  & 0.721    & 0.722 & \textbf{0.762} & \textbf{0.762} & \textbf{0.762} & \textbf{0.762} & \textbf{0.762} & 0.707    & 0.709     & 0.707  & 0.706    & 0.707 \\
			A549\_ATCC                  & 0.714    & 0.717     & 0.714  & 0.712    & 0.71  & \textbf{0.767} & \textbf{0.767} & \textbf{0.767} & \textbf{0.767} & \textbf{0.767} & 0.708    & 0.71      & 0.708  & 0.706    & 0.705 \\
			ACHN                        & 0.698    & 0.701     & 0.698  & 0.697    & 0.698 & \textbf{0.745} & \textbf{0.747} & \textbf{0.745} & \textbf{0.745} & \textbf{0.745} & 0.706    & 0.707     & 0.706  & 0.706    & 0.706 \\
			BT\_549                     & 0.719    & 0.718     & 0.719  & 0.716    & 0.711 & \textbf{0.747} & \textbf{0.749} & \textbf{0.747} & \textbf{0.745} & \textbf{0.739} & 0.7      & 0.701     & 0.7    & 0.7      & 0.699 \\
			CAKI\_1                     & 0.714    & 0.714     & 0.714  & 0.714    & 0.714 & \textbf{0.755} & \textbf{0.759} & \textbf{0.755} & \textbf{0.754} & \textbf{0.755} & 0.716    & 0.719     & 0.716  & 0.715    & 0.715 \\
			DLD\_1                      & 0.722    & 0.717     & 0.722  & 0.695    & 0.642 & \textbf{0.779} & \textbf{0.775} & \textbf{0.779} & \textbf{0.775} & \textbf{0.745} & 0.734    & 0.728     & 0.734  & 0.716    & 0.666 \\
			DMS\_114                    & 0.681    & 0.682     & 0.681  & 0.674    & 0.669 & \textbf{0.736} & \textbf{0.737} & \textbf{0.736} & \textbf{0.733} & \textbf{0.727} & 0.683    & 0.688     & 0.683  & 0.673    & 0.668 \\
			DMS\_273                    & 0.705    & 0.701     & 0.705  & 0.698    & 0.679 & \textbf{0.704} & \textbf{0.704} & \textbf{0.704} & \textbf{0.691} & \textbf{0.669} & 0.762    & 0.76      & 0.762  & 0.76     & 0.747 \\
			\hline
			\hline
			Average & 0.716  & 0.716     & 0.716  & 0.712    & 0.703 & \textbf{0.754} & \textbf{0.755} & \textbf{0.754} & \textbf{0.753} & \textbf{0.746} & 0.717    & 0.717     & 0.717  & 0.713    & 0.705 \\
			
		\end{tabular}%
	}
\end{table}

\begin{table}[!htbp]\caption{ Predicting sensitivity or resistivity of the drugs using seven classifiers part II, including LR, RF, ANN, and SVM}\label{Classification2}
	\centering
	\resizebox{\textwidth}{!}{%
		\begin{tabular}{c||ccccc||ccccc||ccccc||ccccc}
			\multirow{2}{*}{Cell lines} & \multicolumn{5}{c||}{LR}                           & \multicolumn{5}{c||}{RF}                           & \multicolumn{5}{c||}{SVM}                          & \multicolumn{5}{c}{ANN}                          \\
			& Accuracy & Precision & Recall & F1 Score & AUC   & Accuracy & Precision & Recall & F1 Score & AUC   & Accuracy & Precision & Recall & F1 Score & AUC   & Accuracy & Precision & Recall & F1 Score & AUC   \\
			\hline
			\hline
			CCRF\_CEM                   & 0.666    & 0.672     & 0.666  & 0.647    & 0.638 & 0.695    & 0.695     & 0.695  & 0.689    & 0.678 & 0.685    & 0.693     & 0.685  & 0.67     & 0.66  & 0.698    & 0.697     & 0.698  & 0.692    & 0.682 \\
			COLO\_205                   & 0.676    & 0.681     & 0.676  & 0.674    & 0.676 & 0.697    & 0.697     & 0.697  & 0.697    & 0.697 & 0.689    & 0.69      & 0.689  & 0.688    & 0.689 & 0.704    & 0.709     & 0.704  & 0.702    & 0.704 \\
			DU\_145                     & 0.672    & 0.681     & 0.672  & 0.665    & 0.667 & 0.698    & 0.699     & 0.698  & 0.698    & 0.697 & 0.687    & 0.69      & 0.687  & 0.685    & 0.685 & 0.707    & 0.709     & 0.707  & 0.706    & 0.708 \\
			EKVX                        & 0.655    & 0.658     & 0.655  & 0.655    & 0.657 & 0.669    & 0.669     & 0.669  & 0.669    & 0.669 & 0.659    & 0.659     & 0.659  & 0.659    & 0.659 & 0.672    & 0.673     & 0.672  & 0.671    & 0.67  \\
			HCC\_2998                   & 0.662    & 0.667     & 0.662  & 0.661    & 0.663 & 0.69     & 0.69      & 0.69   & 0.69     & 0.69  & 0.688    & 0.689     & 0.688  & 0.688    & 0.689 & 0.705    & 0.708     & 0.705  & 0.704    & 0.705 \\
			MDA\_MB\_435                & 0.682    & 0.687     & 0.682  & 0.664    & 0.65  & 0.699    & 0.698     & 0.699  & 0.691    & 0.677 & 0.694    & 0.699     & 0.694  & 0.678    & 0.663 & 0.706    & 0.705     & 0.706  & 0.7      & 0.687 \\
			SNB\_78                     & 0.735    & 0.736     & 0.735  & 0.704    & 0.644 & 0.755    & 0.75      & 0.755  & 0.74     & 0.687 & 0.732    & 0.748     & 0.732  & 0.689    & 0.627 & 0.742    & 0.741     & 0.742  & 0.716    & 0.656 \\
			NCI\_ADR\_RES               & 0.69     & 0.696     & 0.69   & 0.677    & 0.668 & 0.706    & 0.704     & 0.706  & 0.702    & 0.694 & 0.7      & 0.703     & 0.7    & 0.69     & 0.68  & 0.712    & 0.711     & 0.712  & 0.708    & 0.699 \\
			786\_0                      & 0.682    & 0.687     & 0.682  & 0.679    & 0.681 & 0.698    & 0.698     & 0.698  & 0.698    & 0.698 & 0.695    & 0.697     & 0.695  & 0.694    & 0.694 & 0.706    & 0.706     & 0.706  & 0.706    & 0.706 \\
			A498                        & 0.673    & 0.678     & 0.673  & 0.671    & 0.673 & 0.701    & 0.701     & 0.701  & 0.701    & 0.701 & 0.695    & 0.697     & 0.695  & 0.694    & 0.695 & 0.707    & 0.716     & 0.707  & 0.704    & 0.707 \\
			A549\_ATCC                  & 0.678    & 0.68      & 0.678  & 0.678    & 0.679 & 0.708    & 0.707     & 0.708  & 0.707    & 0.707 & 0.699    & 0.699     & 0.699  & 0.699    & 0.699 & 0.713    & 0.713     & 0.713  & 0.713    & 0.712 \\
			ACHN                        & 0.66     & 0.667     & 0.66   & 0.657    & 0.661 & 0.686    & 0.687     & 0.686  & 0.685    & 0.686 & 0.675    & 0.678     & 0.675  & 0.673    & 0.675 & 0.683    & 0.69      & 0.683  & 0.68     & 0.683 \\
			BT\_549                     & 0.673    & 0.687     & 0.673  & 0.656    & 0.654 & 0.702    & 0.705     & 0.702  & 0.697    & 0.691 & 0.692    & 0.705     & 0.692  & 0.678    & 0.674 & 0.7      & 0.711     & 0.7    & 0.689    & 0.684 \\
			CAKI\_1                     & 0.669    & 0.676     & 0.669  & 0.665    & 0.667 & 0.691    & 0.691     & 0.691  & 0.691    & 0.691 & 0.687    & 0.689     & 0.687  & 0.686    & 0.686 & 0.701    & 0.702     & 0.701  & 0.7      & 0.7   \\
			DLD\_1                      & 0.713    & 0.707     & 0.713  & 0.682    & 0.628 & 0.71     & 0.698     & 0.71   & 0.695    & 0.647 & 0.706    & 0.704     & 0.706  & 0.666    & 0.612 & 0.719    & 0.709     & 0.719  & 0.702    & 0.653 \\
			DMS\_114                    & 0.675    & 0.674     & 0.675  & 0.673    & 0.669 & 0.675    & 0.675     & 0.675  & 0.672    & 0.666 & 0.671    & 0.673     & 0.671  & 0.663    & 0.658 & 0.68     & 0.682     & 0.68   & 0.681    & 0.68  \\
			DMS\_273                    & 0.687    & 0.685     & 0.687  & 0.671    & 0.649 & 0.712    & 0.711     & 0.712  & 0.701    & 0.679 & 0.672    & 0.681     & 0.672  & 0.638    & 0.619 & 0.704    & 0.7       & 0.704  & 0.698    & 0.68  \\
			\hline
			\hline
			Average   & 0.679    & 0.683     & 0.679  & 0.669    & 0.66  & 0.7      & 0.699     & 0.7    & 0.695    & 0.685 & 0.69     & 0.694     & 0.69   & 0.679    & 0.668 & 0.703    & 0.705     & 0.703  & 0.698    & 0.689 \\
			
		\end{tabular}%
	}
\end{table}


\begin{table}[]
	\centering
	\caption{Comparing REFINED CNN classifier with 6 other competing classifiers using McNemar's test to discover statistical significance in predicted classes across selected cell lines.}
	\label{Classification_statistical}
	\resizebox{\textwidth}{!}{%
		\begin{tabular}{c||c||c||c||c||c||c}
			Cell lines     & LR & RF & SVM & ANN & Random CNN & PCACNN \\
			\hline
			\hline
			786\_01        & 0            & 0            & 0             & 0             & 0                   & 0                \\
			A4981          & 0            & 0            & 0             & 0             & 0                   & 0                \\
			A549\_ATCC1    & 0            & 0            & 0             & 0             & 0                   & 0                \\
			ACHN1          & 0            & 0            & 0             & 0             & 0                   & 0                \\
			BT\_5491       & 5.1529E-231  & 7.0138E-192  & 1.5948E-204   & 5.6657E-219   & 1.0162E-261         & 2.0812E-258      \\
			CAKI\_11       & 0            & 0            & 0             & 0             & 0                   & 0                \\
			CCRF\_CEM1     & 9.5566E-299  & 2.4571E-237  & 1.0162E-261   & 0             & 0                   & 0                \\
			COLO\_2051     & 0            & 0            & 0             & 0             & 0                   & 0                \\
			DLD\_11        & 2.2132E-221  & 8.2447E-230  & 1.2882E-231   & 3.377E-226    & 4.2213E-227         & 1.0553E-227      \\
			DMS\_1141      & 7.2835E-158  & 9.5467E-153  & 1.7782E-161   & 6.2565E-148   & 5.4267E-166         & 5.5569E-163      \\
			DMS\_2731      & 5.7457E-188  & 3.5911E-189  & 3.4247E-195   & 3.0124E-182   & 7.8968E-177         & 1.9279E-180      \\
			DU\_1451       & 1.424E-306   & 1.0912E-252  & 2.152E-283    & 0             & 0                   & 0                \\
			EKVX1          & 0            & 0            & 0             & 0             & 0                   & 0                \\
			HCC\_29981     & 0            & 0            & 0             & 0             & 0                   & 0                \\
			MDA\_MB\_4351  & 2.2132E-221  & 7.1821E-189  & 1.6722E-198   & 1.2286E-237   & 1.9572E-295         & 4.203E-286       \\
			NCI\_ADR\_RES1 & 2.2347E-249  & 1.5209E-210  & 6.9161E-223   & 1.364E-253    & 8.2091E-289         & 2.1015E-286      \\
			SNB\_781       & 1.6489E-229  & 1.5725E-235  & 1.9196E-239   & 5.0321E-234   & 3.377E-226          & 1.0064E-233     
		\end{tabular}%
	}
\end{table}

\begin{table}[]
	\centering
	\caption{Predicting sensitivity or resistivity 95 \% intervals of the drugs using seven classifiers}
	\label{Classification_inervals}
	\resizebox{\textwidth}{!}{%
		\begin{tabular}{c||ccccc||ccccc||ccccc||ccccc||ccccc||ccccc||ccccc}
			\multirow{2}{*}{Cell lines} & \multicolumn{5}{c||}{Random CNN}  & \multicolumn{5}{c||}{REFIND CNN}                   & \multicolumn{5}{c||}{PCA CNN}  & \multicolumn{5}{c||}{LR} & \multicolumn{5}{c||}{RF}  & \multicolumn{5}{c||}{SVM}  & \multicolumn{5}{c}{ANN}                          \\
			& Accuracy & Precision & Recall & F1 Score & AUC   & Accuracy & Precision & Recall & F1 Score & AUC   & Accuracy & Precision & Recall & F1 Score & AUC   & Accuracy & Precision & Recall & F1 Score & AUC   & Accuracy & Precision & Recall & F1 Score & AUC   & Accuracy & Precision & Recall & F1 Score & AUC   & Accuracy & Precision & Recall & F1 Score & AUC   \\
			\hline
			\hline
			CCRF\_CEM                   & 0.038    & 0.038     & 0.038  & 0.038    & 0.039 & 0.037    & 0.037     & 0.037  & 0.037    & 0.038 & 0.039    & 0.039     & 0.039  & 0.039    & 0.039 & 0.041    & 0.040     & 0.041  & 0.041    & 0.041 & 0.040    & 0.040     & 0.040  & 0.040    & 0.040 & 0.040    & 0.040     & 0.040  & 0.040    & 0.041 & 0.039    & 0.039     & 0.039  & 0.040    & 0.040 \\
			COLO\_205                   & 0.037    & 0.037     & 0.037  & 0.037    & 0.037 & 0.037    & 0.037     & 0.037  & 0.037    & 0.037 & 0.038    & 0.038     & 0.038  & 0.038    & 0.038 & 0.040    & 0.040     & 0.040  & 0.040    & 0.040 & 0.039    & 0.039     & 0.039  & 0.039    & 0.039 & 0.040    & 0.040     & 0.040  & 0.040    & 0.040 & 0.039    & 0.039     & 0.039  & 0.039    & 0.039 \\
			DU\_145                     & 0.038    & 0.038     & 0.038  & 0.038    & 0.038 & 0.037    & 0.037     & 0.037  & 0.037    & 0.037 & 0.039    & 0.039     & 0.039  & 0.039    & 0.039 & 0.040    & 0.040     & 0.040  & 0.041    & 0.041 & 0.039    & 0.039     & 0.039  & 0.039    & 0.039 & 0.040    & 0.040     & 0.040  & 0.040    & 0.040 & 0.039    & 0.039     & 0.039  & 0.039    & 0.039 \\
			EKVX                        & 0.039    & 0.039     & 0.039  & 0.039    & 0.039 & 0.037    & 0.037     & 0.037  & 0.037    & 0.037 & 0.039    & 0.039     & 0.039  & 0.039    & 0.039 & 0.041    & 0.041     & 0.041  & 0.041    & 0.041 & 0.040    & 0.040     & 0.040  & 0.040    & 0.040 & 0.041    & 0.041     & 0.041  & 0.041    & 0.041 & 0.040    & 0.040     & 0.040  & 0.040    & 0.040 \\
			HCC\_2998                   & 0.039    & 0.039     & 0.039  & 0.039    & 0.039 & 0.037    & 0.037     & 0.037  & 0.037    & 0.037 & 0.039    & 0.039     & 0.039  & 0.039    & 0.039 & 0.041    & 0.041     & 0.041  & 0.041    & 0.041 & 0.040    & 0.040     & 0.040  & 0.040    & 0.040 & 0.040    & 0.040     & 0.040  & 0.040    & 0.040 & 0.039    & 0.039     & 0.039  & 0.039    & 0.039 \\
			MDA\_MB\_435                & 0.039    & 0.039     & 0.039  & 0.039    & 0.040 & 0.037    & 0.037     & 0.037  & 0.037    & 0.038 & 0.040    & 0.040     & 0.040  & 0.040    & 0.040 & 0.040    & 0.040     & 0.040  & 0.041    & 0.041 & 0.039    & 0.039     & 0.039  & 0.040    & 0.040 & 0.040    & 0.039     & 0.040  & 0.040    & 0.041 & 0.039    & 0.039     & 0.039  & 0.039    & 0.040 \\
			SNB\_78                     & 0.038    & 0.038     & 0.038  & 0.039    & 0.040 & 0.036    & 0.036     & 0.036  & 0.036    & 0.038 & 0.037    & 0.037     & 0.037  & 0.038    & 0.040 & 0.038    & 0.038     & 0.038  & 0.039    & 0.041 & 0.037    & 0.037     & 0.037  & 0.038    & 0.040 & 0.038    & 0.037     & 0.038  & 0.040    & 0.042 & 0.038    & 0.038     & 0.038  & 0.039    & 0.041 \\
			NCI\_ADR\_RES               & 0.039    & 0.039     & 0.039  & 0.039    & 0.039 & 0.036    & 0.036     & 0.036  & 0.036    & 0.036 & 0.039    & 0.039     & 0.039  & 0.039    & 0.039 & 0.040    & 0.040     & 0.040  & 0.040    & 0.040 & 0.039    & 0.039     & 0.039  & 0.039    & 0.040 & 0.039    & 0.039     & 0.039  & 0.040    & 0.040 & 0.039    & 0.039     & 0.039  & 0.039    & 0.039 \\
			786\_0                      & 0.039    & 0.039     & 0.039  & 0.039    & 0.039 & 0.036    & 0.036     & 0.036  & 0.036    & 0.036 & 0.039    & 0.039     & 0.039  & 0.039    & 0.039 & 0.040    & 0.040     & 0.040  & 0.040    & 0.040 & 0.039    & 0.039     & 0.039  & 0.039    & 0.039 & 0.040    & 0.039     & 0.040  & 0.040    & 0.040 & 0.039    & 0.039     & 0.039  & 0.039    & 0.039 \\
			A498                        & 0.039    & 0.038     & 0.039  & 0.039    & 0.039 & 0.037    & 0.037     & 0.037  & 0.037    & 0.037 & 0.039    & 0.039     & 0.039  & 0.039    & 0.039 & 0.040    & 0.040     & 0.040  & 0.040    & 0.040 & 0.039    & 0.039     & 0.039  & 0.039    & 0.039 & 0.040    & 0.039     & 0.040  & 0.040    & 0.040 & 0.039    & 0.039     & 0.039  & 0.039    & 0.039 \\
			A549\_ATCC                  & 0.039    & 0.039     & 0.039  & 0.039    & 0.039 & 0.036    & 0.036     & 0.036  & 0.036    & 0.036 & 0.039    & 0.039     & 0.039  & 0.039    & 0.039 & 0.040    & 0.040     & 0.040  & 0.040    & 0.040 & 0.039    & 0.039     & 0.039  & 0.039    & 0.039 & 0.039    & 0.039     & 0.039  & 0.039    & 0.039 & 0.039    & 0.039     & 0.039  & 0.039    & 0.039 \\
			ACHN                        & 0.039    & 0.039     & 0.039  & 0.039    & 0.039 & 0.037    & 0.037     & 0.037  & 0.037    & 0.037 & 0.039    & 0.039     & 0.039  & 0.039    & 0.039 & 0.041    & 0.041     & 0.041  & 0.041    & 0.041 & 0.040    & 0.040     & 0.040  & 0.040    & 0.040 & 0.040    & 0.040     & 0.040  & 0.040    & 0.040 & 0.040    & 0.040     & 0.040  & 0.040    & 0.040 \\
			BT\_549                     & 0.039    & 0.039     & 0.039  & 0.039    & 0.039 & 0.037    & 0.037     & 0.037  & 0.037    & 0.038 & 0.039    & 0.039     & 0.039  & 0.039    & 0.039 & 0.040    & 0.040     & 0.040  & 0.041    & 0.041 & 0.039    & 0.039     & 0.039  & 0.039    & 0.040 & 0.040    & 0.039     & 0.040  & 0.040    & 0.040 & 0.039    & 0.039     & 0.039  & 0.040    & 0.040 \\
			CAKI\_1                     & 0.039    & 0.039     & 0.039  & 0.039    & 0.039 & 0.037    & 0.037     & 0.037  & 0.037    & 0.037 & 0.039    & 0.039     & 0.039  & 0.039    & 0.039 & 0.040    & 0.040     & 0.040  & 0.041    & 0.041 & 0.040    & 0.040     & 0.040  & 0.040    & 0.040 & 0.040    & 0.040     & 0.040  & 0.040    & 0.040 & 0.039    & 0.039     & 0.039  & 0.039    & 0.039 \\
			DLD\_1                      & 0.039    & 0.039     & 0.039  & 0.040    & 0.041 & 0.036    & 0.036     & 0.036  & 0.036    & 0.037 & 0.038    & 0.038     & 0.038  & 0.039    & 0.041 & 0.039    & 0.039     & 0.039  & 0.040    & 0.042 & 0.039    & 0.039     & 0.039  & 0.040    & 0.041 & 0.039    & 0.039     & 0.039  & 0.041    & 0.042 & 0.039    & 0.039     & 0.039  & 0.039    & 0.041 \\
			DMS\_114                    & 0.040    & 0.040     & 0.040  & 0.040    & 0.040 & 0.038    & 0.038     & 0.038  & 0.038    & 0.038 & 0.040    & 0.040     & 0.040  & 0.040    & 0.040 & 0.040    & 0.040     & 0.040  & 0.040    & 0.040 & 0.040    & 0.040     & 0.040  & 0.040    & 0.041 & 0.040    & 0.040     & 0.040  & 0.041    & 0.041 & 0.040    & 0.040     & 0.040  & 0.040    & 0.040 \\
			DMS\_273                    & 0.039    & 0.039     & 0.039  & 0.039    & 0.040 & 0.039    & 0.039     & 0.039  & 0.040    & 0.040 & 0.037    & 0.037     & 0.037  & 0.037    & 0.037 & 0.040    & 0.040     & 0.040  & 0.040    & 0.041 & 0.039    & 0.039     & 0.039  & 0.039    & 0.040 & 0.040    & 0.040     & 0.040  & 0.041    & 0.042 & 0.039    & 0.039     & 0.039  & 0.039    & 0.040 \\
			\hline
			\hline
			Average                     & 0.039    & 0.039     & 0.039  & 0.039    & 0.039 & 0.037    & 0.037     & 0.037  & 0.037    & 0.037 & 0.039    & 0.039     & 0.039  & 0.039    & 0.039 & 0.040    & 0.040     & 0.040  & 0.040    & 0.041 & 0.039    & 0.039     & 0.039  & 0.040    & 0.040 & 0.040    & 0.040     & 0.040  & 0.040    & 0.040 & 0.039    & 0.039     & 0.039  & 0.039    & 0.040
		\end{tabular}%
	}
\end{table}

\begin{table}[!htbp]\caption{Drug sensitivity prediction using seven regression models. The NRMSE, PCC, and bias of each model is used for comparison.}\label{Reression}
	\centering
	\resizebox{\textwidth}{!}{%
		\begin{tabular}{c||ccc||ccc||ccc||ccc||ccc||ccc||ccc}
			\multirow{2}{*}{Cell lines} & \multicolumn{3}{c||}{Random CNN} & \multicolumn{3}{c||}{PCA CNN} & \multicolumn{3}{c||}{REFINED CNN}                  & \multicolumn{3}{c||}{RF} & \multicolumn{3}{c||}{SVR} & \multicolumn{3}{c||}{ANN} & \multicolumn{3}{c}{EN} \\
			& NRMSE    & PCC      & Bias     & NRMSE   & PCC     & Bias    & NRMSE          & PCC            & Bias           & NRMSE  & PCC   & Bias  & NRMSE  & PCC    & Bias  & NRMSE  & PCC    & Bias  & NRMSE  & PCC   & Bias  \\
			\hline
			\hline
			CCRF\_CEM                   & 0.868    & 0.536    & 0.818    & 0.884   & 0.529   & 0.671   & \textbf{0.774} & \textbf{0.653} & \textbf{0.493} & 0.893  & 0.465 & 0.839 & 0.874  & 0.521  & 0.750 & 0.828  & 0.563  & 0.705 & 0.978  & 0.259 & 0.973 \\
			COLO\_205                   & 0.958    & 0.538    & 0.709    & 0.823   & 0.601   & 0.564   & \textbf{0.741} & \textbf{0.686} & \textbf{0.448} & 0.892  & 0.467 & 0.837 & 0.867  & 0.535  & 0.746 & 0.811  & 0.587  & 0.680 & 0.974  & 0.288 & 0.968 \\
			DU\_145                     & 0.839    & 0.572    & 0.594    & 0.882   & 0.555   & 0.553   & \textbf{0.786} & \textbf{0.647} & \textbf{0.458} & 0.903  & 0.434 & 0.838 & 0.882  & 0.507  & 0.773 & 0.842  & 0.539  & 0.713 & 0.976  & 0.268 & 0.970 \\
			EKVX   & 0.867    & 0.553    & 0.579    & 0.894   & 0.572   & 0.514   & \textbf{0.804} & \textbf{0.618} & \textbf{0.535} & 0.904  & 0.433 & 0.842 & 0.881  & 0.503  & 0.769 & 0.848  & 0.530  & 0.713 & 0.978  & 0.252 & 0.972 \\
			HCC\_2998                   & 0.930    & 0.578    & 0.744    & 0.961   & 0.535   & 0.628   & \textbf{0.774} & \textbf{0.654} & \textbf{0.447} & 0.880  & 0.488 & 0.815 & 0.858  & 0.542  & 0.740 & 0.820  & 0.572  & 0.662 & 0.968  & 0.312 & 0.961 \\
			MDA\_MB\_435                & 0.884    & 0.532    & 0.760    & 0.982   & 0.557   & 0.518   & \textbf{0.787} & \textbf{0.651} & \textbf{0.469} & 0.912  & 0.412 & 0.849 & 0.897  & 0.477  & 0.781 & 0.858  & 0.514  & 0.732 & 0.982  & 0.234 & 0.977 \\
			SNB\_78                     & 0.864    & 0.516    & 0.730    & 0.917   & 0.544   & 0.528   & \textbf{0.784} & \textbf{0.652} & \textbf{0.448} & 0.842  & 0.558 & 0.769 & 0.847  & 0.555  & 0.745 & 0.822  & 0.573  & 0.648 & 0.964  & 0.352 & 0.958 \\
			NCI\_ADR\_RES               & 0.945    & 0.470    & 0.751    & 0.834   & 0.580   & 0.565   & \textbf{0.798} & \textbf{0.638} & \textbf{0.475} & 0.908  & 0.434 & 0.860 & 0.903  & 0.476  & 0.799 & 0.857  & 0.518  & 0.737 & 0.984  & 0.275 & 0.981 \\
			786\_0                      & 0.877    & 0.558    & 0.604    & 0.944   & 0.599   & 0.543   & \textbf{0.752} & \textbf{0.665} & \textbf{0.450} & 0.887  & 0.481 & 0.832 & 0.878  & 0.521  & 0.767 & 0.832  & 0.571  & 0.707 & 0.974  & 0.289 & 0.968 \\
			A498   & 0.845    & 0.604    & 0.712    & 0.883   & 0.576   & 0.545   & \textbf{0.785} & \textbf{0.635} & \textbf{0.433} & 0.890  & 0.465 & 0.827 & 0.859  & 0.551  & 0.750 & 0.840  & 0.567  & 0.700 & 0.972  & 0.312 & 0.967 \\
			A549\_ATCC                  & 0.913    & 0.548    & 0.560    & 0.814   & 0.590   & 0.623   & \textbf{0.769} & \textbf{0.645} & \textbf{0.536} & 0.870  & 0.511 & 0.806 & 0.857  & 0.555  & 0.743 & 0.810  & 0.592  & 0.692 & 0.969  & 0.308 & 0.962 \\
			ACHN                        & 0.830    & 0.571    & 0.616    & 0.837   & 0.562   & 0.732   & \textbf{0.747} & \textbf{0.665} & \textbf{0.560} & 0.880  & 0.494 & 0.822 & 0.868  & 0.529  & 0.759 & 0.816  & 0.585  & 0.675 & 0.975  & 0.284 & 0.970 \\
			BT\_549                     & 0.941    & 0.529    & 0.595    & 0.885   & 0.545   & 0.769   & \textbf{0.843} & \textbf{0.584} & \textbf{0.560} & 0.888  & 0.478 & 0.832 & 0.870  & 0.514  & 0.755 & 0.840  & 0.545  & 0.714 & 0.977  & 0.279 & 0.972 \\
			CAKI\_1   & 0.866    & 0.561    & 0.589    & 0.886   & 0.475   & 0.753   & \textbf{0.775} & \textbf{0.640} & \textbf{0.423} & 0.901  & 0.444 & 0.845 & 0.885  & 0.507  & 0.773 & 0.863  & 0.537  & 0.692 & 0.982  & 0.241 & 0.977 \\
			DLD\_1                      & 0.923    & 0.620    & 0.611    & 0.812   & 0.584   & 0.666   & \textbf{0.781} & \textbf{0.625} & \textbf{0.499} & 0.847  & 0.557 & 0.783 & 0.867  & 0.529  & 0.763 & 0.824  & 0.568  & 0.667 & 0.975  & 0.270 & 0.968 \\
			DMS\_114                    & 0.873    & 0.546    & 0.575    & 0.953   & 0.565   & 0.487   & \textbf{0.738} & \textbf{0.680} & \textbf{0.484} & 0.832  & 0.568 & 0.746 & 0.834  & 0.571  & 0.689 & 0.804  & 0.596  & 0.631 & 0.985  & 0.332 & 0.958 \\
			DMS\_273                    & 0.810    & 0.587    & 0.652    & 0.909   & 0.586   & 0.619   & \textbf{0.758} & \textbf{0.670} & \textbf{0.454} & 0.829  & 0.568 & 0.735 & 0.860  & 0.538  & 0.728 & 0.817  & 0.578  & 0.645 & 0.984  & 0.318 & 0.955 \\
			\hline
			\hline
			Average   & 0.884    & 0.554    & 0.659    & 0.888   & 0.562   & 0.605   & \textbf{0.776} & \textbf{0.647} & \textbf{0.481} & 0.880  & 0.486 & 0.816 & 0.870  & 0.525  & 0.755 & 0.831  & 0.561  & 0.689 & 0.976  & 0.287 & 0.968 \\
			
		\end{tabular}%
	}
\end{table}

\begin{table}[]
	\centering
	\caption{Robustness analysis to compare REFINED CNN with 6 other competing models per each metric and each cell line. Each cell of the table represents the percentage for which REFINED CNN outperforms the paired competing model.}
	\label{NCI_Robustness}
	\resizebox{\textwidth}{!}{%
		\begin{tabular}{c||ccc||ccc||ccc||ccc||ccc||ccc}
			\multirow{2}{*}{Cell lines} & \multicolumn{3}{c||}{REFIEND CNN Versus PCA CNN}      & \multicolumn{3}{c||}{REFIEND CNN Versus Random CNN}   & \multicolumn{3}{c||}{REFIEND CNN Versus ANN}          & \multicolumn{3}{c||}{REFIEND CNN Versus SVR}          & \multicolumn{3}{c||}{REFIEND CNN Versus RF}           & \multicolumn{3}{c}{REFIEND CNN Versus EN}           \\
			& NRMSE Percentage & PCC Percentage & Bias Percentage & NRMSE Percentage & PCC Percentage & Bias Percentage & NRMSE Percentage & PCC Percentage & Bias Percentage & NRMSE Percentage & PCC Percentage & Bias Percentage & NRMSE Percentage & PCC Percentage & Bias Percentage & NRMSE Percentage & PCC Percentage & Bias Percentage \\
			\hline
			\hline
			786\_0                      & 100              & 94.3           & 81.5            & 99.9             & 99.7           & 95.2            & 99.3             & 98.7           & 100             & 100              & 100            & 100             & 100              & 100            & 100             & 100              & 100            & 100             \\
			A498                        & 99.8             & 91.8           & 95.4            & 100              & 100            & 100             & 96.8             & 96.1           & 99.8            & 99.5             & 98.1           & 100             & 100              & 100            & 100             & 100              & 100            & 100             \\
			A549\_ATCC                  & 88.8             & 90.3           & 98.4            & 100              & 98.1           & 72.9            & 87.7             & 88.6           & 100             & 99.6             & 98.3           & 100             & 99.8             & 99.9           & 100             & 100              & 100            & 100             \\
			ACHN                        & 99.8             & 99             & 100             & 99.2             & 98.9           & 93.7            & 98.8             & 97.9           & 99.7            & 100              & 99.8           & 100             & 100              & 100            & 100             & 100              & 100            & 100             \\
			BT\_549                     & 88.3             & 80.6           & 100             & 98.9             & 91.9           & 81.8            & 45               & 83.6           & 100             & 80.2             & 95.2           & 100             & 91.2             & 99.2           & 100             & 100              & 100            & 100             \\
			CAKI\_1                     & 99.8             & 100            & 100             & 100              & 95.2           & 100             & 99.5             & 99.5           & 100             & 100              & 99.9           & 100             & 100              & 100            & 100             & 100              & 100            & 100             \\
			DLD\_1                      & 78.1             & 78.1           & 88.4            & 100              & 98.8           & 100             & 91.5             & 90.8           & 99.4            & 99.9             & 98.7           & 100             & 100              & 100            & 100             & 100              & 100            & 100             \\
			DMS\_114                    & 100              & 99.3           & 51.8            & 100              & 100            & 100             & 98.8             & 99.1           & 100             & 100              & 100            & 100             & 100              & 100            & 100             & 100              & 100            & 100             \\
			DMS\_273                    & 100              & 97.6           & 100             & 99.9             & 87.4           & 82              & 94.3             & 99.2           & 100             & 99.6             & 99.8           & 100             & 98.9             & 98.9           & 100             & 100              & 100            & 100             \\
			HCC\_2998                   & 100              & 99.7           & 68.7            & 99.5             & 98.9           & 95.3            & 79.2             & 89.7           & 82.4            & 96.1             & 96.9           & 99.4            & 96.8             & 97.2           & 100             & 100              & 100            & 100             \\
			CCRF\_CEM                   & 100              & 99.9           & 100             & 84               & 94             & 100             & 88.5             & 95.5           & 100             & 99.8             & 99.6           & 100             & 94.8             & 97.1           & 100             & 100              & 100            & 100             \\
			COLO\_205                   & 95.1             & 93.7           & 99.8            & 99.4             & 99.6           & 100             & 99.8             & 100            & 100             & 100              & 100            & 100             & 100              & 100            & 100             & 100              & 100            & 100             \\
			DU\_145                     & 98.8             & 99             & 98.8            & 100              & 100            & 100             & 96.4             & 99.9           & 100             & 99.8             & 100            & 100             & 99.8             & 100            & 100             & 100              & 100            & 100             \\
			EKVX                        & 96.9             & 85             & 91.6            & 100              & 80.1           & 100             & 67.3             & 86.7           & 100             & 93.1             & 96.2           & 100             & 98.6             & 99.8           & 100             & 100              & 100            & 100             \\
			MDA\_MB\_435                & 100              & 99.9           & 87.7            & 99.8             & 100            & 100             & 98.9             & 100            & 100             & 99.9             & 100            & 100             & 100              & 100            & 100             & 100              & 100            & 100             \\
			NCI\_ADR\_RES               & 98               & 98.9           & 98.7            & 100              & 100            & 100             & 97.1             & 100            & 100             & 100              & 100            & 100             & 100              & 100            & 100             & 100              & 100            & 100             \\
			SNB\_78                     & 99.3             & 98             & 88.5            & 96.7             & 99.4           & 100             & 82.7             & 94.5           & 100             & 92               & 97             & 100             & 91.3             & 96.8           & 100             & 100              & 100            & 100             \\
			\hline
			\hline
			Average                     & 96.63            & 94.42          & 91.14           & 98.66            & 96.59          & 95.35           & 89.51            & 95.28          & 98.90           & 97.62            & 98.79          & 99.96           & 98.31            & 99.35          & 100.00          & 100.00           & 100.00         & 100.00         
		\end{tabular}%
	}
\end{table}

\begin{table}[]
	\centering
	\caption{Gap statistics analysis to compare REFINED CNN with 6 other competing models per each metric and each cell line paired with the null model. The wider (larger) Gap value indicates better performance.}
	\label{NCI_Gap}
	\resizebox{\textwidth}{!}{%
		\begin{tabular}{c||ccc||ccc||ccc||ccc||ccc||ccc||ccc}
			\multirow{2}{*}{Cell lines} & \multicolumn{3}{c||}{REFINED CNN Gap} & \multicolumn{3}{c||}{PCA CNN Gap} & \multicolumn{3}{c||}{Random CNN Gap} & \multicolumn{3}{c||}{RF Gap} & \multicolumn{3}{c||}{SVR Gap} & \multicolumn{3}{c||}{ANN Gap} & \multicolumn{3}{c}{EN Gap} \\
			& NRMSE      & PCC        & Bias      & NRMSE     & PCC      & Bias     & NRMSE      & PCC       & Bias      & NRMSE   & PCC     & Bias   & NRMSE   & PCC     & Bias    & NRMSE   & PCC     & Bias    & NRMSE   & PCC     & Bias   \\
			\hline
			\hline
			786\_0                      & 0.735      & 0.661      & 0.468     & 0.544     & 0.593    & 0.565    & 0.604      & 0.554     & 0.407     & 0.598   & 0.481   & 0.407  & 0.611   & 0.520   & 0.241   & 0.652   & 0.566   & 0.301   & 0.478   & 0.095   & 0.145  \\
			A498                        & 0.699      & 0.630      & 0.400     & 0.597     & 0.570    & 0.468    & 0.501      & 0.469     & 0.588     & 0.592   & 0.462   & 0.588  & 0.628   & 0.549   & 0.260   & 0.640   & 0.562   & 0.309   & 0.477   & 0.327   & 0.039  \\
			A549\_ATCC                  & 0.716      & 0.641      & 0.479     & 0.668     & 0.585    & 0.389    & 0.564      & 0.541     & 0.453     & 0.615   & 0.507   & 0.453  & 0.631   & 0.552   & 0.265   & 0.676   & 0.588   & 0.319   & 0.477   & 0.179   & 0.037  \\
			ACHN                        & 0.730      & 0.658      & 0.450     & 0.641     & 0.556    & 0.274    & 0.647      & 0.567     & 0.396     & 0.596   & 0.490   & 0.396  & 0.612   & 0.525   & 0.250   & 0.661   & 0.580   & 0.334   & 0.471   & 0.222   & 0.036  \\
			BT\_549                     & 0.806      & 0.581      & 0.454     & 0.616     & 0.548    & 0.235    & 0.555      & 0.529     & 0.422     & 0.613   & 0.477   & 0.422  & 0.633   & 0.514   & 0.253   & 0.660   & 0.541   & 0.294   & 0.495   & 0.283   & 0.029  \\
			CAKI\_1                     & 0.708      & 0.635      & 0.483     & 0.597     & 0.476    & 0.254    & 0.595      & 0.571     & 0.591     & 0.583   & 0.445   & 0.591  & 0.602   & 0.506   & 0.234   & 0.618   & 0.534   & 0.315   & 0.477   & 0.222   & 0.140  \\
			DLD\_1                      & 0.710      & 0.622      & 0.403     & 0.676     & 0.580    & 0.348    & 0.609      & 0.522     & 0.190     & 0.594   & 0.462   & 0.190  & 0.617   & 0.519   & 0.256   & 0.663   & 0.558   & 0.301   & 0.476   & 0.320   & 0.143  \\
			DMS\_114                    & 0.764      & 0.673      & 0.530     & 0.555     & 0.562    & 0.535    & 0.633      & 0.513     & 0.300     & 0.608   & 0.464   & 0.300  & 0.638   & 0.533   & 0.263   & 0.693   & 0.583   & 0.327   & 0.492   & 0.288   & 0.168  \\
			DMS\_273                    & 0.725      & 0.667      & 0.561     & 0.563     & 0.581    & 0.397    & 0.557      & 0.610     & 0.513     & 0.631   & 0.555   & 0.513  & 0.618   & 0.530   & 0.251   & 0.656   & 0.565   & 0.348   & 0.470   & 0.331   & 0.029  \\
			HCC\_2998                   & 0.713      & 0.648      & 0.417     & 0.521     & 0.529    & 0.401    & 0.607      & 0.540     & 0.441     & 0.654   & 0.566   & 0.441  & 0.655   & 0.569   & 0.326   & 0.681   & 0.591   & 0.384   & 0.481   & 0.077   & 0.029  \\
			RF\_CEM                     & 0.711      & 0.649      & 0.521     & 0.601     & 0.529    & 0.341    & 0.672      & 0.582     & 0.361     & 0.653   & 0.568   & 0.361  & 0.628   & 0.544   & 0.285   & 0.666   & 0.576   & 0.370   & 0.479   & 0.244   & 0.134  \\
			OLO\_205                    & 0.727      & 0.681      & 0.562     & 0.661     & 0.621    & 0.446    & 0.625      & 0.567     & 0.417     & 0.564   & 0.437   & 0.417  & 0.588   & 0.507   & 0.233   & 0.625   & 0.536   & 0.294   & 0.461   & 0.243   & 0.132  \\
			DU\_145                     & 0.708      & 0.643      & 0.556     & 0.612     & 0.551    & 0.462    & 0.605      & 0.470     & 0.153     & 0.591   & 0.435   & 0.153  & 0.618   & 0.499   & 0.240   & 0.647   & 0.526   & 0.296   & 0.490   & 0.318   & 0.150  \\
			EKVX                        & 0.684      & 0.613      & 0.475     & 0.591     & 0.567    & 0.501    & 0.557      & 0.579     & 0.258     & 0.606   & 0.481   & 0.258  & 0.634   & 0.538   & 0.266   & 0.668   & 0.567   & 0.344   & 0.481   & 0.296   & 0.140  \\
			MDA\_MB\_435                & 0.699      & 0.647      & 0.545     & 0.403     & 0.509    & 0.496    & 0.604      & 0.531     & 0.245     & 0.572   & 0.409   & 0.245  & 0.593   & 0.476   & 0.228   & 0.627   & 0.511   & 0.274   & 0.481   & 0.329   & 0.138  \\
			NCI\_ADR\_RES               & 0.701      & 0.633      & 0.537     & 0.621     & 0.541    & 0.448    & 0.554      & 0.472     & 0.259     & 0.593   & 0.434   & 0.259  & 0.602   & 0.476   & 0.210   & 0.644   & 0.516   & 0.269   & 0.494   & 0.428   & 0.147  \\
			SNB\_78                     & 0.716      & 0.649      & 0.575     & 0.568     & 0.539    & 0.496    & 0.633      & 0.526     & 0.297     & 0.653   & 0.554   & 0.297  & 0.656   & 0.556   & 0.276   & 0.672   & 0.568   & 0.371   & 0.479   & 0.282   & 0.183  \\
			\hline
			\hline
			Average                     & 0.721      & 0.643      & 0.495     & 0.590     & 0.555    & 0.415    & 0.596      & 0.538     & 0.370     & 0.607   & 0.484   & 0.370  & 0.622   & 0.524   & 0.255   & 0.656   & 0.557   & 0.321   & 0.480   & 0.264   & 0.107 
		\end{tabular}%
	}
\end{table}

\begin{figure*}
	\begin{multicols}{2}
		\includegraphics[width=\linewidth,height=0.25\textheight]{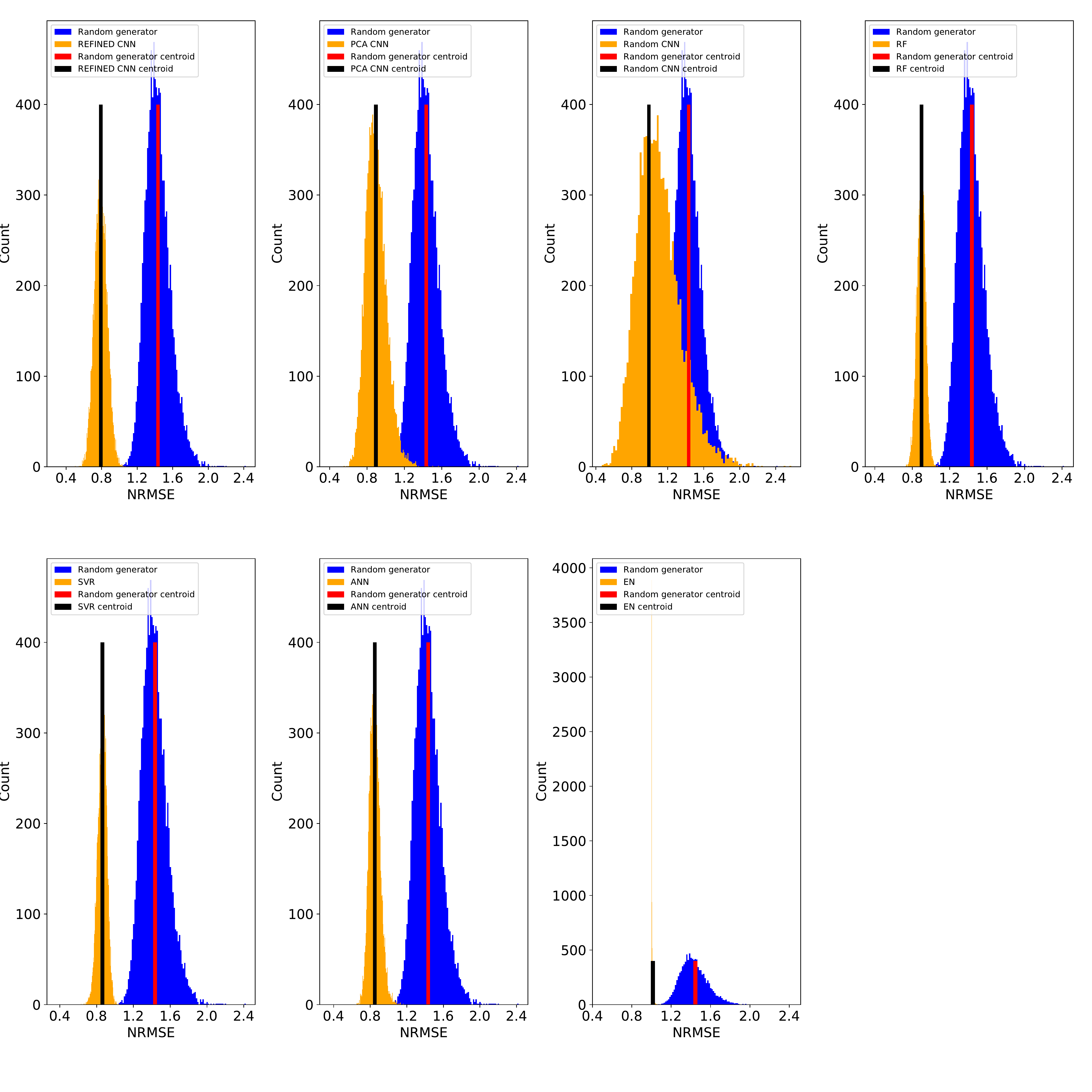}\par 
		\subcaption{NRMSEs of A498 cell line}
		\includegraphics[width=\linewidth,height=0.25\textheight]{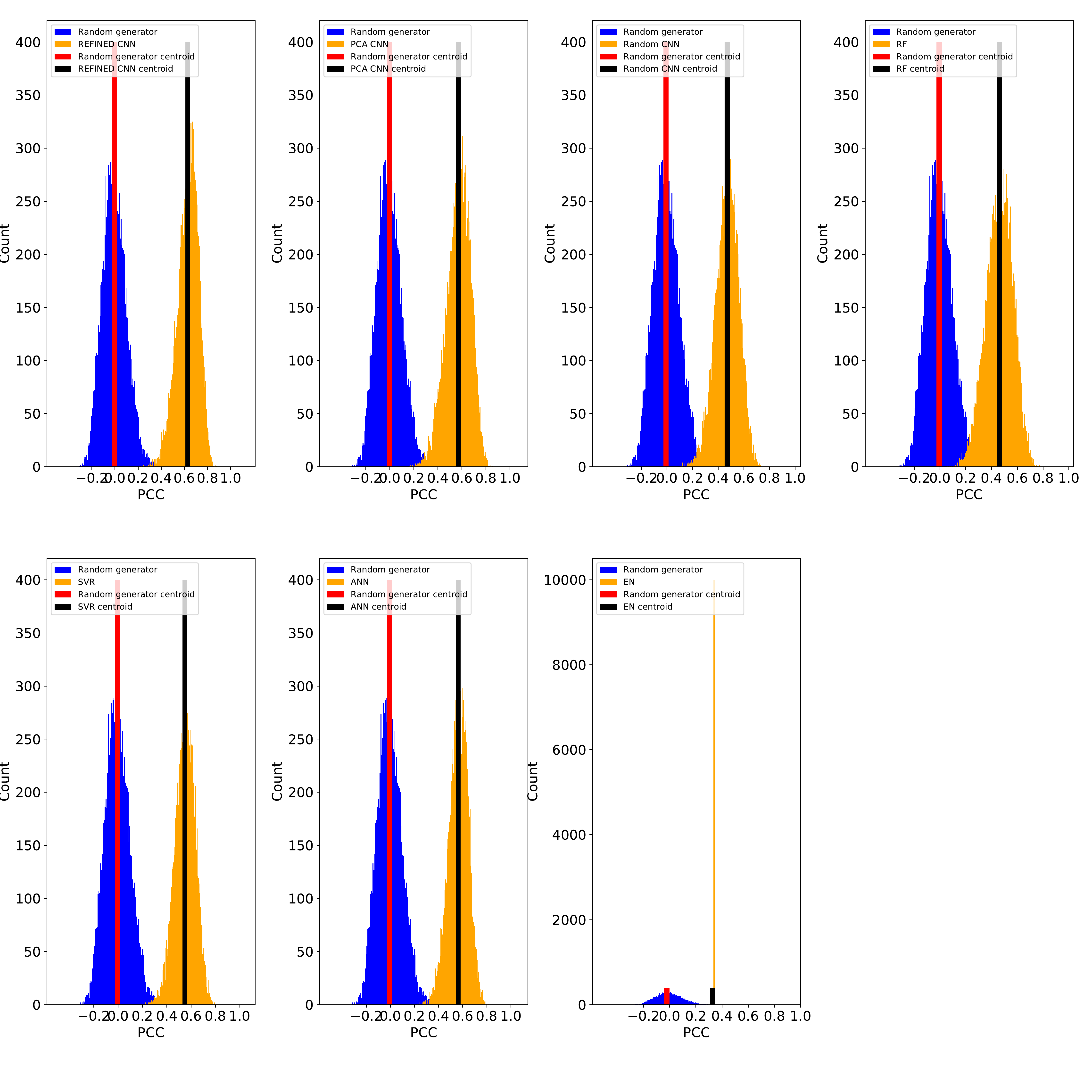}\par 
		\subcaption{PCCs of A498 cell line}
	\end{multicols}
	\begin{multicols}{2}
		\includegraphics[width=\linewidth,height=0.25\textheight]{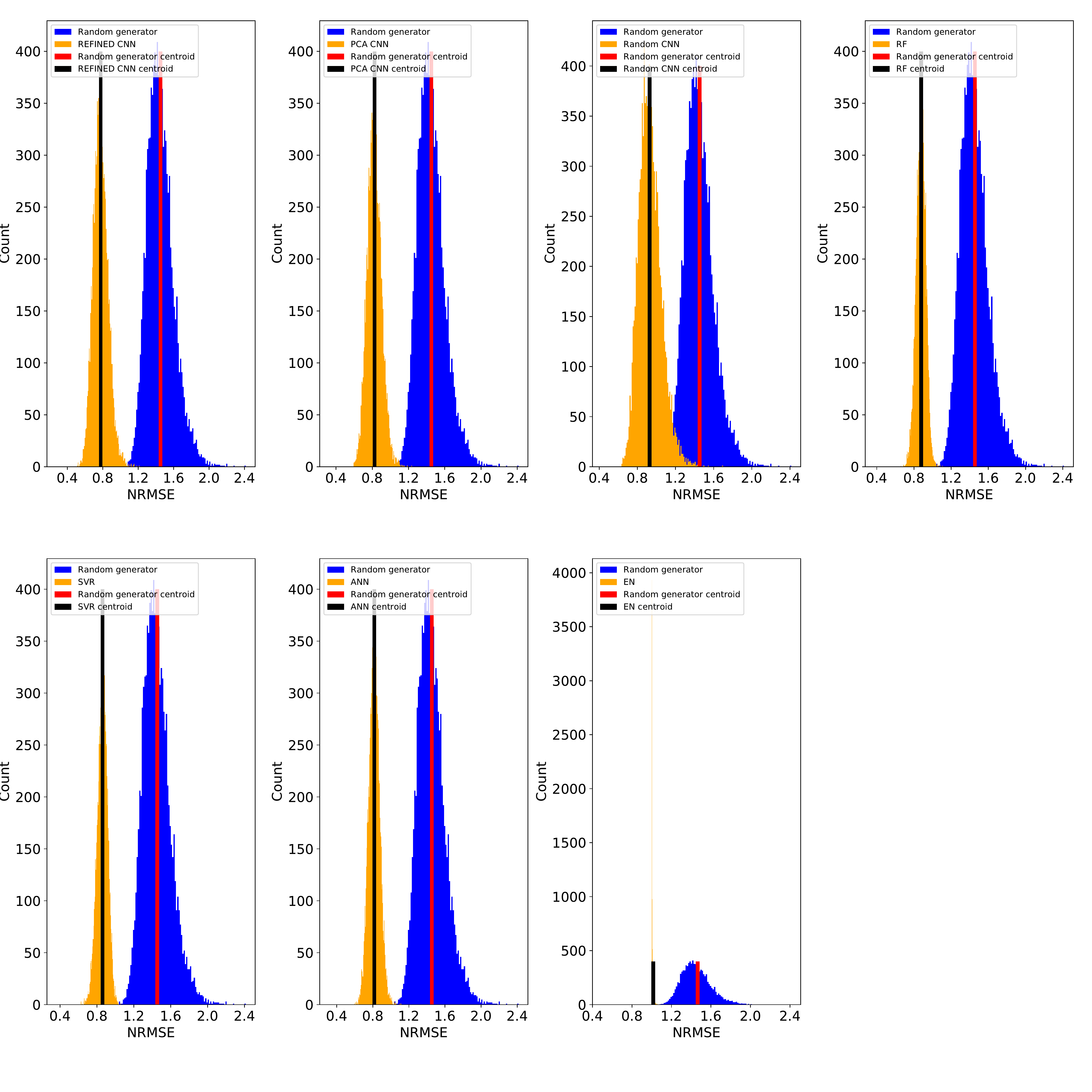}\par
		\subcaption{NRMSEs of A549\_ATCC cell line}
		\includegraphics[width=\linewidth,height=0.25\textheight]{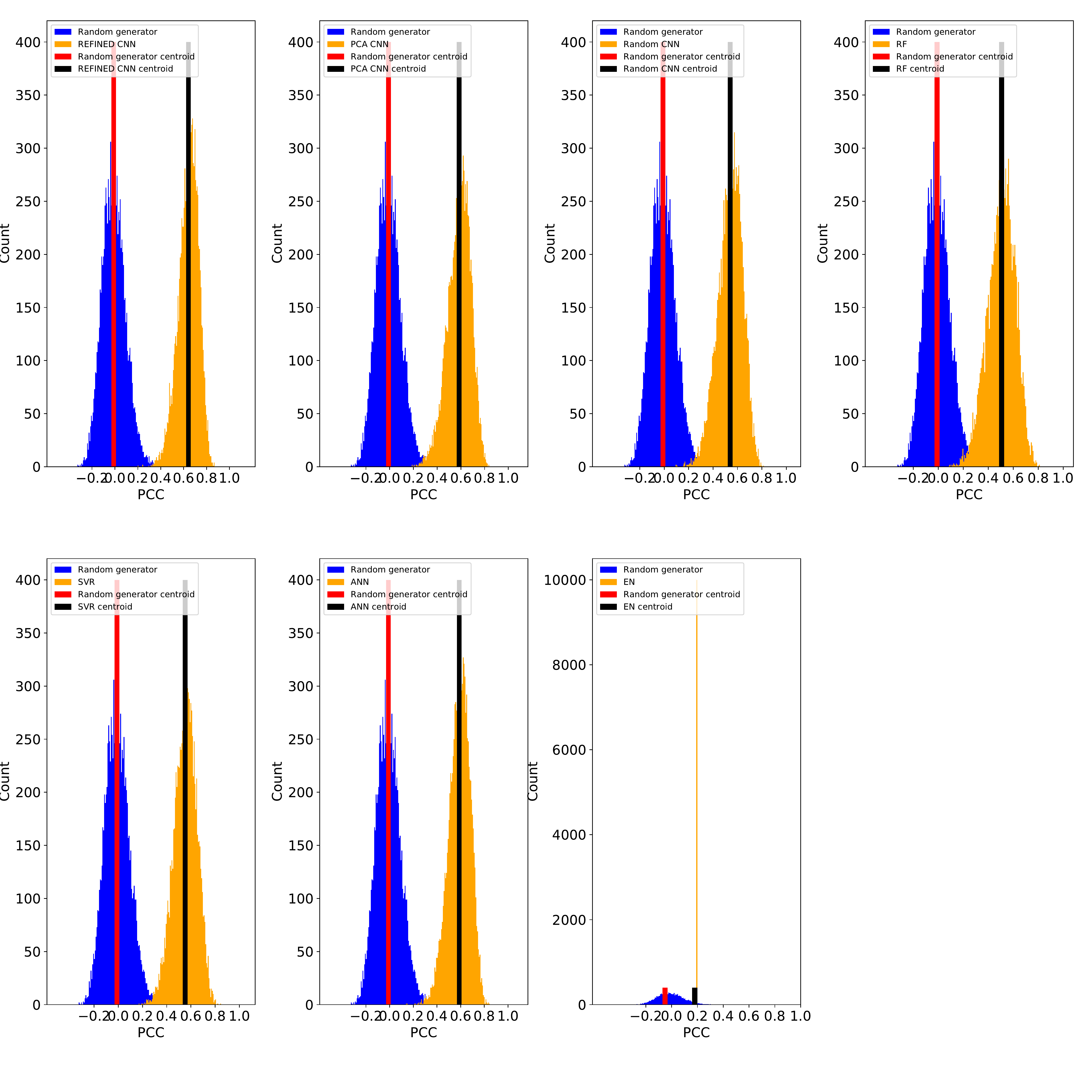}\par
		\subcaption{PCCs of A549\_ATCC cell line}
	\end{multicols}
	\begin{multicols}{2}
		\includegraphics[width=\linewidth,height=0.25\textheight]{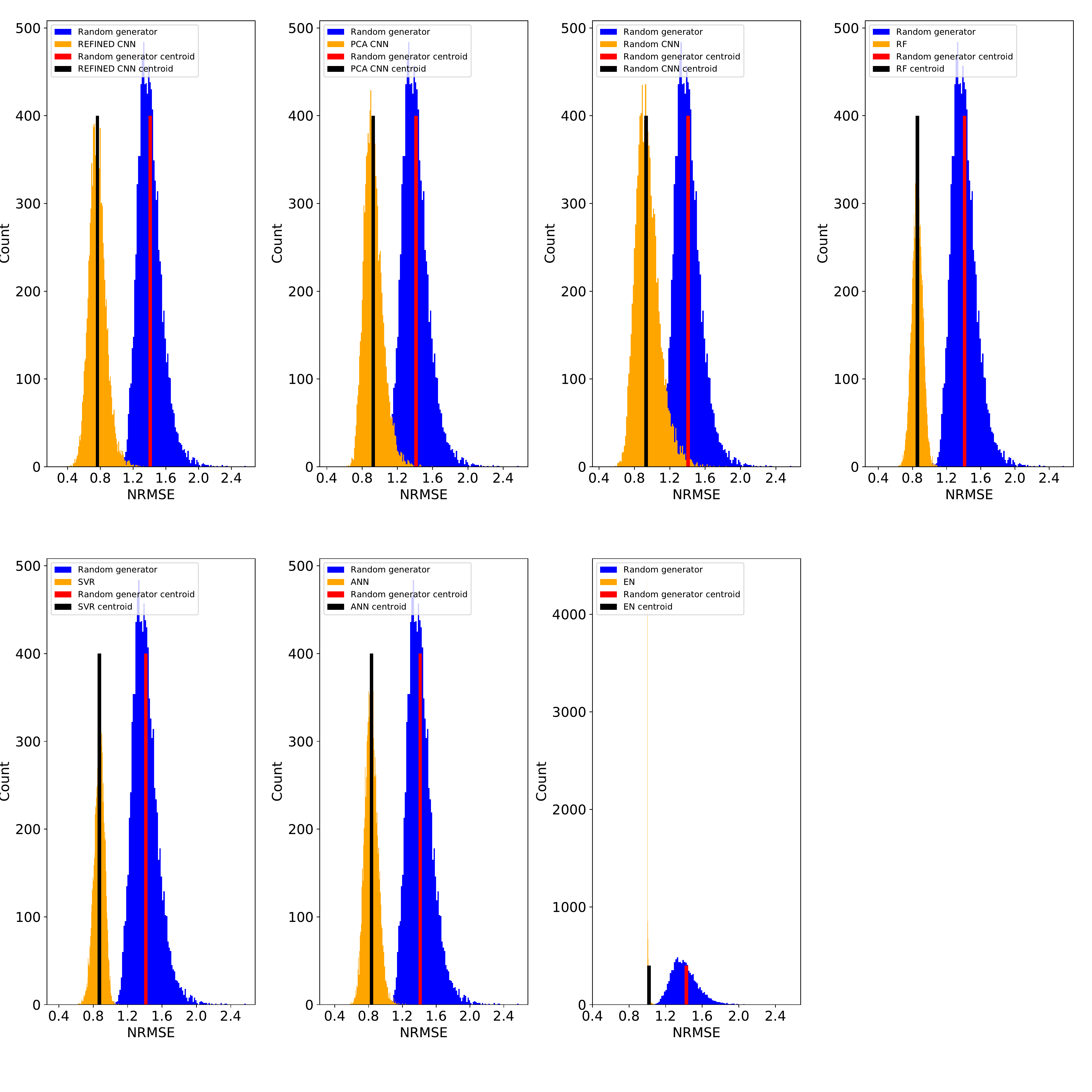}\par
		\subcaption{NRMSEs of DMS\_273 cell line}
		\includegraphics[width=\linewidth,height=0.25\textheight]{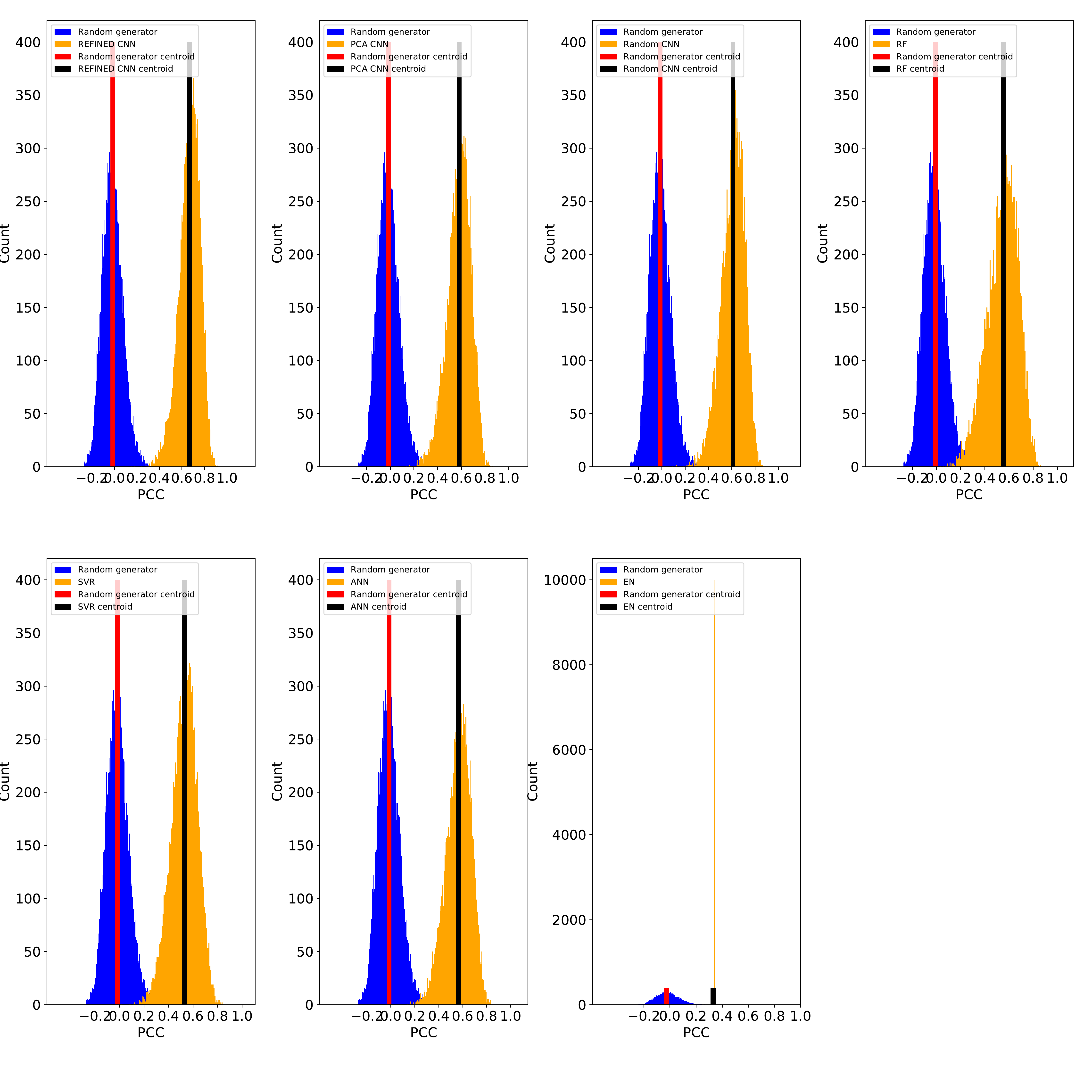}\par
		\subcaption{PCCs of DMS\_273 cell line}
	\end{multicols}
	\caption{Distribution of NRMSEs and PCCs metric of all seven models drawn from the Gap statistics test for three cell lines of the NCI60 dataset. The distributions clustered into two groups and their associated cluster centroids are shown with a vertical bar on the histogram plots.}
	\label{GapDist}
\end{figure*}

\begin{figure}[!htbp]
	\centering
	\includegraphics[width=\textwidth]{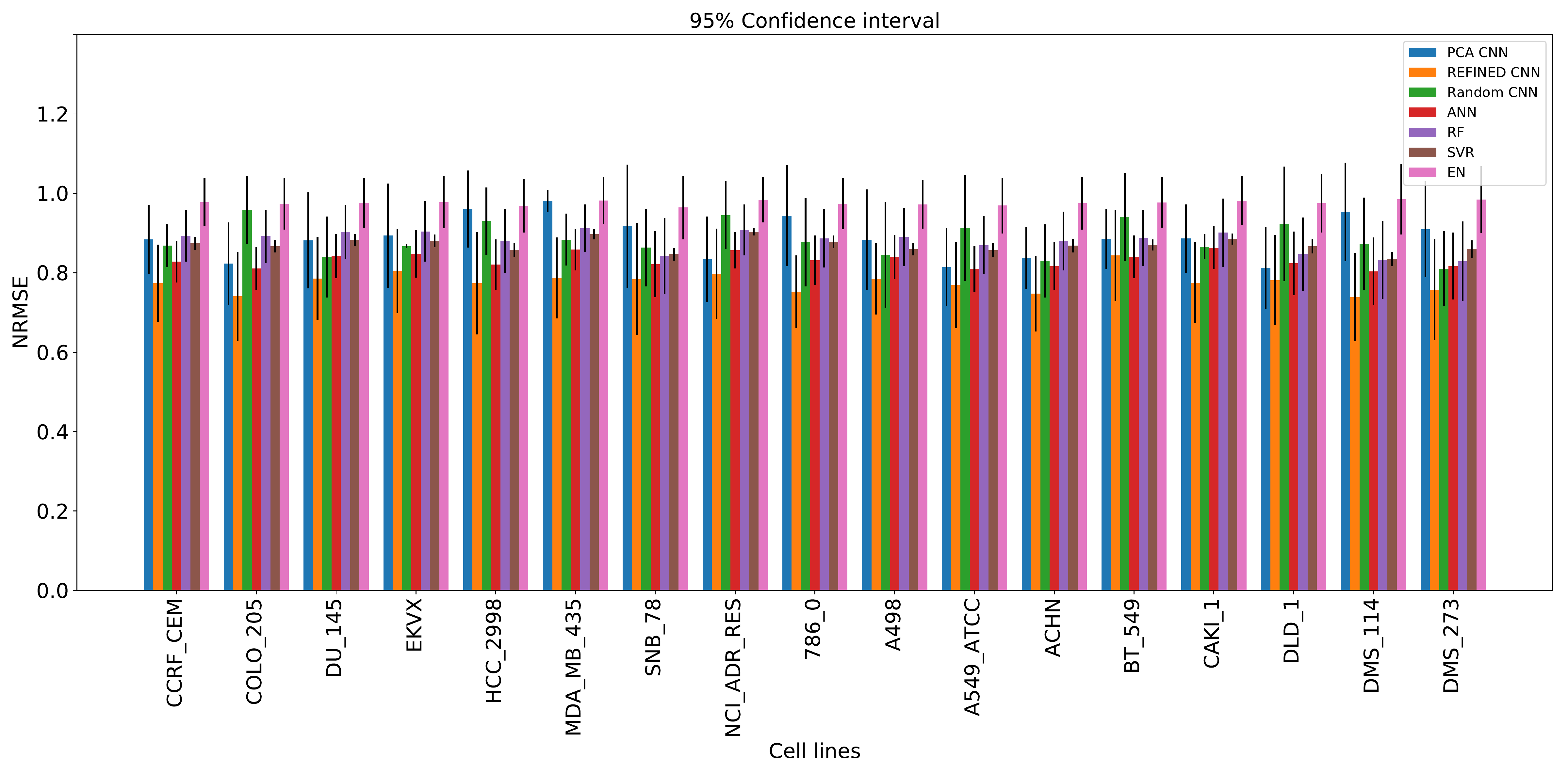} 
	\caption{NRMSE with 95 \% confidence interval of each model trained on 17 cell lines of NCI60 dataset, reported for each cell line per model separately.}
	\label{NCI_NRMSE_Conf}  
\end{figure}

\begin{figure}[!htbp]
	\centering
	\includegraphics[width=\textwidth]{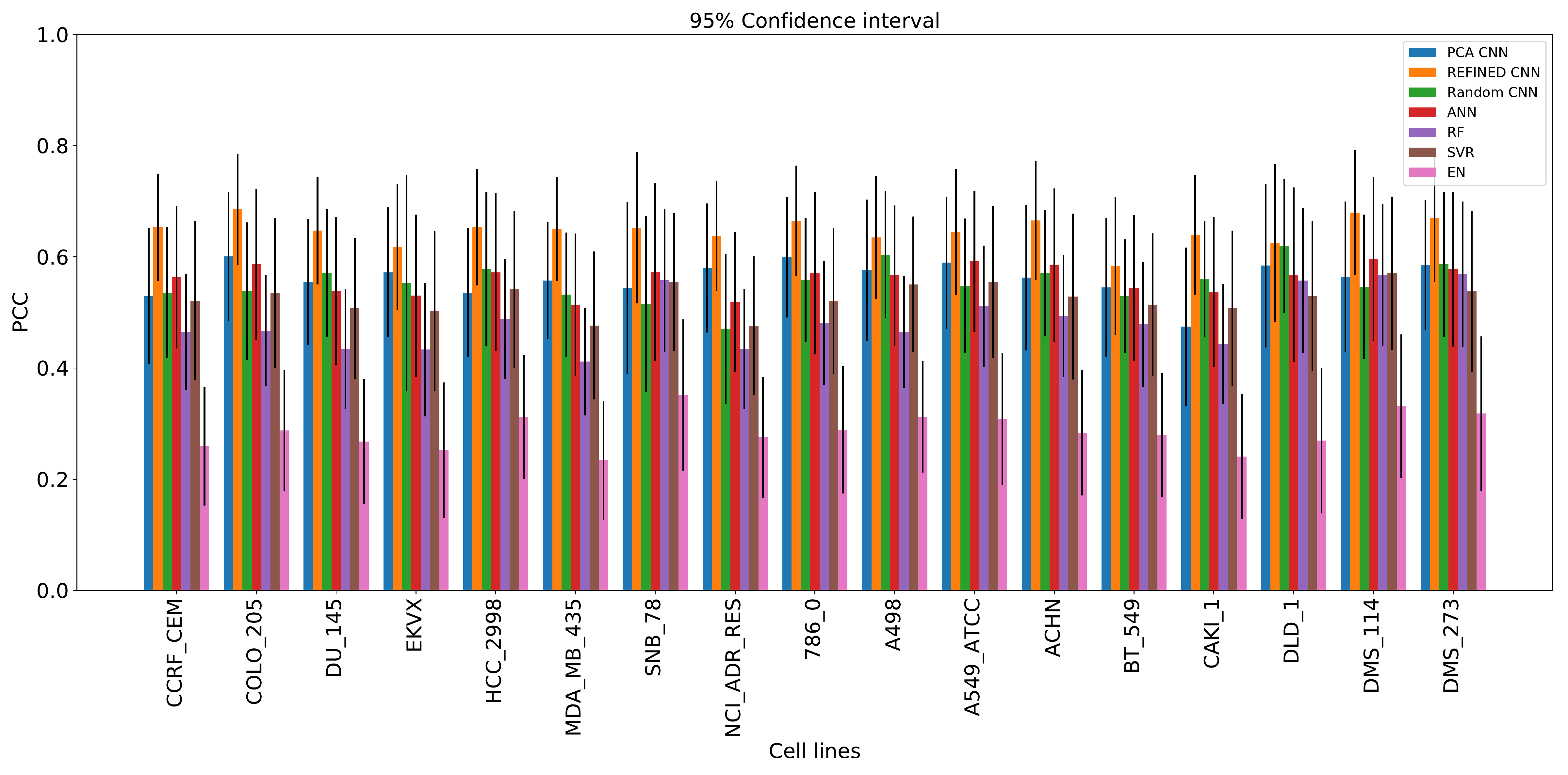} 
	\caption{PCC with 95 \% confidence interval of each model trained on 17 cell lines of NCI60 dataset, reported for each cell line per model separately.}
	\label{NCI_PCC_Conf}  
\end{figure}

\begin{figure}[!htbp]
	\centering
	\includegraphics[width=\textwidth]{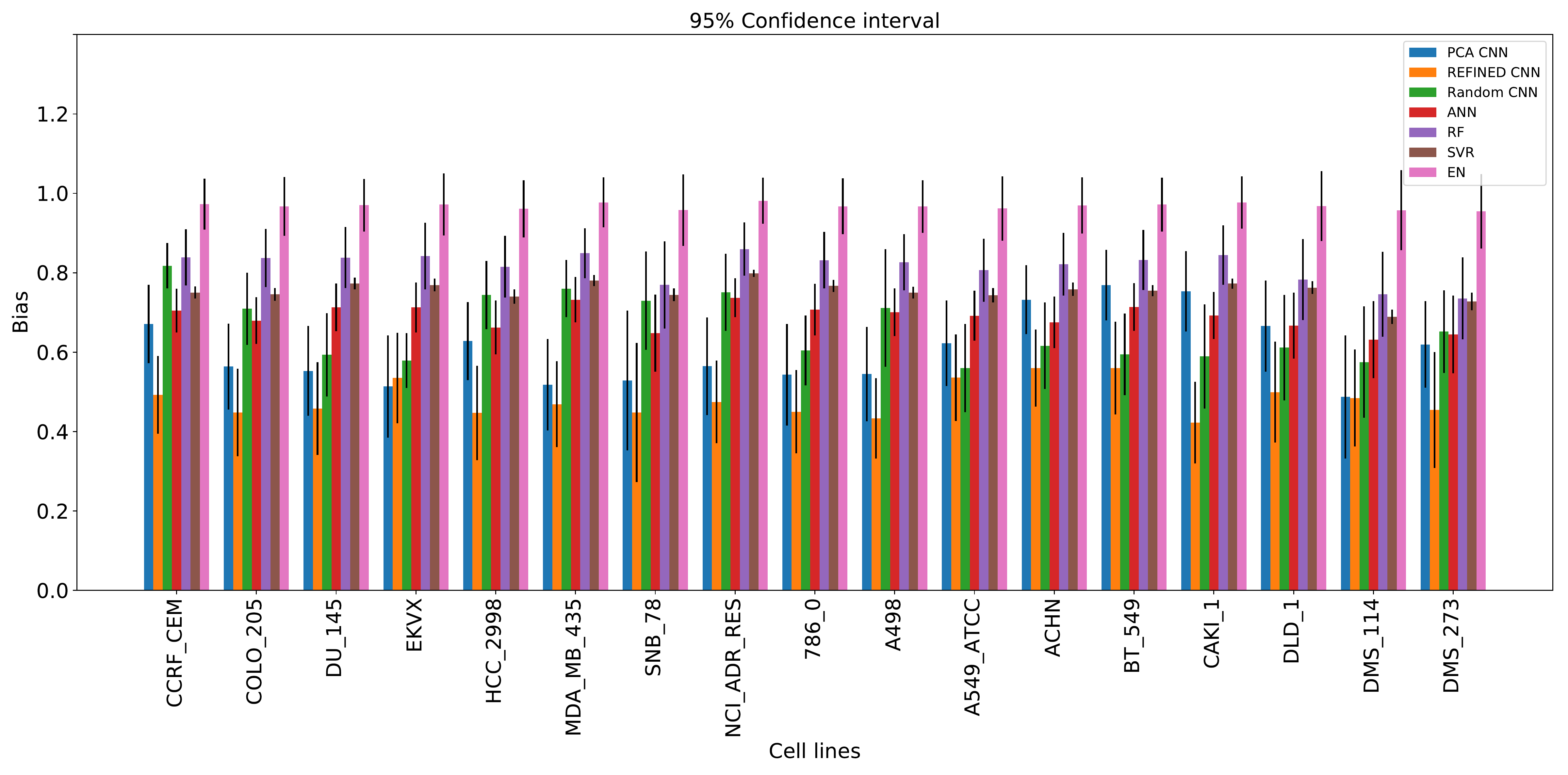} 
	\caption{Bias with 95 \% confidence interval of each model trained on 17 cell lines of NCI60 dataset, reported for each cell line per model separately.}
	\label{NCI_Bias_Conf}  
\end{figure}

\begin{table}[!htbp]
	\centering
	\caption{REFINED on different training size for randomly selected five cell lines }
	\label{SampleSize}
	\resizebox{\textwidth}{!}{%
		\begin{tabular}{c||cc||cc||cc||cc||cc}
			& \multicolumn{2}{c||}{CCRF\_CEM} & \multicolumn{2}{c||}{EKVX} & \multicolumn{2}{c||}{MDA\_MB\_435} & \multicolumn{2}{c||}{NCI\_ADR\_RES} & \multicolumn{2}{c}{SNB\_78} \\
			Training size & NRMSE         & PCC           & NRMSE       & PCC        & NRMSE           & PCC            & NRMSE           & PCC             & NRMSE        & PCC          \\
			\hline
			\hline
			20 \%  & 1.279 & 0.373 & 0.899 & 0.4980 & 0.949 & 0.454 & 0.953  & 0.464 & 0.876 & 0.537 \\
			40 \%  & 0.985 & 0.478 & 0.881 & 0.511  & 0.911 & 0.494 & 0.958  & 0.536 & 0.825 & 0.590 \\
			60 \%  & 0.903 & 0.538 & 0.849 & 0.525  & 0.839 & 0.575 & 0.857  & 0.528 & 0.806 & 0.591 \\
			80 \%  & 0.774 & 0.653 & 0.804 & 0.618  & 0.787 & 0.651 & 0.798  & 0.638 & 0.784 & 0.652       
		\end{tabular}%
	}
\end{table}



\begin{figure}[]
	\centering
	\includegraphics[width=\textwidth]{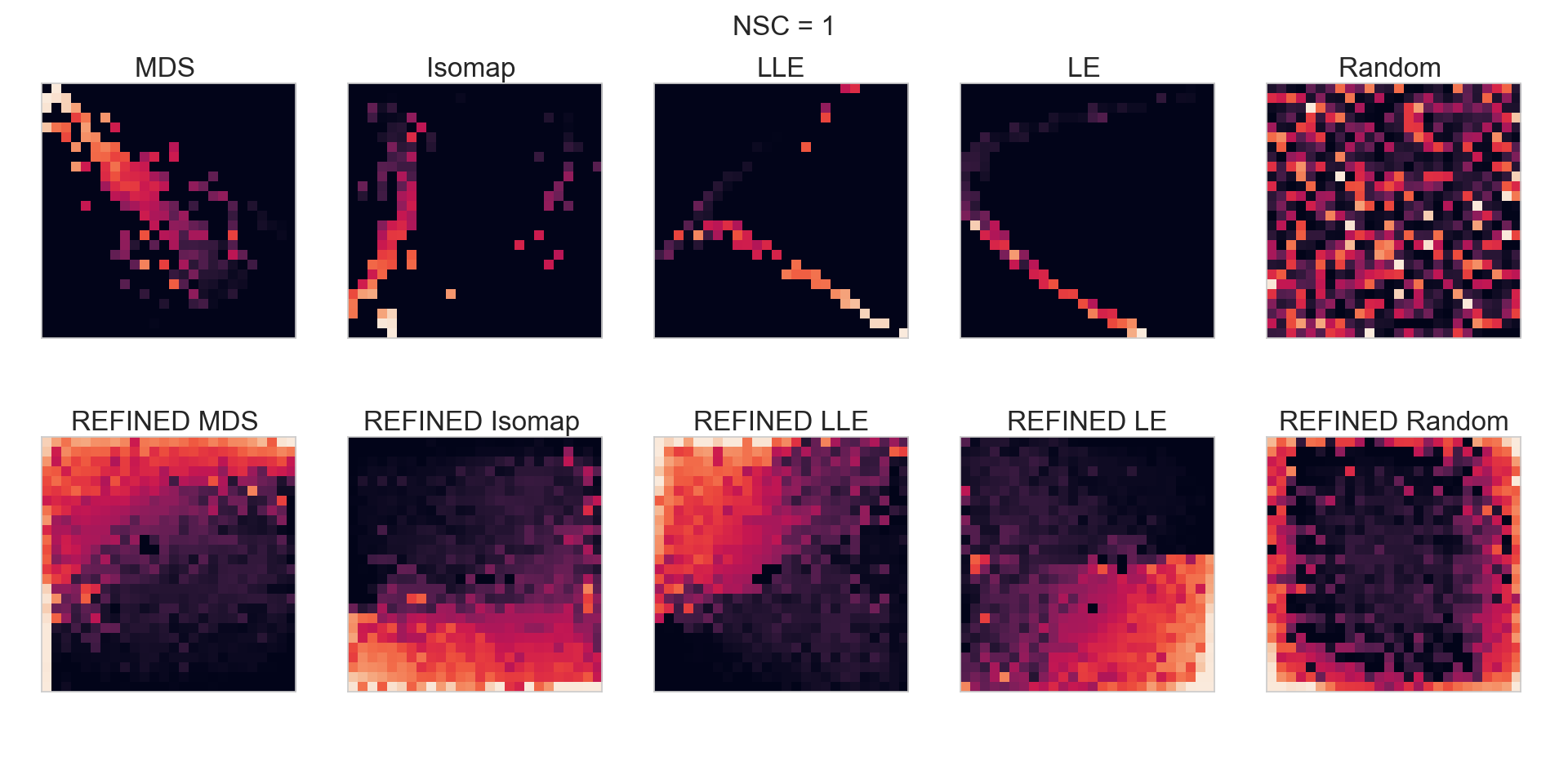} 
	\caption{First row: Images created using different dimensionality reduction techniques. Second row:  Images created using REFINED initialized with different dimensionality reduction techniques. Corresponding NSC = 1 }
	\label{Image_Generation_2}  
\end{figure}

\begin{figure}[]
	\centering
	\includegraphics[width=\textwidth]{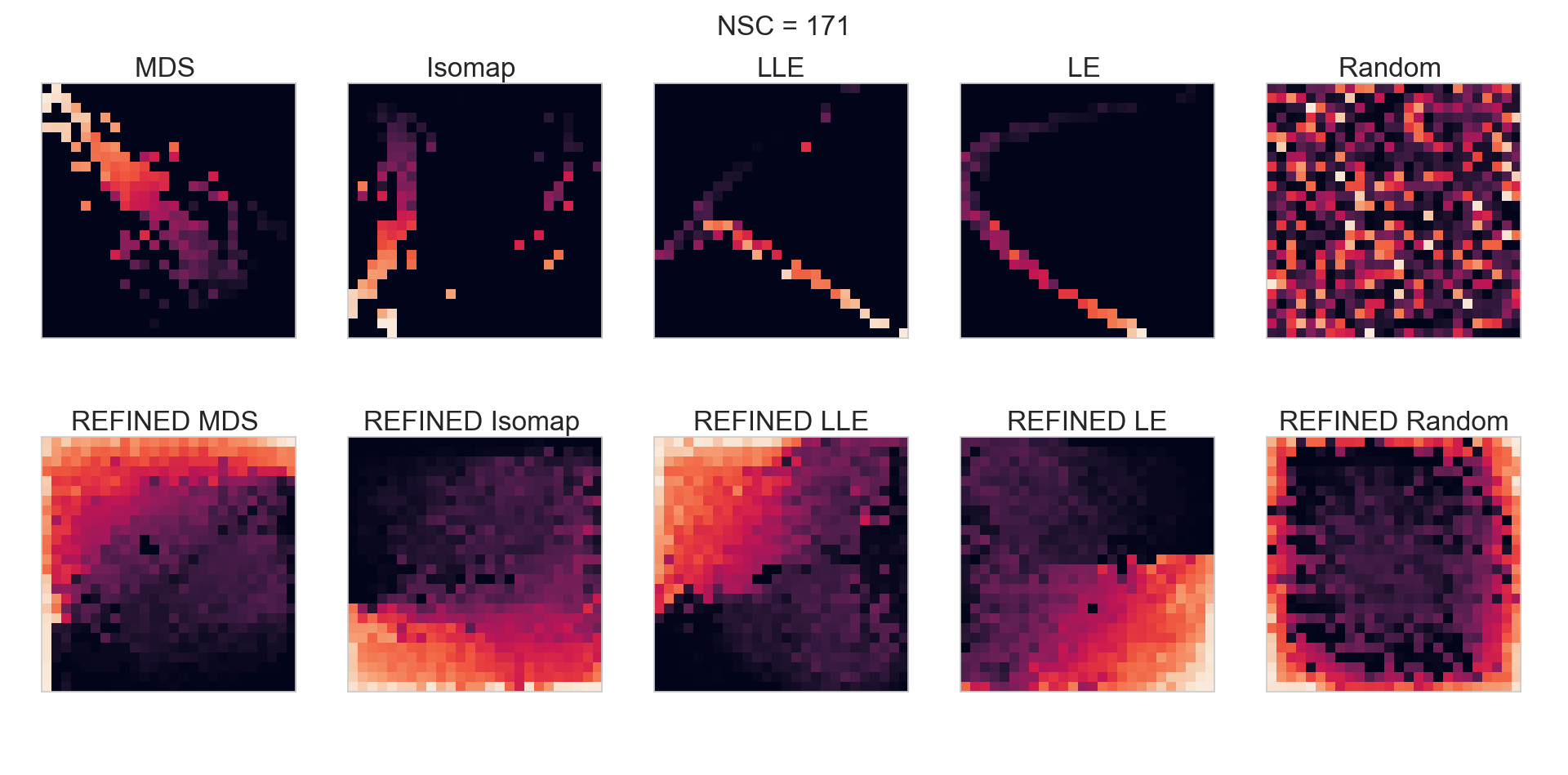} 
	\caption{First row: Images created using different dimensionality reduction techniques. Second row:  Images created using REFINED initialized with different dimensionality reduction techniques. Corresponding NSC = 171}
	\label{Image_Generation_3}  
\end{figure}

\begin{figure}[]
	\centering
	\includegraphics[width=\textwidth]{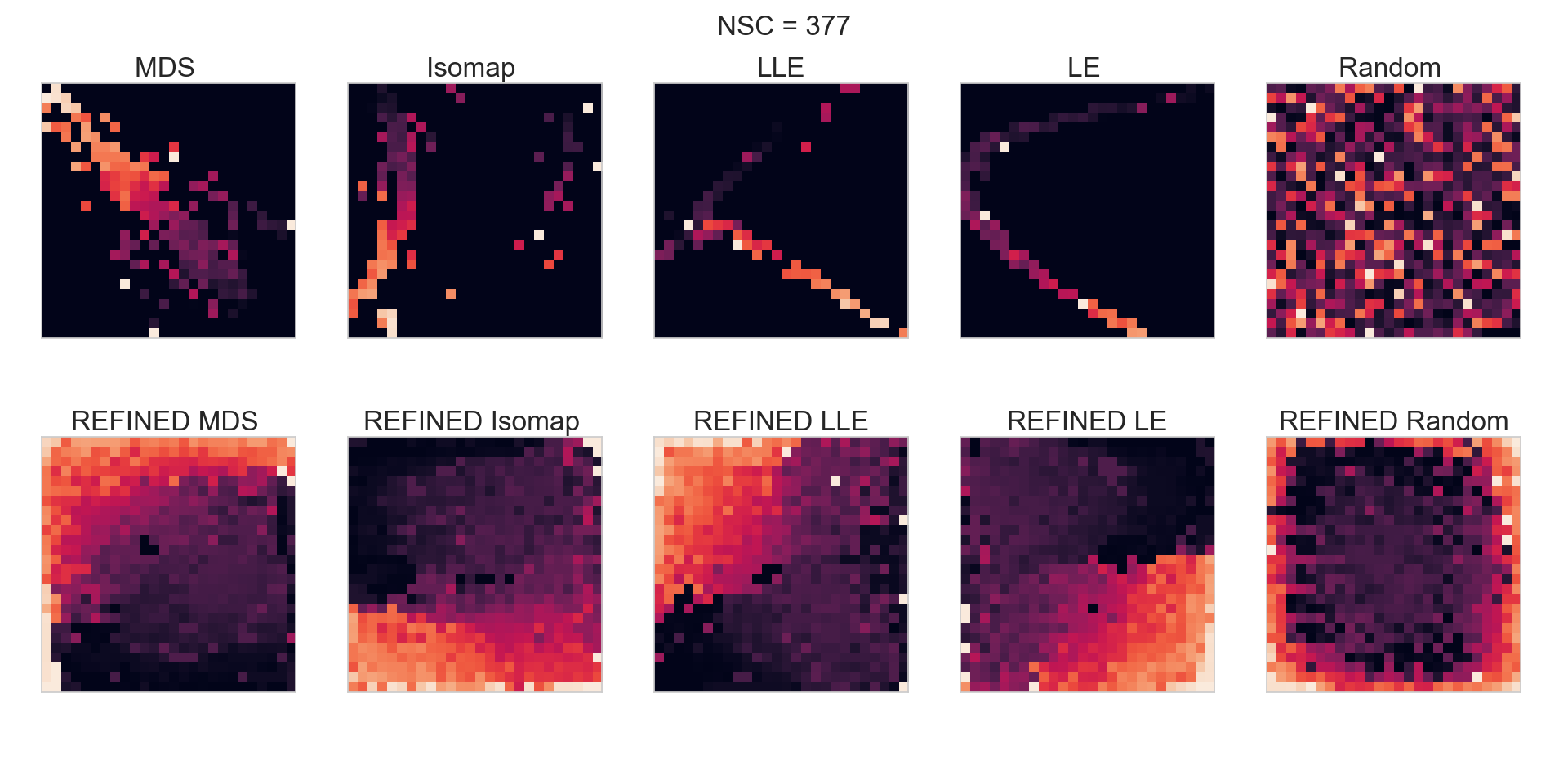} 
	\caption{First row: Images created using different dimensionality reduction techniques. Second row:  Images created using REFINED initialized with different dimensionality reduction techniques. Corresponding NSC = 377}
	\label{Image_Generation_4}  
\end{figure}

\begin{figure}[]
	\centering
	\includegraphics[width=\textwidth]{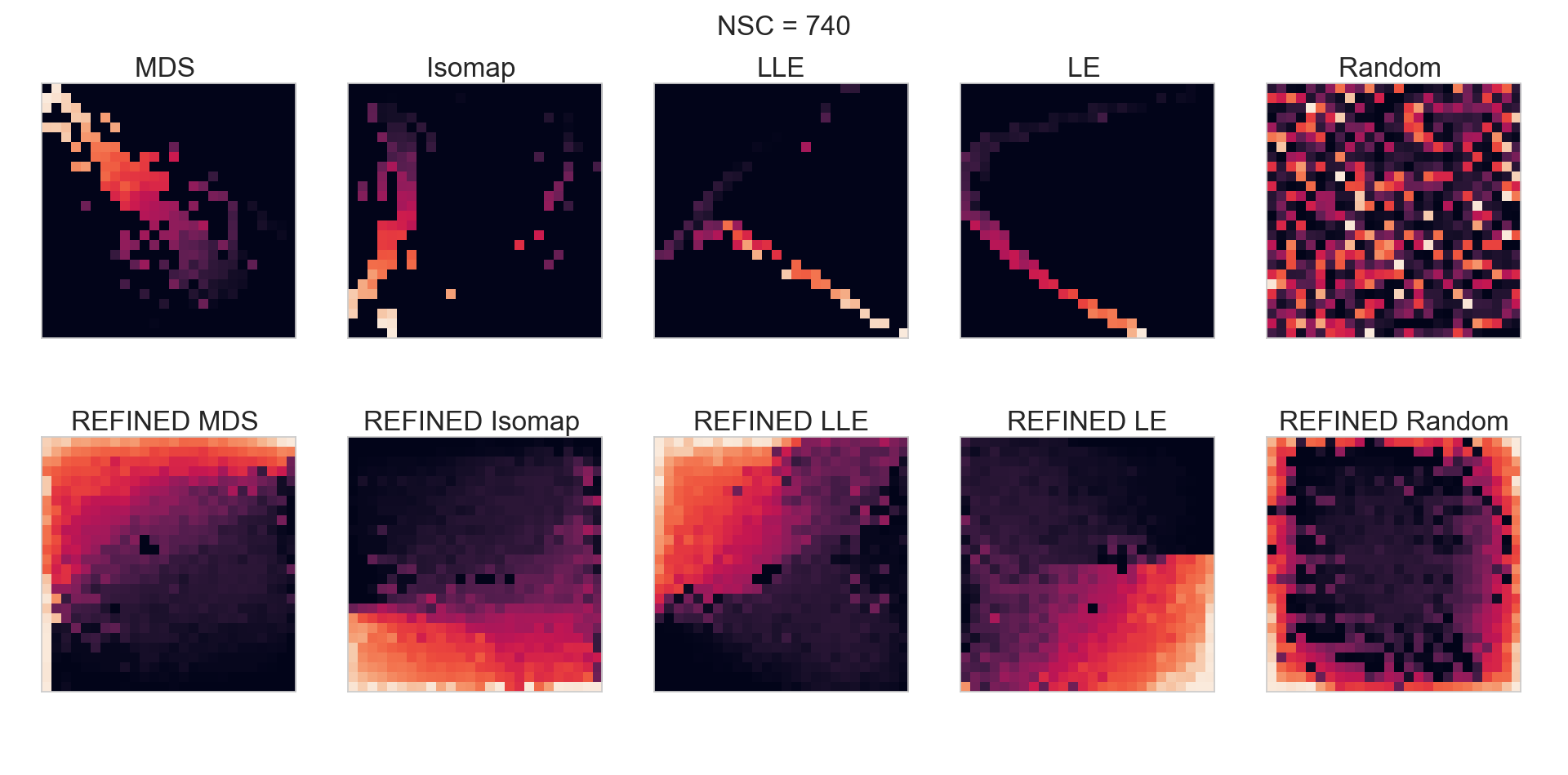} 
	\caption{First row: Images created using different dimensionality reduction techniques. Second row:  Images created using REFINED initialized with different dimensionality reduction techniques. Corresponding NSC = 740}
	\label{Image_Generation_5}  
\end{figure}

\begin{figure}[]
	\centering
	\includegraphics[width=\textwidth]{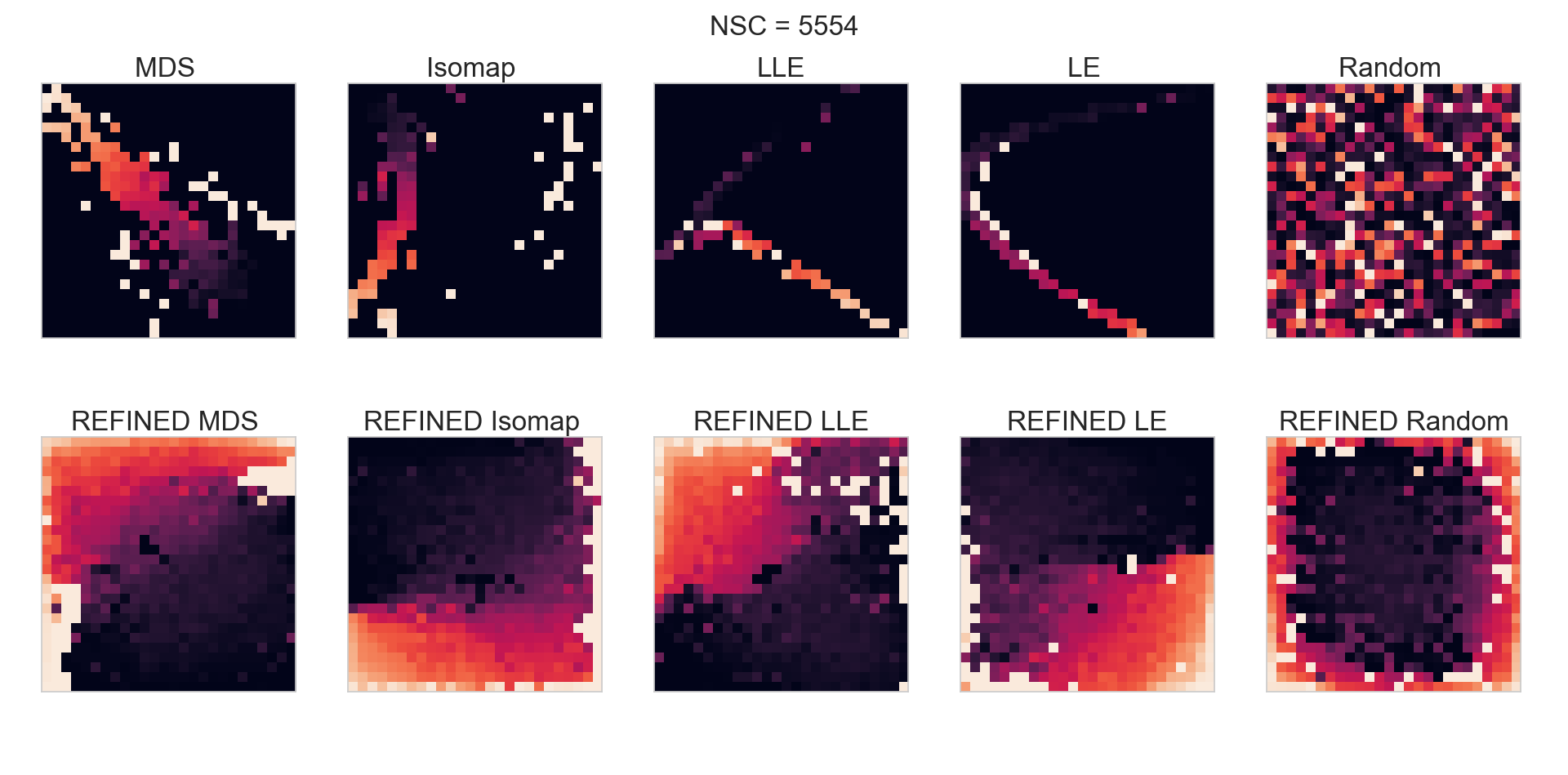} 
	\caption{First row: Images created using different dimensionality reduction techniques. Second row:  Images created using REFINED initialized with different dimensionality reduction techniques. Corresponding NSC = 5554}
	\label{Image_Generation_6}  
\end{figure}

\begin{table}[]
	\centering
	\caption{Results of training CNN on the images created from NCI60 dataset using different dimensionality reduction techniques explained in the Ablation study.}
	\label{Ablation_init}
	\resizebox{\textwidth}{!}{%
	\begin{tabular}{c||ccc||ccc||ccc||ccc||ccc}
		\multirow{2}{*}{Cell line} & \multicolumn{3}{c||}{Isomap} & \multicolumn{3}{c||}{LLE} & \multicolumn{3}{c||}{MDS} & \multicolumn{3}{c||}{LE} & \multicolumn{3}{c}{Random} \\
		& NRMSE      & PCC        & Bias      & NRMSE     & PCC       & Bias     & NRMSE     & PCC       & Bias     & NRMSE     & PCC      & Bias     & NRMSE      & PCC        & Bias      \\
		\hline
		\hline
		CCRF\_CEM     & 0.962 & 0.470 & 0.701 & 0.980 & 0.233 & 0.963 & 0.831 & 0.622 & 0.667 & 0.797 & 0.614 & 0.561 & 0.868 & 0.536 & 0.818 \\
		COLO\_205     & 0.891 & 0.584 & 0.514 & 1.262 & 0.000 & 1.000 & 0.787 & 0.622 & 0.59  & 0.785 & 0.641 & 0.483 & 0.958 & 0.538 & 0.709 \\
		DU\_145       & 0.868 & 0.507 & 0.724 & 0.979 & 0.406 & 0.835 & 0.864 & 0.592 & 0.523 & 0.806 & 0.614 & 0.533 & 0.839 & 0.572 & 0.594 \\
		EKVX          & 0.853 & 0.572 & 0.665 & 1.021 & 0.397 & 0.812 & 0.867 & 0.545 & 0.587 & 0.841 & 0.570 & 0.574 & 0.867 & 0.553 & 0.579 \\
		HCC\_2998     & 0.830 & 0.565 & 0.630 & 0.832 & 0.591 & 0.562 & 0.825 & 0.628 & 0.467 & 0.929 & 0.550 & 0.545 & 0.93  & 0.578 & 0.744 \\
		MDA\_MB\_435  & 0.911 & 0.453 & 0.773 & 0.922 & 0.499 & 0.643 & 1.019 & 0.429 & 0.748 & 0.821 & 0.586 & 0.632 & 0.884 & 0.532 & 0.76  \\
		SNB\_78       & 0.828 & 0.576 & 0.598 & 0.859 & 0.519 & 0.754 & 0.819 & 0.588 & 0.684 & 0.817 & 0.585 & 0.684 & 0.864 & 0.516 & 0.73  \\
		NCI\_ADR\_RES & 1.315 & 0.470 & 0.481 & 0.895 & 0.474 & 0.793 & 0.938 & 0.488 & 0.746 & 0.824 & 0.576 & 0.616 & 0.945 & 0.47  & 0.751 \\
		786\_0        & 0.820 & 0.602 & 0.546 & 0.827 & 0.574 & 0.677 & 0.814 & 0.647 & 0.66  & 0.852 & 0.618 & 0.477 & 0.877 & 0.558 & 0.604 \\
		A498          & 0.892 & 0.573 & 0.534 & 1.018 & 0.045 & 0.998 & 0.844 & 0.582 & 0.681 & 0.797 & 0.618 & 0.537 & 0.845 & 0.604 & 0.712 \\
		A549\_ATCC    & 0.801 & 0.614 & 0.557 & 0.851 & 0.559 & 0.724 & 0.896 & 0.599 & 0.765 & 0.825 & 0.631 & 0.470 & 0.913 & 0.548 & 0.56  \\
		ACHN          & 1.109 & 0.494 & 0.602 & 0.914 & 0.462 & 0.815 & 0.789 & 0.631 & 0.514 & 0.900 & 0.601 & 0.449 & 0.83  & 0.571 & 0.616 \\
		BT\_549       & 0.850 & 0.530 & 0.701 & 0.890 & 0.465 & 0.768 & 0.926 & 0.499 & 0.764 & 0.835 & 0.562 & 0.624 & 0.941 & 0.529 & 0.595 \\
		CAKI\_1       & 0.874 & 0.548 & 0.649 & 1.439 & 0.376 & 0.781 & 0.935 & 0.546 & 0.546 & 0.799 & 0.609 & 0.569 & 0.866 & 0.561 & 0.589 \\
		DLD\_1        & 0.880 & 0.587 & 0.531 & 0.902 & 0.468 & 0.829 & 0.786 & 0.629 & 0.533 & 0.792 & 0.636 & 0.652 & 0.923 & 0.62  & 0.611 \\
		DMS\_114      & 0.830 & 0.577 & 0.599 & 0.854 & 0.539 & 0.709 & 0.847 & 0.619 & 0.467 & 0.898 & 0.602 & 0.503 & 0.873 & 0.546 & 0.575 \\
		DMS\_273      & 0.780 & 0.639 & 0.529 & 0.813 & 0.598 & 0.588 & 0.801 & 0.62  & 0.52  & 0.771 & 0.639 & 0.579 & 0.81  & 0.587 & 0.652 \\
		\hline
		\hline
		Average       & 0.900 & 0.551 & 0.608 & 0.956 & 0.424 & 0.780 & 0.858 & 0.582 & 0.616 & 0.829 & 0.603 & 0.558 & 0.884 & 0.554 & 0.659
	\end{tabular}
}
\end{table}

\begin{table}[]\caption{Results of training CNN on the images created from NCI60 dataset using REFINED initialized with different dimensionality reduction techniques explained in the Ablation study.}\label{Ablation_REFINED}
	\centering
	\resizebox{\textwidth}{!}{%
		\begin{tabular}{c||ccc||ccc||ccc||ccc||ccc}
			\multirow{2}{*}{Cell line} & \multicolumn{3}{c||}{REFINED\_Isomap} & \multicolumn{3}{c||}{REFINED\_LLE} & \multicolumn{3}{c||}{REFINED\_MDS} & \multicolumn{3}{c||}{REFINED\_LE} & \multicolumn{3}{c}{REFINED\_Random} \\
			& NRMSE      & PCC        & Bias      & NRMSE     & PCC       & Bias     & NRMSE     & PCC       & Bias     & NRMSE     & PCC      & Bias     & NRMSE      & PCC        & Bias      \\
			\hline
			\hline
			CCRF\_CEM                  & 0.799      & 0.630      & 0.498     & 0.781     & 0.638     & 0.515    & 0.774     & 0.653     & 0.493    & 0.790     & 0.633    & 0.510    & 0.816      & 0.586      & 0.725     \\
			COLO\_205                  & 0.752      & 0.673      & 0.493     & 0.765     & 0.652     & 0.509    & 0.741     & 0.686     & 0.448    & 0.782     & 0.658    & 0.433    & 0.812      & 0.606      & 0.662     \\
			DU\_145                    & 0.858      & 0.628      & 0.443     & 0.825     & 0.613     & 0.531    & 0.786     & 0.647     & 0.458    & 0.822     & 0.621    & 0.595    & 0.935      & 0.518      & 0.604     \\
			EKVX                       & 0.790      & 0.634      & 0.504     & 0.839     & 0.587     & 0.526    & 0.804     & 0.618     & 0.535    & 0.824     & 0.601    & 0.622    & 0.998      & 0.528      & 0.580     \\
			HCC\_2998                  & 0.758      & 0.668      & 0.456     & 0.834     & 0.623     & 0.465    & 0.774     & 0.654     & 0.447    & 0.772     & 0.660    & 0.451    & 0.833      & 0.608      & 0.454     \\
			MDA\_MB\_435               & 0.817      & 0.622      & 0.495     & 0.893     & 0.574     & 0.688    & 0.787     & 0.651     & 0.469    & 0.824     & 0.609    & 0.559    & 0.811      & 0.627      & 0.586     \\
			SNB\_78                    & 0.827      & 0.636      & 0.436     & 0.831     & 0.626     & 0.458    & 0.784     & 0.652     & 0.448    & 0.755     & 0.661    & 0.547    & 0.846      & 0.542      & 0.778     \\
			NCI\_ADR\_RES              & 0.791      & 0.632      & 0.517     & 0.823     & 0.603     & 0.583    & 0.798     & 0.638     & 0.475    & 0.811     & 0.608    & 0.547    & 0.810      & 0.542      & 0.593     \\
			786\_0                     & 0.751      & 0.678      & 0.504     & 0.763     & 0.663     & 0.467    & 0.752     & 0.665     & 0.450    & 0.756     & 0.672    & 0.451    & 0.784      & 0.637      & 0.661     \\
			A498                       & 0.799      & 0.624      & 0.508     & 0.785     & 0.662     & 0.435    & 0.785     & 0.635     & 0.433    & 0.789     & 0.633    & 0.569    & 0.787      & 0.631      & 0.669     \\
			A549\_ATCC                 & 0.754      & 0.673      & 0.472     & 0.779     & 0.650     & 0.465    & 0.769     & 0.645     & 0.536    & 0.750     & 0.675    & 0.469    & 0.858      & 0.556      & 0.473     \\
			ACHN                       & 0.768      & 0.684      & 0.526     & 0.766     & 0.656     & 0.503    & 0.747     & 0.665     & 0.560    & 0.756     & 0.659    & 0.530    & 0.922      & 0.585      & 0.551     \\
			BT\_549                    & 0.812      & 0.594      & 0.583     & 0.821     & 0.606     & 0.521    & 0.843     & 0.584     & 0.560    & 0.801     & 0.609    & 0.585    & 0.822      & 0.591      & 0.708     \\
			CAKI\_1                    & 0.785      & 0.639      & 0.491     & 0.817     & 0.590     & 0.608    & 0.775     & 0.640     & 0.423    & 0.816     & 0.617    & 0.510    & 0.836      & 0.571      & 0.666     \\
			DLD\_1                     & 0.729      & 0.687      & 0.506     & 0.821     & 0.622     & 0.527    & 0.781     & 0.625     & 0.499    & 0.750     & 0.680    & 0.433    & 0.832      & 0.594      & 0.537     \\
			DMS\_114                   & 0.764      & 0.671      & 0.510     & 0.736     & 0.690     & 0.431    & 0.738     & 0.680     & 0.484    & 0.749     & 0.676    & 0.467    & 0.857      & 0.563      & 0.576     \\
			DMS\_273                   & 0.754      & 0.667      & 0.492     & 0.752     & 0.693     & 0.478    & 0.758     & 0.670     & 0.454    & 0.722     & 0.701    & 0.453    & 0.866      & 0.546      & 0.463     \\
			\hline
			\hline
			Average                    & 0.783      & 0.649      & 0.496     & 0.802     & 0.632     & 0.512    & 0.776     & 0.647     & 0.481    & 0.781     & 0.645    & 0.514    & 0.849      & 0.578      & 0.605    
		\end{tabular}%
	}
\end{table}

\newpage
\subsection{GDSC dataset}

\begin{figure}[!htbp]
	\centering
	\includegraphics[width=\textwidth]{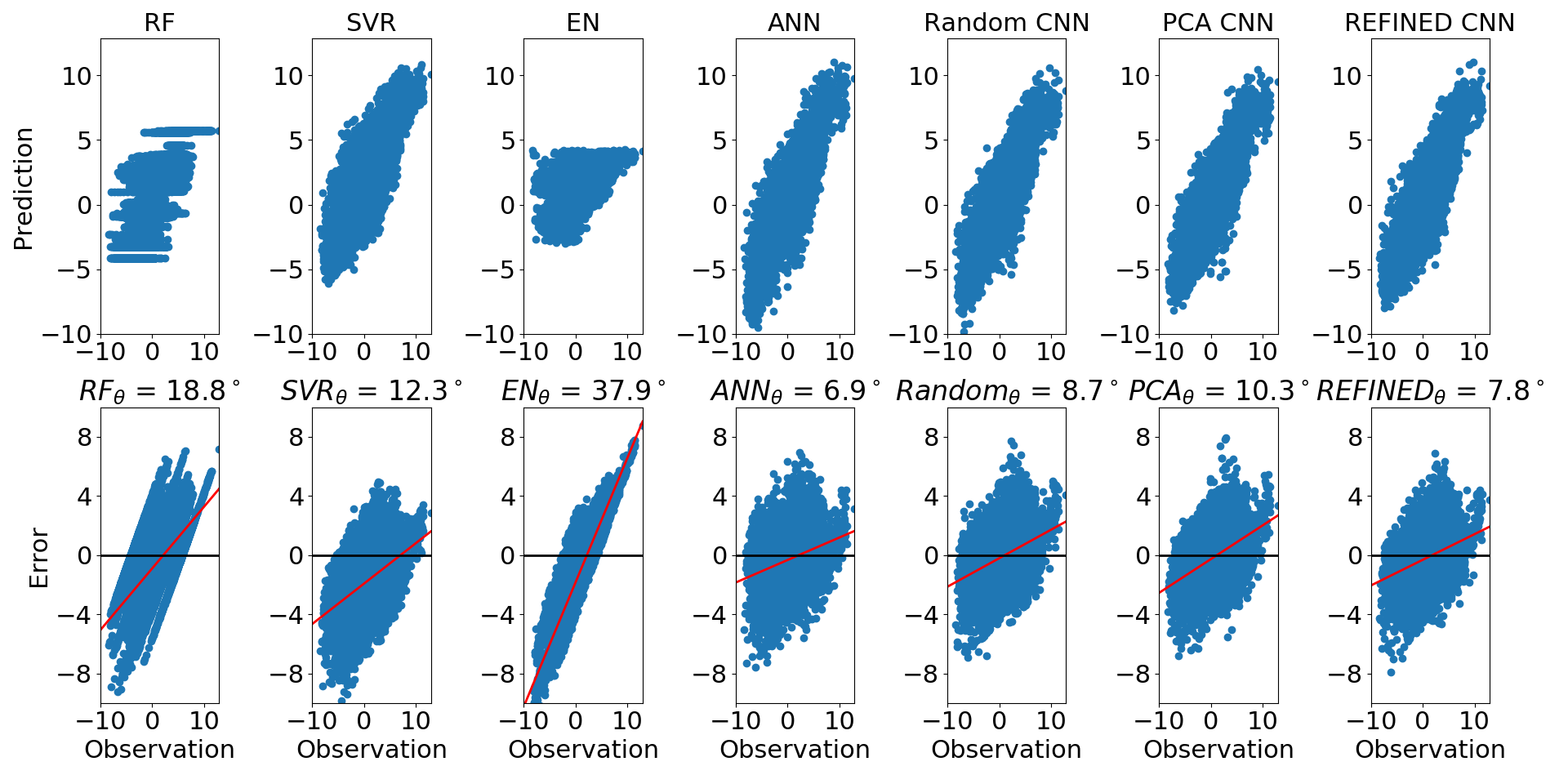} 
	\caption{Scatter plot of predicted NLOGIC50s and their residual (error) for each model, in the case where models were trained on 50 \% of the available data}
	\label{GSC 0.5}  
\end{figure}

\begin{figure}[!htbp]
	\centering
	\includegraphics[width=\textwidth]{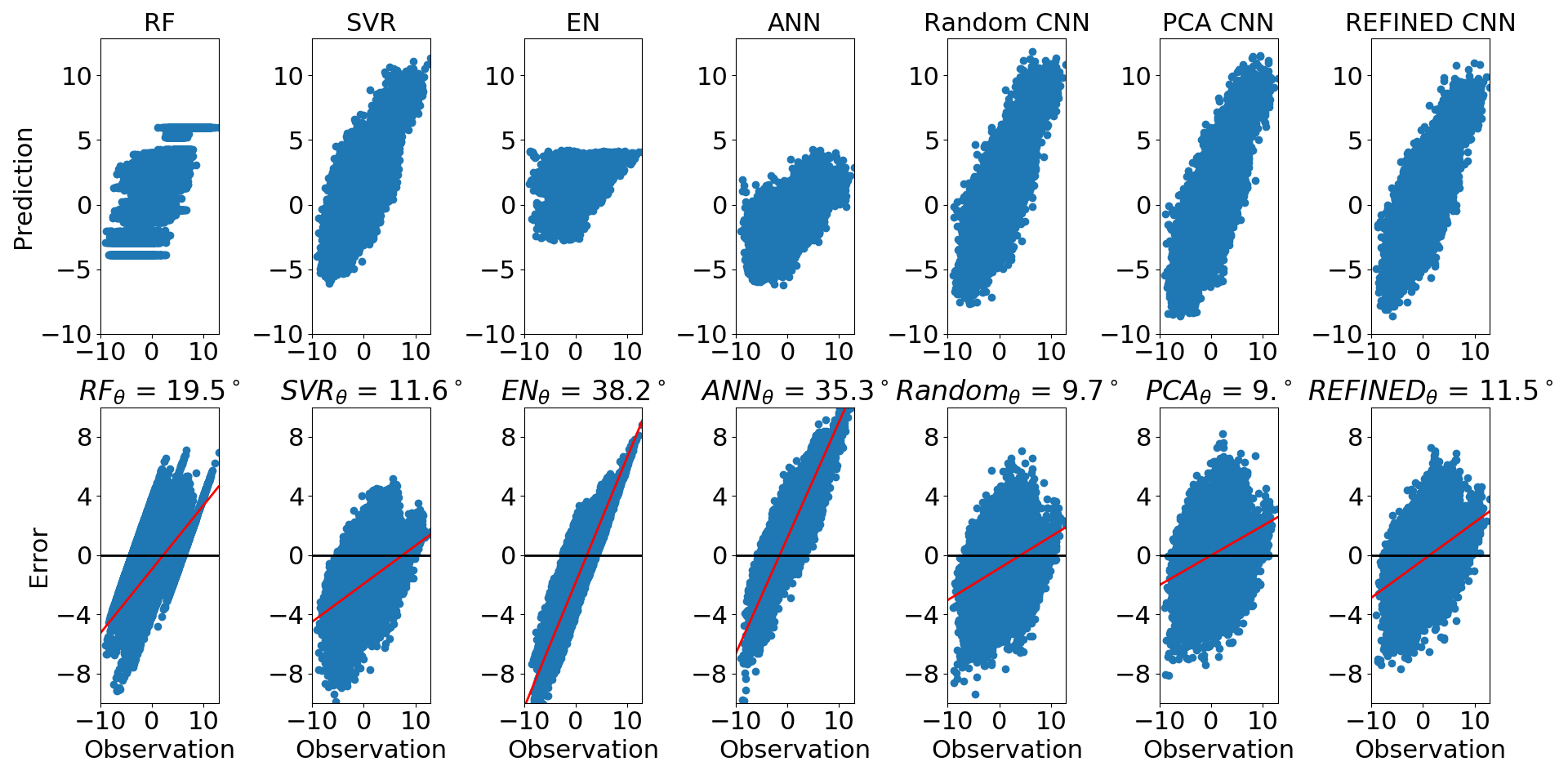} 
	\caption{Scatter plot of predicted NLOGIC50s and their residual (error) for each model, in the case where models were trained on 20 \% of the available data}
	\label{GSC 0.2}  
\end{figure}

\begin{table}[]
	\centering
	\caption{Robustness analysis to compare REFINED CNN with 8 other competing models per each metric. Each cell of the table represents that in what percentage REFINED CNN outperforms the paired competing model.}
	\label{GDSC_Robust}
	\resizebox{0.5\textwidth}{!}{%
		\begin{tabular}{c||ccc}
			Comparison                & NRMSE & PCC   & Bias  \\
			\hline
			\hline
			REFINED CNN vs ANN        & 62.23 & 60.68 & 73.49 \\
			REFINED CNN vs RF         & 98.39 & 98.19 & 97.95 \\
			REFINED CNN vs SVR        & 93.83 & 93.36 & 73.47 \\
			REFINED CNN vs EN         & 100   & 100   & 100   \\
			REFINED CNN vs PCA CNN    & 66.58 & 62.04 & 68.85 \\
			REFINED CNN vs Random CNN & 65.78 & 69.77 & 65.27 \\
			REFINED CNN vs DRF        & 100   & 100   & 100   \\
			REFINED CNN vs HGNN       & 99.87 & 98.81 & 99.99
		\end{tabular}%
	}
\end{table}

\begin{table}[]
	\centering
	\caption{Gap statistics analysis to compare REFINED CNN with 8 other competing models per each metric paired with the null model. The wider (larger) Gap value indicates better performance.}
	\label{GapGDSC}
	\resizebox{0.5\textwidth}{!}{%
		\begin{tabular}{c||ccc}
			Model       & NRMSE & PCC   & Bias  \\
			\hline
			\hline
			REFINED CNN & 1.010 & 0.902 & 0.853 \\
			PCA CNN     & 0.980 & 0.892 & 0.871 \\
			Random CNN  & 0.983 & 0.895 & 0.828 \\
			RF          & 0.852 & 0.812 & 0.710 \\
			SVR         & 0.898 & 0.844 & 0.808 \\
			ANN         & 0.988 & 0.893 & 0.815 \\
			EN          & 0.535 & 0.476 & 0.200 \\
			DRF         & 0.432 & 0.184 & 0.120 \\
			HGNN        & 0.782 & 0.792 & 0.550
		\end{tabular}%
	}
\end{table}

\begin{figure}[]
	\centering
	\includegraphics[width=\textwidth]{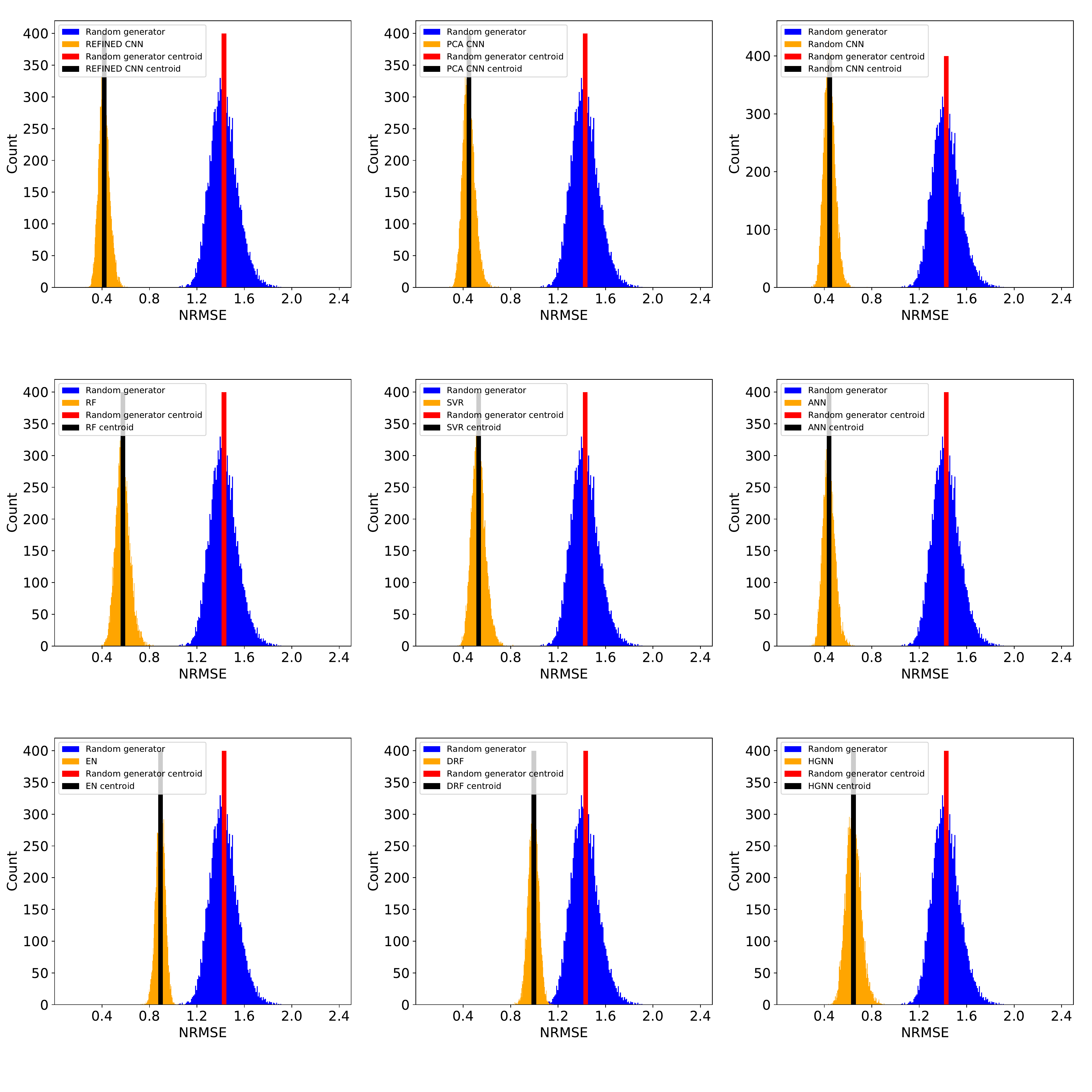} 
	\caption{Distribution of NRMSEs of all nine models drawn from the Gap statistics test of the GDSC dataset. The distributions clustered into two groups and their associated cluster centroids are shown with a vertical bar on the histogram plots.}
	\label{GapGDSCNRMSE}  
\end{figure}

\begin{figure}[]
	\centering
	\includegraphics[width=\textwidth]{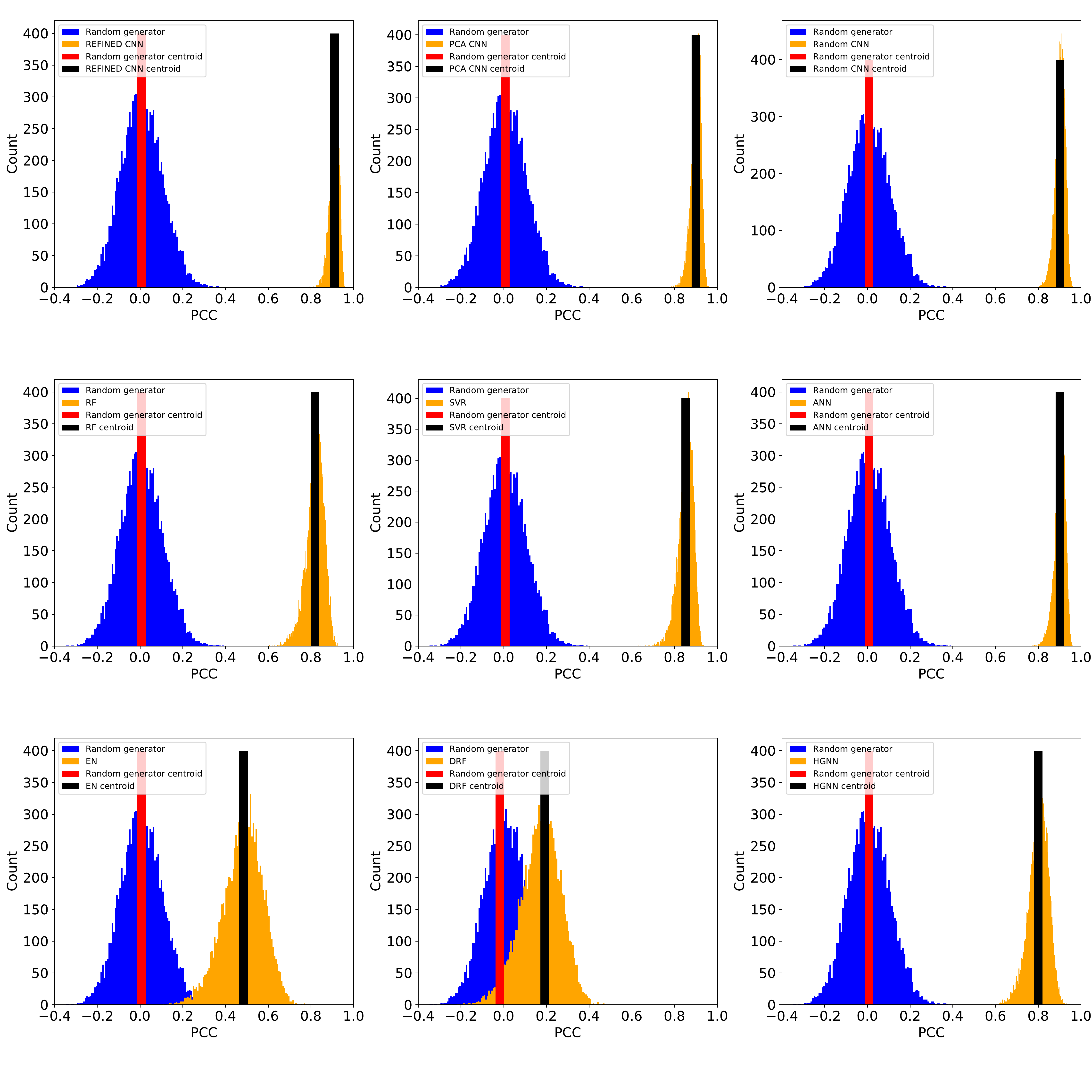} 
	\caption{Distribution of PCCs of all nine models drawn from the Gap statistics test of the GDSC dataset. The distributions clustered into two groups and their associated cluster centroids are shown with a vertical bar on the histogram plots.}
	\label{GapGDSCPCC}  
\end{figure}

\begin{figure}[]
	\centering
	\includegraphics[width=\textwidth]{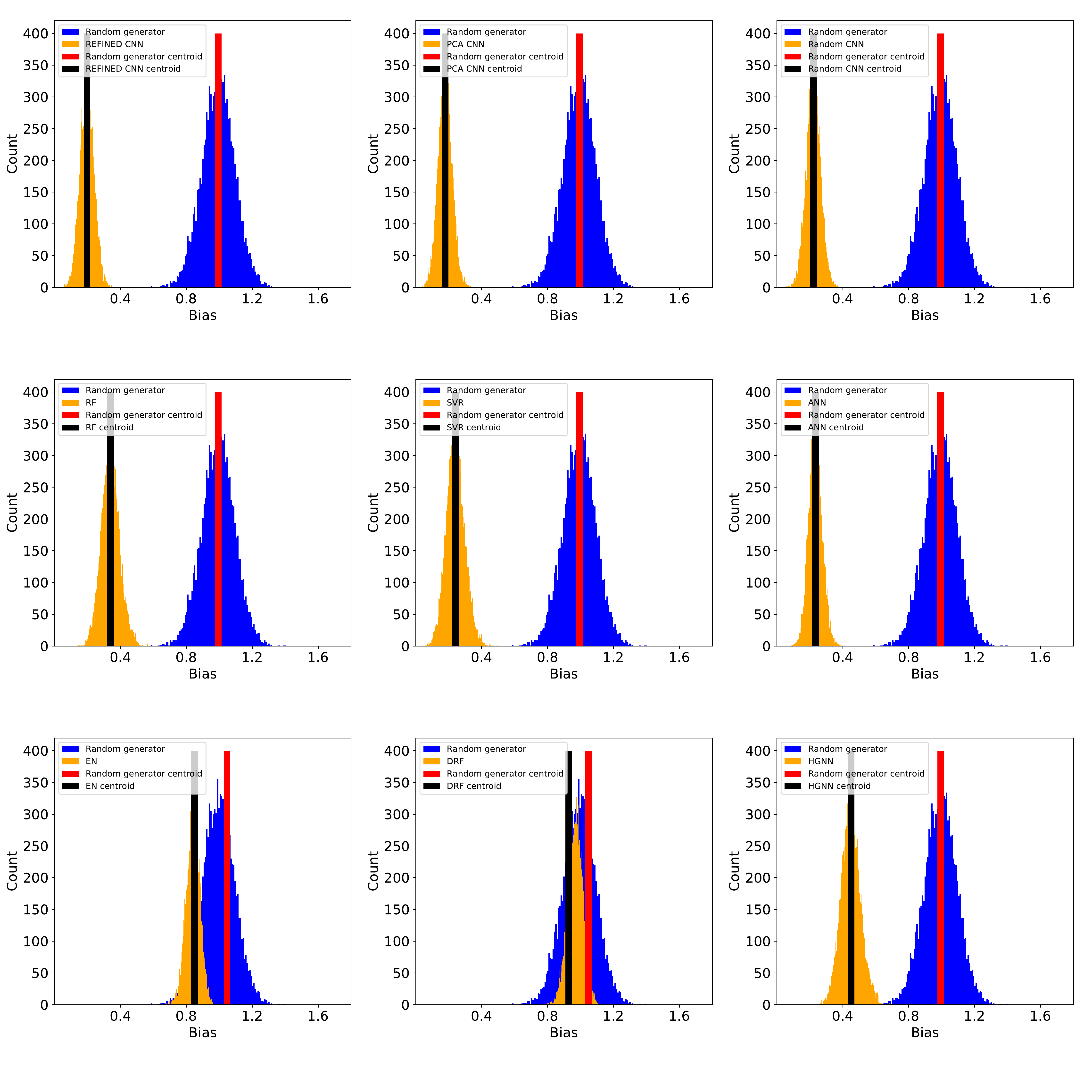} 
	\caption{Distribution of Biases of all nine models drawn from the Gap statistics test of the GDSC dataset. The distributions clustered into two groups and their associated cluster centroids are shown with a vertical bar on the histogram plots.}
	\label{GapGDSCBias}  
\end{figure}

\begin{figure}[]
	\centering
	\includegraphics[width=\textwidth]{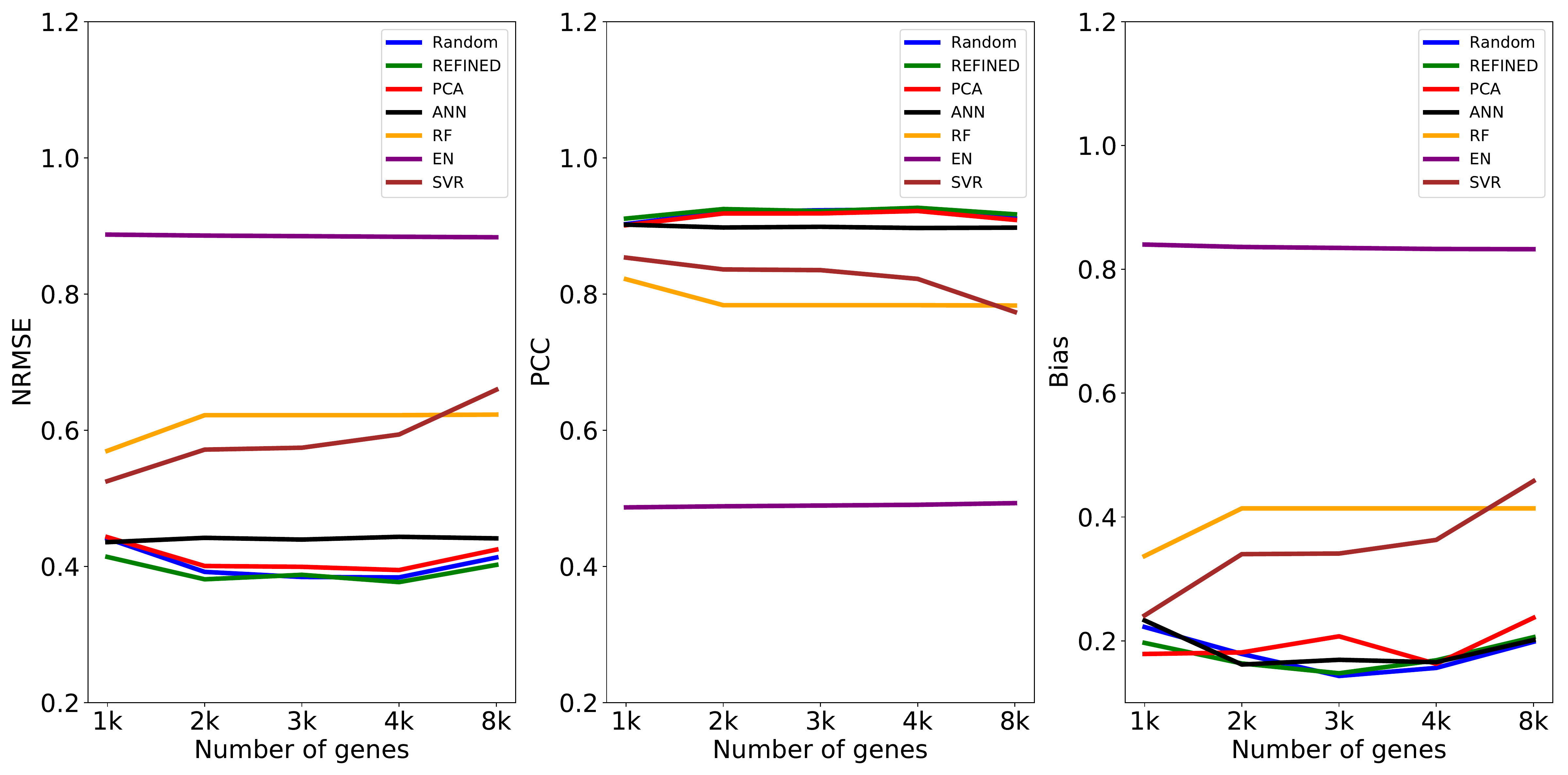} 
	\caption{Modeling GDSC dataset using different number of genes.}
	\label{GDSC_gene_figure}  
\end{figure}

\begin{table}[]
	\centering
	\caption{Detailed results of modeling GDSC dataset using different number of genes.}
	\label{GDSC_gene_table}
	\resizebox{\textwidth}{!}{%
		\begin{tabular}{c||ccc||ccc||ccc||ccc||ccc}
			\multirow{2}{*}{Metrics} & \multicolumn{3}{c||}{1k Genes} & \multicolumn{3}{c||}{2k Genes} & \multicolumn{3}{c||}{3k Genes} & \multicolumn{3}{c||}{4k Genes} & \multicolumn{3}{c}{8k Genes} \\
			& NRMSE    & PCC      & Bias     & NRMSE    & PCC      & Bias     & NRMSE    & PCC      & Bias     & NRMSE    & PCC      & Bias     & NRMSE    & PCC      & Bias     \\
			\hline
			\hline
			EN                       & 0.887    & 0.487    & 0.840    & 0.886    & 0.488    & 0.836    & 0.885    & 0.490    & 0.835    & 0.884    & 0.491    & 0.833    & 0.883    & 0.493    & 0.833    \\
			RF                       & 0.570    & 0.822    & 0.337    & 0.622    & 0.784    & 0.414    & 0.622    & 0.784    & 0.414    & 0.622    & 0.784    & 0.414    & 0.623    & 0.783    & 0.414    \\
			SVR                      & 0.525    & 0.854    & 0.241    & 0.572    & 0.836    & 0.340    & 0.574    & 0.835    & 0.341    & 0.594    & 0.822    & 0.363    & 0.660    & 0.774    & 0.458    \\
			ANN                      & 0.436    & 0.902    & 0.233    & 0.442    & 0.898    & 0.162    & 0.440    & 0.899    & 0.169    & 0.444    & 0.897    & 0.166    & 0.441    & 0.898    & 0.201    \\
			Random CNN               & 0.441    & 0.903    & 0.222    & 0.392    & 0.921    & 0.179    & 0.385    & 0.923    & 0.143    & 0.384    & 0.924    & 0.156    & 0.413    & 0.911    & 0.198    \\
			PCA CNN                  & 0.443    & 0.901    & 0.179    & 0.401    & 0.918    & 0.181    & 0.400    & 0.918    & 0.207    & 0.395    & 0.922    & 0.163    & 0.425    & 0.909    & 0.237    \\
			REFINED CNN              & 0.414    & 0.911    & 0.197    & 0.381    & 0.925    & 0.163    & 0.388    & 0.922    & 0.148    & 0.377    & 0.927    & 0.169    & 0.402    & 0.917    & 0.206   
		\end{tabular}%
	}
\end{table}

\begin{figure}[!htbp]
	\centering
	\includegraphics[width=\textwidth]{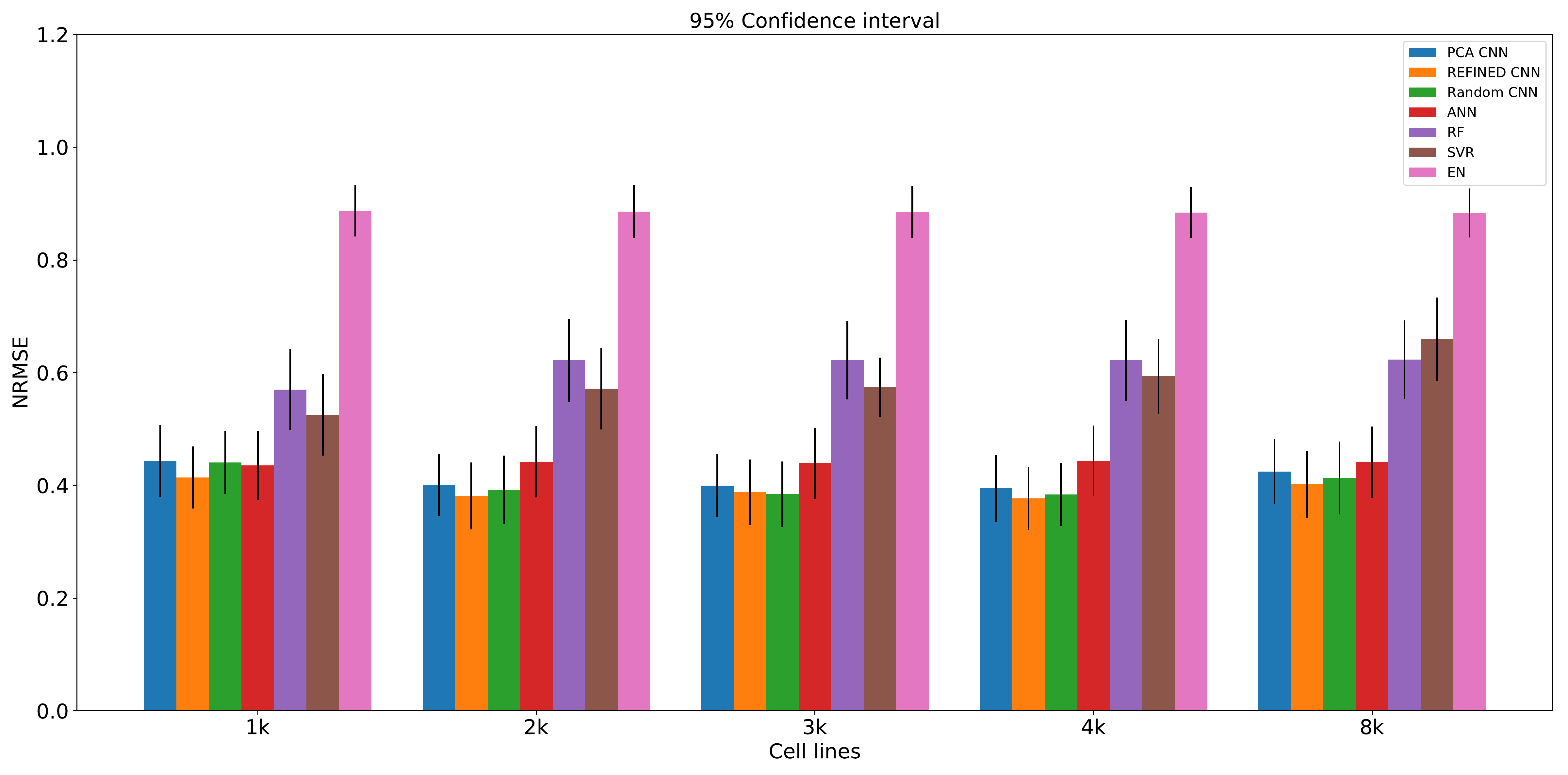} 
	\caption{NRMSE with 95 \% confidence interval of each model trained on GDSC dataset.}
	\label{GDSC_NRMSE_Conf}  
\end{figure}

\begin{figure}[!htbp]
	\centering
	\includegraphics[width=\textwidth]{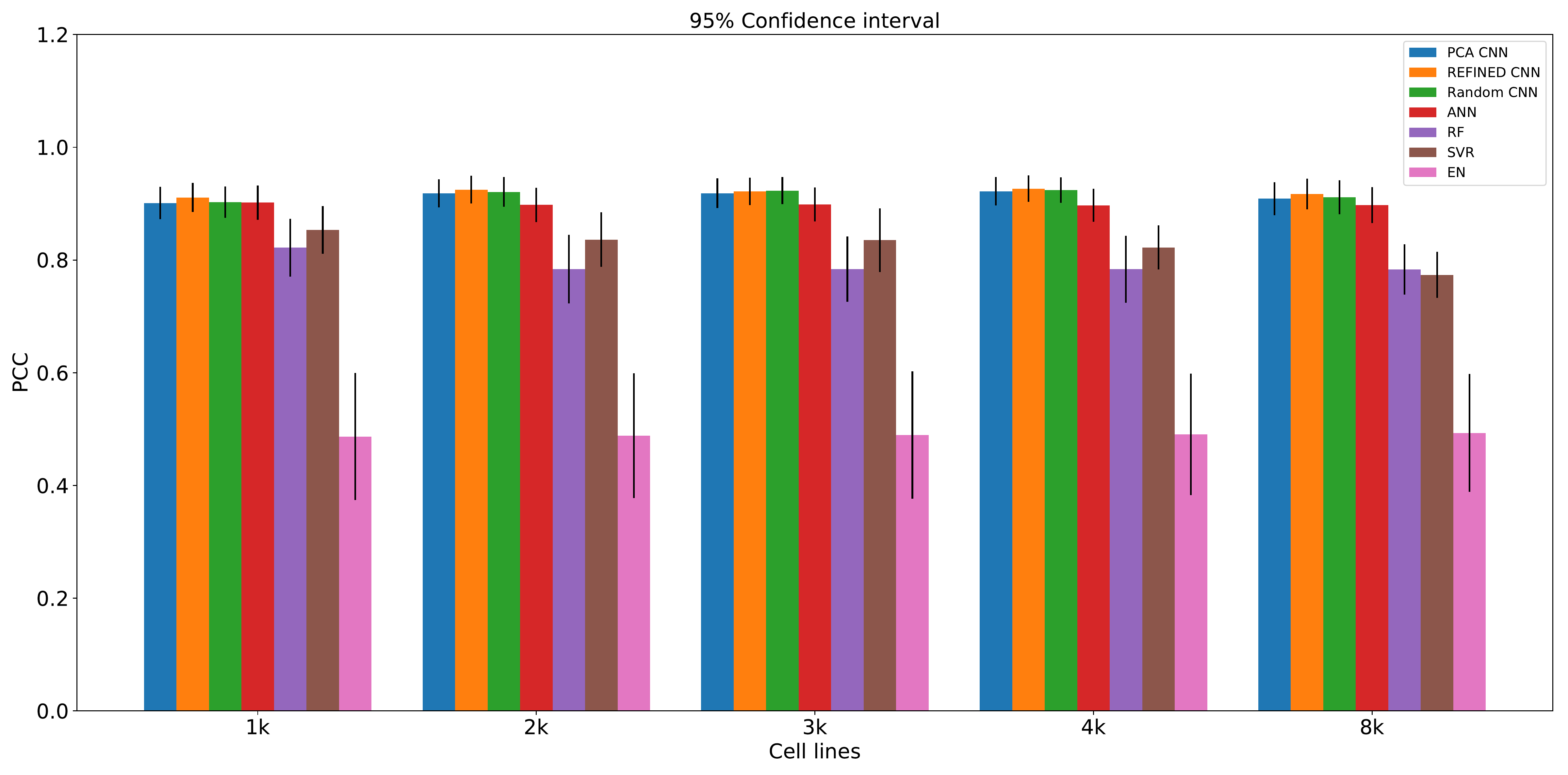} 
	\caption{PCC with 95 \% confidence interval of each model trained on GDSC dataset.}
	\label{GDSC_PCC_Conf}  
\end{figure}

\begin{figure}[!htbp]
	\centering
	\includegraphics[width=\textwidth]{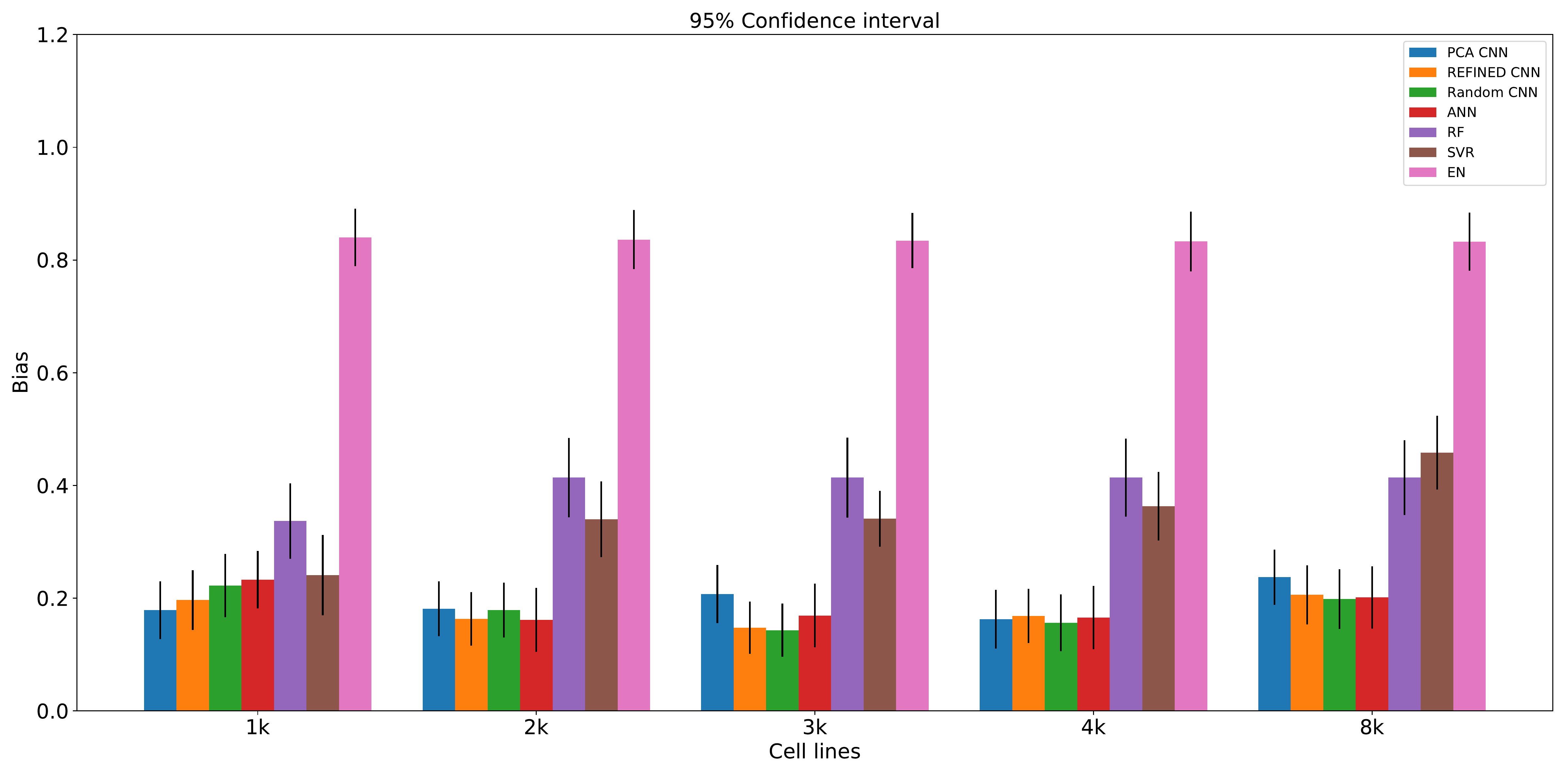} 
	\caption{Bias with 95 \% confidence interval of each model trained on GDSC dataset}
	\label{GDSC_Bias_Conf}  
\end{figure}

\newpage
\subsection{REFINED CNN architectures}

\begin{figure}[!htbp]
	\begin{multicols}{2}
		\includegraphics[width=\linewidth,height=0.8\textheight]{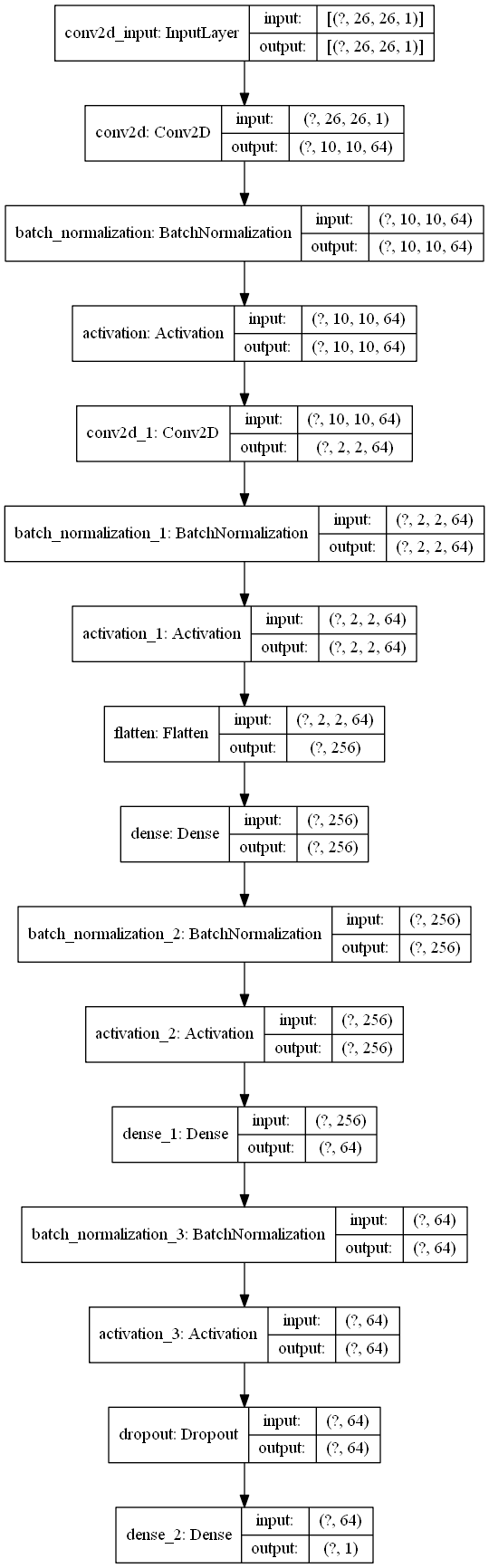}\par 
		\includegraphics[width=\linewidth,height=0.8\textheight]{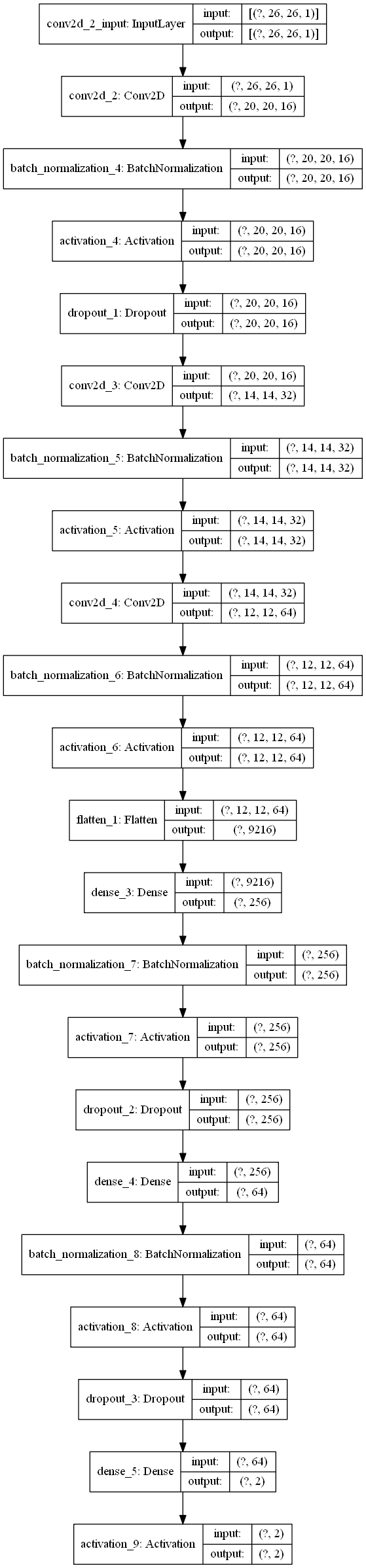}\par 
	\end{multicols}
	\caption{The architecture of the CNN models used to train the NCI60 dataset (left model is the CNN regressor, right model is the CNN classifier). Plot created by 'graphviz' utility of keras \cite{chollet2015keras}.}
	\label{NCI_Arch}
\end{figure}

\begin{figure}[]
	\centering
	\includegraphics[width=\textwidth,height=0.95\textheight]{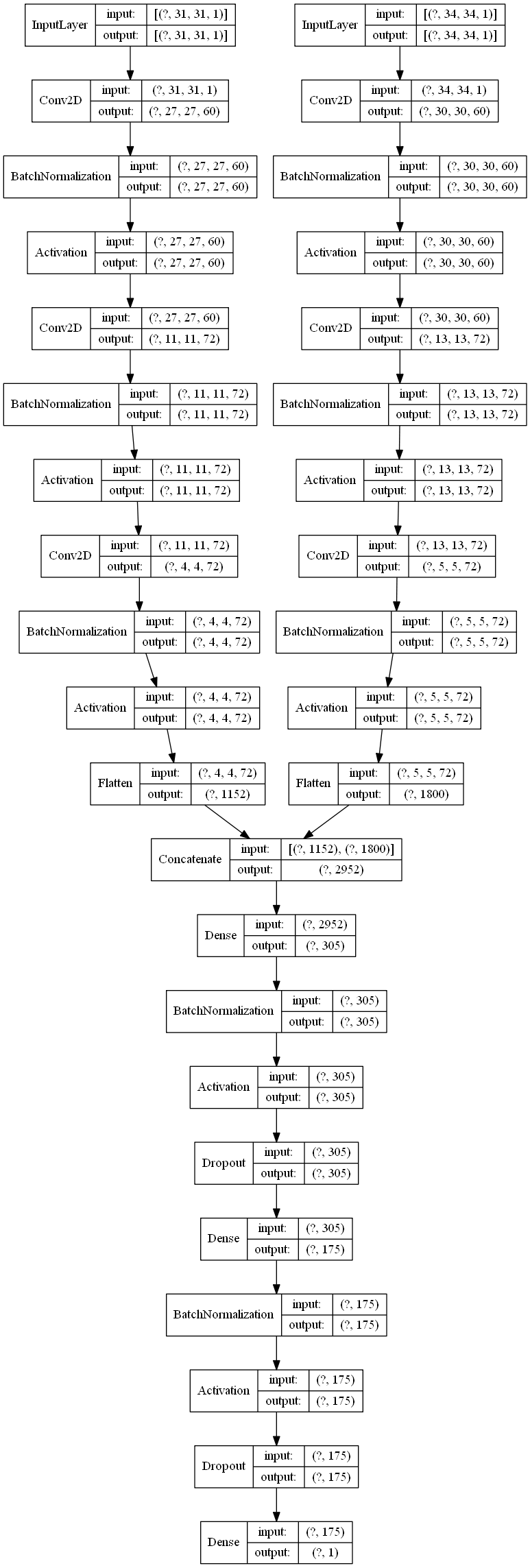} 
	\caption{The architecture of the CNN model used to train the GDSC dataset. Plot created by 'graphviz' utility of keras \cite{chollet2015keras}.}
	\label{GDSC_Arch}  
\end{figure}

\end{subappendices}

\end{document}